\documentclass[twoside]{article}

\usepackage{PRIMEarxiv}

\usepackage[utf8]{inputenc} 
\usepackage[T1]{fontenc}    
\usepackage{url}            
\usepackage{booktabs}       
\usepackage{amsfonts}       
\usepackage{nicefrac}       
\usepackage{microtype}      
\usepackage{lipsum}
\usepackage{fancyhdr}       
\usepackage{graphicx}       

\usepackage{amsmath,amssymb}
\usepackage[colorlinks=true, allcolors=blue]{hyperref}
\usepackage{amsthm}
\usepackage{mathtools}
\usepackage{color}
\usepackage{cite}
\usepackage[export]{adjustbox}
\usepackage{tikz}
\usetikzlibrary{angles,quotes}
\usepackage{subcaption}
\usepackage{caption}
\usepackage{multirow,makecell}
\usepackage{bbold}
\usepackage{ mathrsfs }
\usepackage{float}
\usepackage{enumitem}

\title{Synthetic Defect Geometries of Cast Metal Objects Modeled via 2d Voronoi Tessellations
}

\author{
  Natascha Jeziorski \\
  RPTU University Kaiserslautern-Landau\\
  Fraunhofer Institute for Industrial Mathematics ITWM, Kaiserslautern\\
  \And
  Petra Gospodneti\'c\\
  Fraunhofer Institute for Industrial Mathematics ITWM, Kaiserslautern\\
  \And
  Claudia Redenbach \\
  RPTU University Kaiserslautern-Landau
}

\begin{document}
\maketitle

\begin{abstract}
In industry, defect detection is crucial for quality control.
Non-destructive testing (NDT) methods are preferred as they do not influence the functionality of the object while inspecting.
Automated data evaluation for automated defect detection is a growing field of research.
In particular, machine learning approaches show promising results.
To provide training data in sufficient amount and quality, synthetic data can be used.
Rule-based approaches enable synthetic data generation in a controllable environment.
Therefore, a digital twin of the inspected object including synthetic defects is needed.
We present parametric methods to model 3d mesh objects of various defect types that can then be added to the object geometry to obtain synthetic defective objects.
The models are motivated by common defects in metal casting but can be transferred to other machining procedures that produce similar defect shapes.	
Synthetic data resembling the real inspection data can then be created by using a physically based Monte Carlo simulation of the respective testing method.
Using our defect models, a variable and arbitrarily large synthetic data set can be generated with the possibility to include rarely occurring defects in sufficient quantity.
Pixel-perfect annotation can be created in parallel.
As an example, we will use visual surface inspection, but the procedure can be applied in combination with simulations for any other NDT method.
\end{abstract}

\keywords{Stochastic geometry modeling \and Defect modeling \and Voronoi tessellations \and Synthetic training data \and Quality control}

\section{Introduction}    
    Visual surface inspection is a commonly used method for quality control in industry. 
    In general, the demands on inspection systems are robustness, reliability, objectivity, and speed. 
    Manual defect detection is widely used, but is slow, subjective, and expensive. 
    Automated inspection systems using machine learning approaches offer a promising alternative \cite{Bozic2021KSDD2,Honzatko2021CSEM}. 
    Classical image processing algorithms are also frequently used for automatic defect detection \cite{Luo2020OverviewClassicalDetection}, but they are often less flexible in adapting to new inspection setups than learning-based methods.
    However, both automated inspection methods using machine learning approaches and people performing manual inspections need to be trained to detect production-relevant defects.
    Thus, example data of defective objects acquired by visual inspection systems are required.
    
    For training of inspection workers, it is sufficient to have a specification of standards and a few examples, whereas for machine learning models, one needs thousands of examples.
    This is problematic in applications where real data are not available in the quantity and diversity needed for reliable defect detection. 
    It is impossible to account for the large variety and variability of possible defects as manufacturers try to prevent defect formation during production. 
    In particular, safety-related defects occur rarely as they are avoided best possible.
    Moreover, the data have to be annotated to be used for training machine learning approaches.
    Typically, annotation is done manually, which requires the knowledge of domain experts and is extremely cumbersome and time-consuming.
    
    Synthetic data that resemble the appearance of inspection data can overcome this problem \cite{nikolenko2021synthetic}. 
    AI-based techniques are of increasing interest for the generation of synthetic image data \cite{Jain2020Steel,Zhang2021DefectGAN,Wei2023GAN}.
    While generative methods increase the quantity of the data set, the diversity of the synthetic data is restricted to the diversity of the training data.
    Moreover, learning-based approaches are known to introduce hallucinations \cite{Schmedemann2023}, which may have a negative impact on subsequent training.
    In addition, synthetic images generated using generative methods are not always directly annotated and require subsequent labeling.
    
    Another method of increasing the amount of image data showing defective areas is to insert synthetic defects into real images of defect-free samples.
    Haselmann and Gruber \cite{Haselmann2019defect_segmentation} present a fake sample generator that adds simple fabric defects to visual image data.
    For the detection of cracks in concrete using micro-computed tomography (\textmu CT) imaging, a database of semi-synthetic images is available \cite{Jung2024Dataset}.
    The images are obtained by simulating crack geometries using a stochastic model and then inserting them into real \textmu CT images of concrete.
    
    An alternative approach for the generation of synthetic data is to use rule-based methods, meaning to physically simulate the inspection environment using techniques from computer graphics \cite{Dahmen2019digital_reality_survey,Tsirikoglou2020Survey,2024MicroStruc}.
    This includes simulations of the considered inspection method and of the object to be inspected.
    Approaches for CT images of cast aluminum objects \cite{Fuchs2021SynCT,2023synthCT} or 3d woven composites \cite{2025synthCTcfk} and visual surface inspection of metal surfaces \cite{boikov2021synthetic,Roovere2024,abubakr2021learning,kim2022synthetic,Fulir2023Kupplung,Fulir2025SynosIs} exist.
    Advantages of this procedure are the automatic pixel-perfect annotation of the synthetic data and fast data generation in sufficient quantity.
    Furthermore, various synthetic defects can be added to the object's digital twin to generate a diverse data set of defective products.
    This requires models to generate defect geometries of different defect types.
    
    We restrict on cast metal objects and their typical defects.
    Jolly and Katgerman \cite{2021JOLLYmodelingCastDefects} give an extensive overview of modeling approaches for various defect types of cast aluminum products.
    Most methods are based on modeling the physics of the casting process including the temperature difference during the filling of the mold and the cooling of the material.
    Physically correct defect geometries are provided by investigating the physical material properties.
    Including the crystalline structure of the material and its behavior under stress results in complex defect models.
    Preliminary studies or expert knowledge are needed to obtain exact information on the composition and strength of the used material such that those specialized models cannot easily be transferred to other applications.
    Sistaninia et al. \cite{2013Sistaninia_Hottear} propose a detailed model for crack formation as metal solidifies.
    They include the poly-crystalline structure of metal and the heat change during the cooling process.
    
    Approaches based on computer graphic techniques offer an alternative.
    These methods do usually not rely on the physical material properties and the physically correct forming process, but rather model realistically appearing defects in an artistic way.
    Examples are provided by Bosch \cite{Bosch2007Phd} for modeling surface grooves and scratches and by Bosnar and Gospodneti\'c \cite{Bosnar2023Defects} for surface dents and scratches.
    Merillou et al. \cite{2001scratchesMerillou} proposed a model for scratch simulation integrating real-world measurements.
    Moreover, Fan et al. \cite{Fan2022BrittleFracture} provide an approach for modeling brittle material fracturing of the object.
    Therefore, deformations of the discretization of the 3d object are computed, which is commonly given by tetrahedrons.
    Instead of adding individual defects, the whole object is usually destroyed and broken down into parts.
    
    Fulir et al. \cite{Fulir2023Kupplung} have shown how synthetic data generated in a virtual inspection environment can improve learning-based defect detection for visual inspection, especially if only little real data is available.
    They achieve the best results when training deep learning models with synthetic data, followed by fine-tuning with real data to overcome the domain gap.
    
    \subsection{Our contribution}
\begin{figure}
        \centering
        \includegraphics[width=\textwidth]{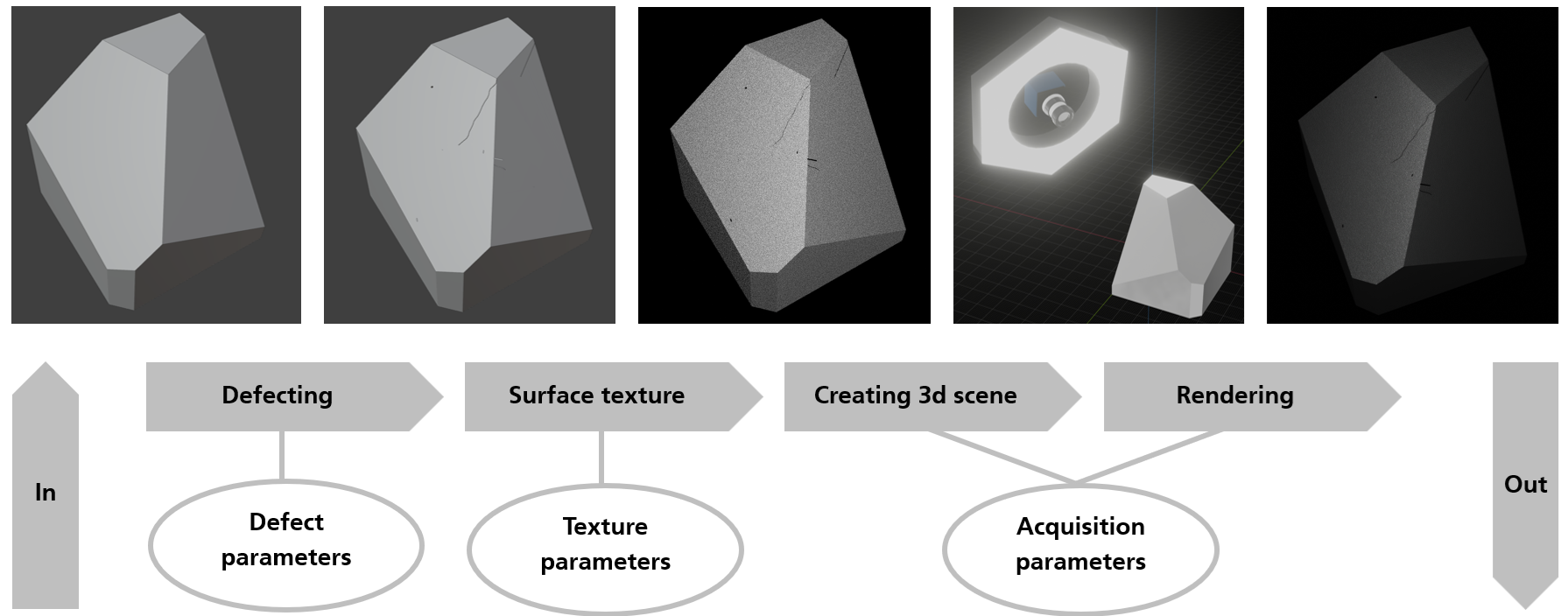}
        \caption{Visualization of the synthetic data generation pipeline using rendering in a virtual environment for visual surface inspection.}
        \label{fig:virtual_environment}
    \end{figure}

    We focus on visual surface inspection, where synthetic images can be generated using the pipeline presented in \cite{Gospodnetic2020Pipeline,Bosnar2020ImageSynthesis,Fulir2025SynosIs}, see Figure \ref{fig:virtual_environment} for illustration.
    Therefore, digital twins of real samples are placed in a virtual inspection environment resembling the real inspection set up consisting of a suitable camera and light sources. Defect geometries can be imprinted into the object geometry of any object to be inspected if its surface triangulation is provided.
    Then, physically-based rendering is performed to generate synthetic data.
    That is, the imaging process of the real camera is simulated using Monte Carlo methods, so taking 2d images of a 3d scene.
    Note that pixel-perfect annotation can be created in parallel, since shapes and locations of defects are known.
    
    We propose parametric mathematical models to generate various types of surface defects that occur in cast metal objects.
    The output of our defect models are 3d mesh structures given as surface triangulations. 

\begin{figure}
    	\centering
        \begin{subfigure}[t]{0.49\textwidth}
            \subcaptionbox*{Cracks}{\includegraphics[height=3.3cm]{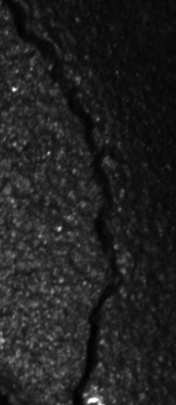}
    		\includegraphics[height=3.3cm]{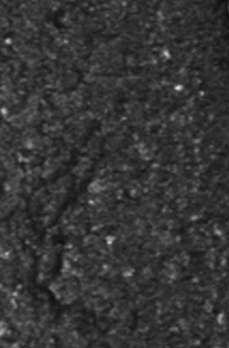}} \hfill
            \subcaptionbox*{Cold shut}{\includegraphics[height=3.3cm]{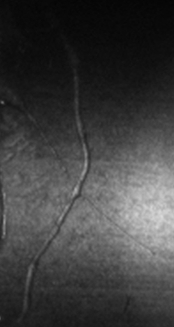}} \hfill
    	    \subcaptionbox*{Bulge}{\includegraphics[height=3.3cm]{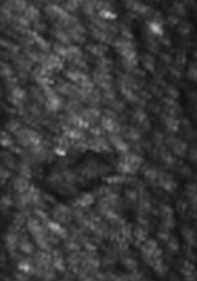}}
           \caption*{Images provided by Fraunhofer ITWM.}
        \end{subfigure}\hfill
    	\begin{subfigure}[t]{0.49\textwidth}
    	    \subcaptionbox*{\centering Lustrous carbon inclusion (similar to coat lift). Image taken as crop from Figure 38a in \cite{2022SertuchaDefectExp}.}{\includegraphics[height=3.3cm]{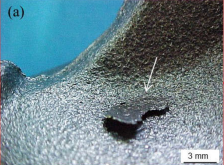}}\hfill
    	    \subcaptionbox*{\centering Scab. Image taken as crop from Figure 23c in \cite{2022SertuchaDefectExp}.}{\includegraphics[height=3.3cm]{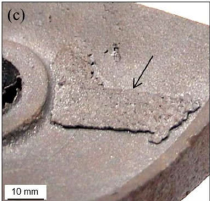}}
    	\end{subfigure}
    	\caption{Examples of common defects in cast metal objects.}
    	\label{fig:ExampleDefects}
    \end{figure}

    Casting is a manufacturing process in which liquid metal is poured into the object mold and then cooled down to solidify.
    Defects can be caused by a number of reasons: a defective mold, unsuitable object geometries, inappropriate material temperature that influences the fluidity, or too cold ambient air temperature that leads to rapid cooling of the material.
    We distinguish between elongated defect types (cold cracks, hot tears, buckles, bulges, coat lifts, cold shuts, and rat tails) and scab defects, see \cite{2022SertuchaDefectExp,AtlasCastingDefects} for an overview.
    Figure \ref{fig:ExampleDefects} shows example images of some defect types.
    The development of the proposed models is motivated by the observation of defect structures and the description of defect formation processes \cite{AtlasCastingDefects,GiessLexikon,2022SertuchaDefectExp,2021JOLLYmodelingCastDefects}.
    
    Using stochastic modeling based on random Voronoi tessellations, we introduce two different pipelines for modeling the elongated and scab defects.
    All elongated defect types can be generated form the same pipeline by adapting the individual model steps.
    The models are parameterized to guarantee defect variation using easily interpretable parameter configurations.
    Furthermore, sampling the parameter values from suitable random distributions ensures variable defect geometries.
    In this way, defect characteristics such as length, width, and depth can be controlled and adapted to the desired parameter ranges.
    The data set content is controllable in terms of defect properties and appearance, also edge-case scenarios can easily be included.
    Thus, it is possible to systematically vary individual aspects such that the resulting synthetic data can be used to train detection or classification algorithms for specific tasks. 
    This provides a significant advantage over AI-based methods, where systematic data generation and control is difficult to perform.
   
    Although we consider casting defects, similar defect shapes can also occur in other manufacturing processes.
    Therefore, our models can be transferred to other applications by using the same synthetic defect geometries but adapting the virtual object in terms of material and surface texture.

\section{Models for elongated defects} \label{sec:autoDefects}
\begin{figure}
		\centering
        \resizebox{\textwidth}{!}{
        \begin{tikzpicture}
		  \node[align=center] at (0,1.9) {\textbf{1d shape}\\Path generation};
		  \node[align=center] at (5.5,1.9) {\textbf{2d shape}\\Assignment of width};
		  \node[align=center] at (11,1.9) {\textbf{3d shape}\\Assignment of height and 3d mesh};
			
		  \node[rotate=90] at (13.7,0) {\textbf{Cold crack}};
	   	\node[rotate=90] at (14.05,0) {\textbf{Hot tear}};
	   	\node[inner sep=0pt] (whitehead) at (0,-3) {\includegraphics[width=.27\textwidth,cframe=black 1pt 0pt]{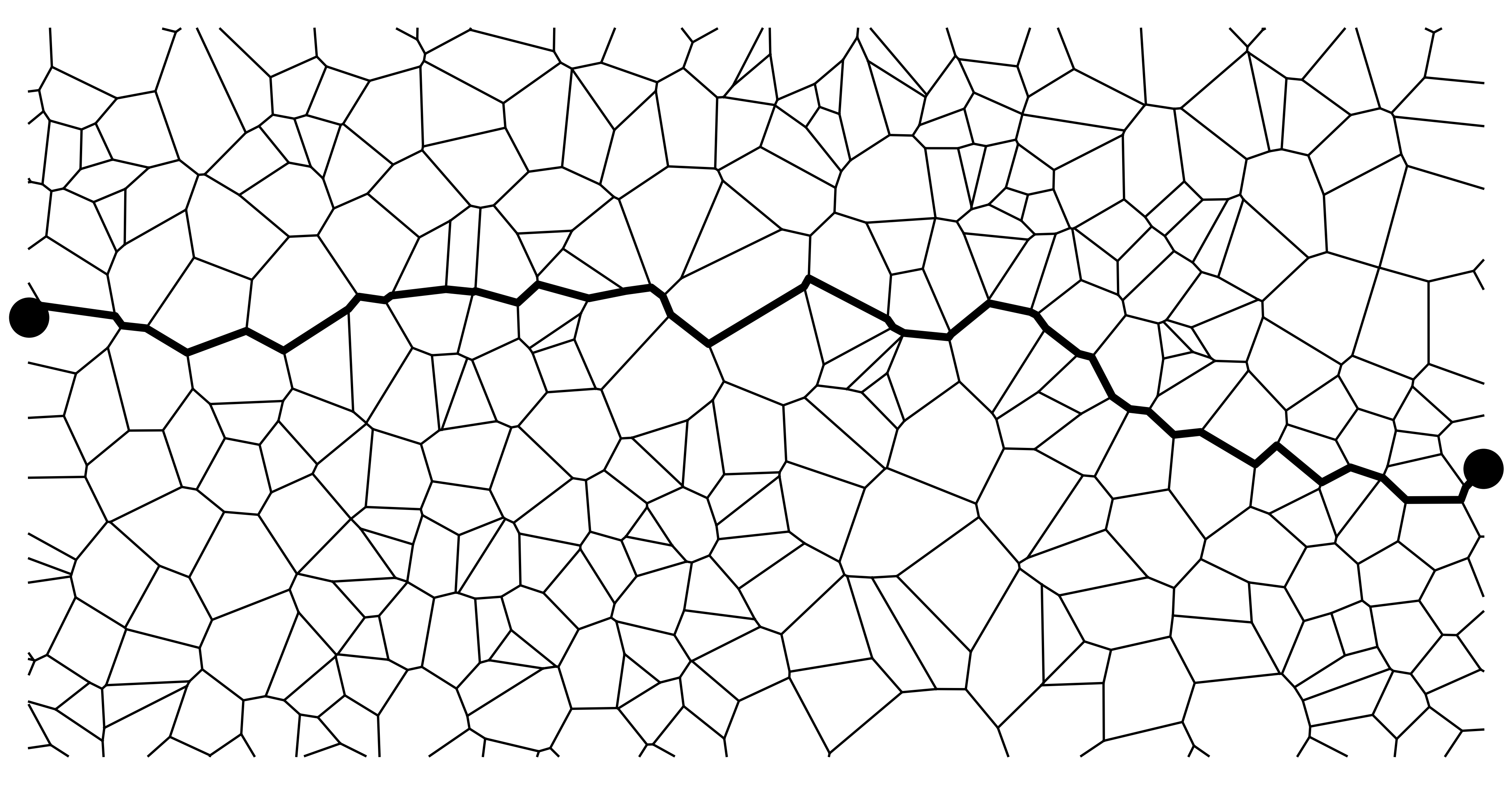}};
	   	\node[inner sep=0pt] (whitehead) at (5.5,0) {\includegraphics[width=.27\textwidth,cframe=black 1pt 0pt]{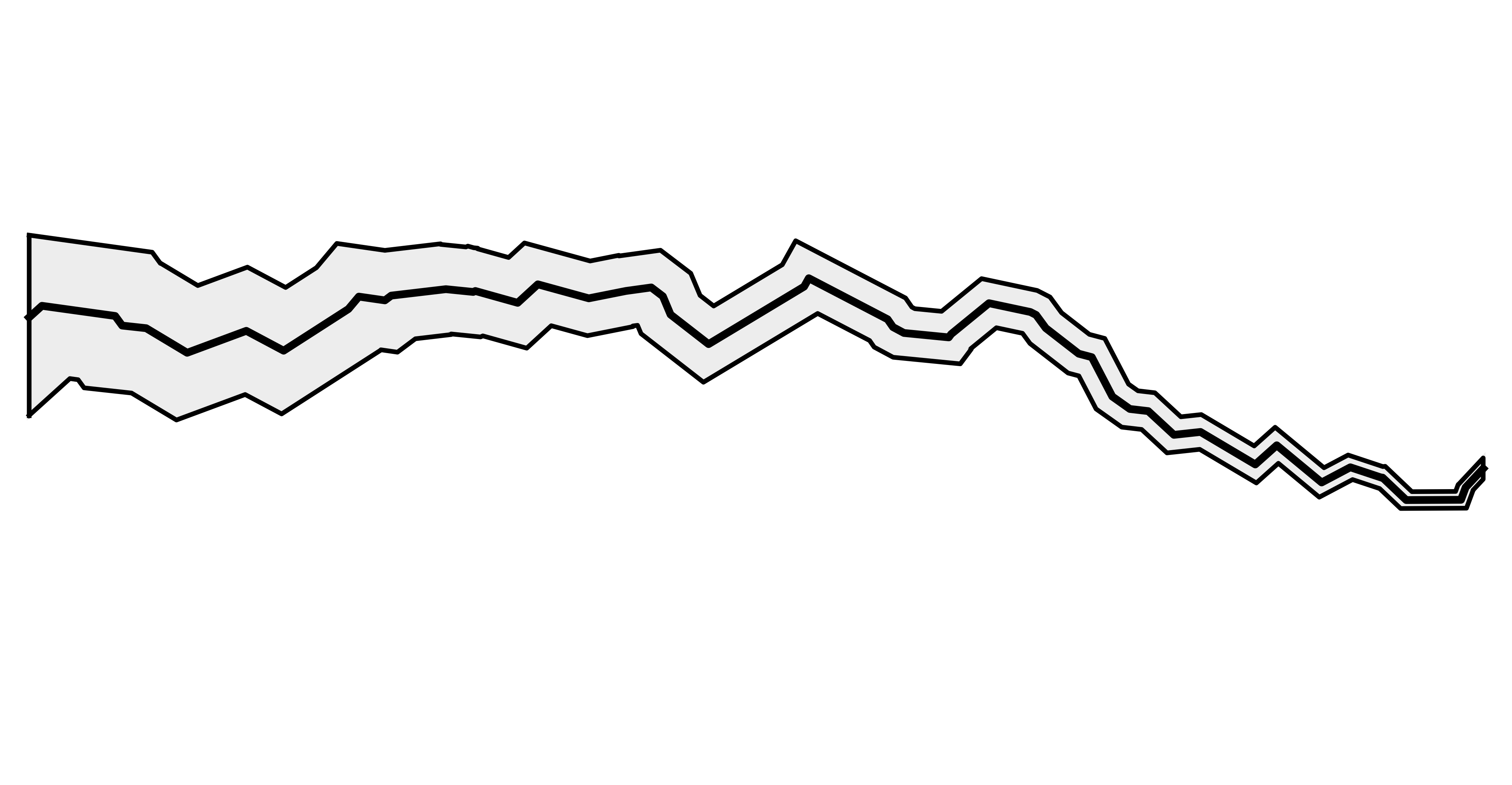}};
	   	\node[inner sep=0pt] (whitehead) at (11,0) {\includegraphics[width=.27\textwidth,cframe=black 1pt 0pt]{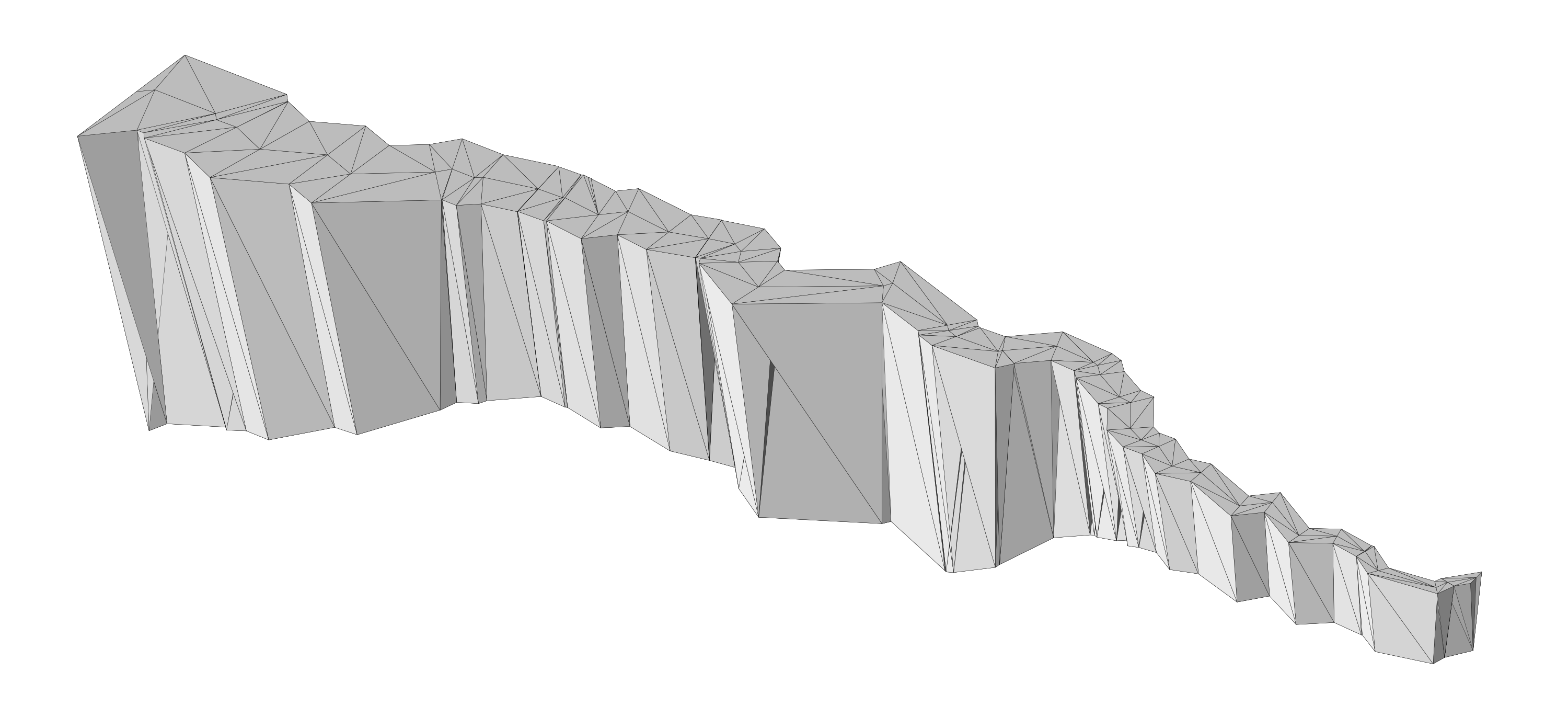}};
			
	   	\node[rotate=90] at (13.7,-3) {\textbf{Buckle}};
	   	\node[inner sep=0pt] (whitehead) at (11,-3) {\includegraphics[width=.27\textwidth,cframe=black 1pt 0pt]{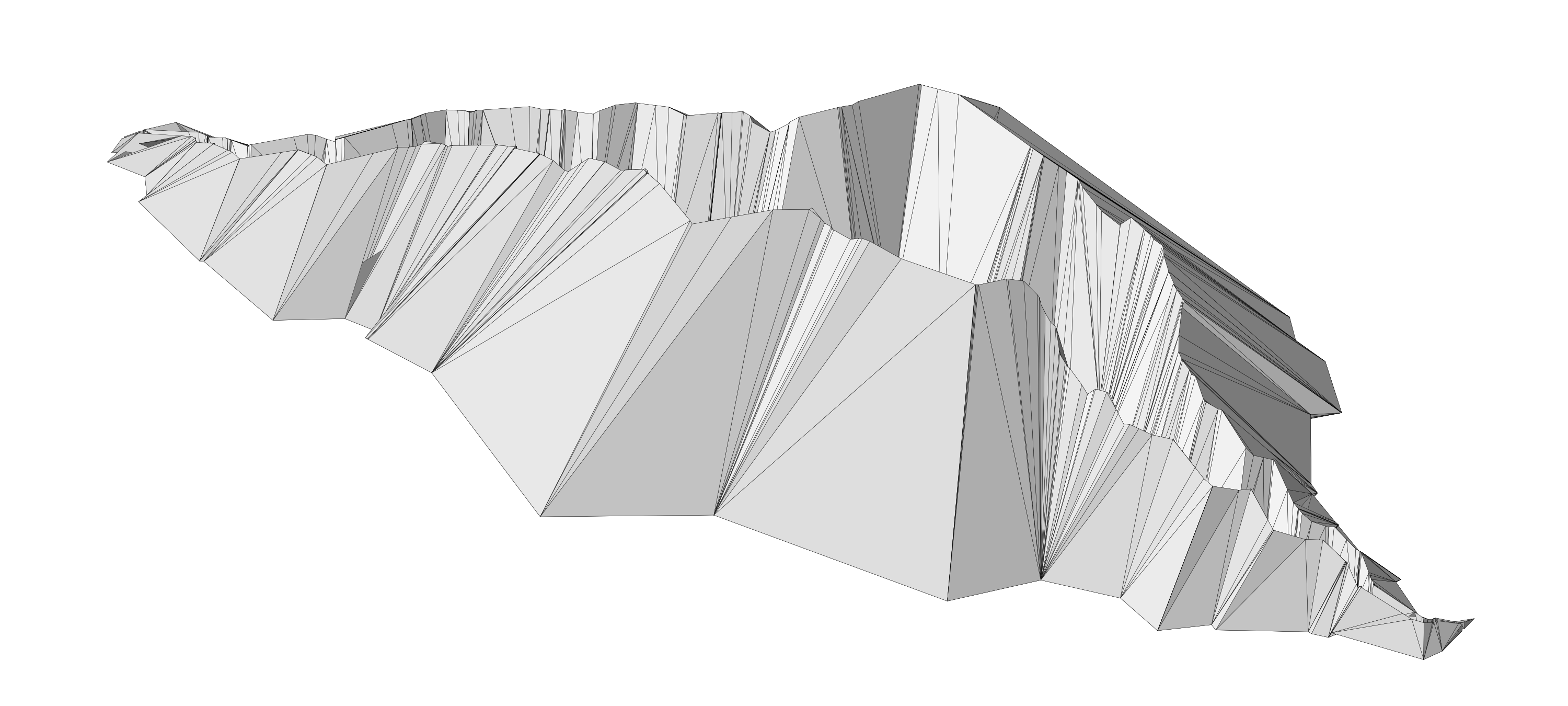}};
			
	   	\node[rotate=90] at (13.7,-6) {\textbf{Bulge}};
	   	\node[inner sep=0pt] (whitehead) at (5.5,-6) {\includegraphics[width=.27\textwidth,cframe=black 1pt 0pt]{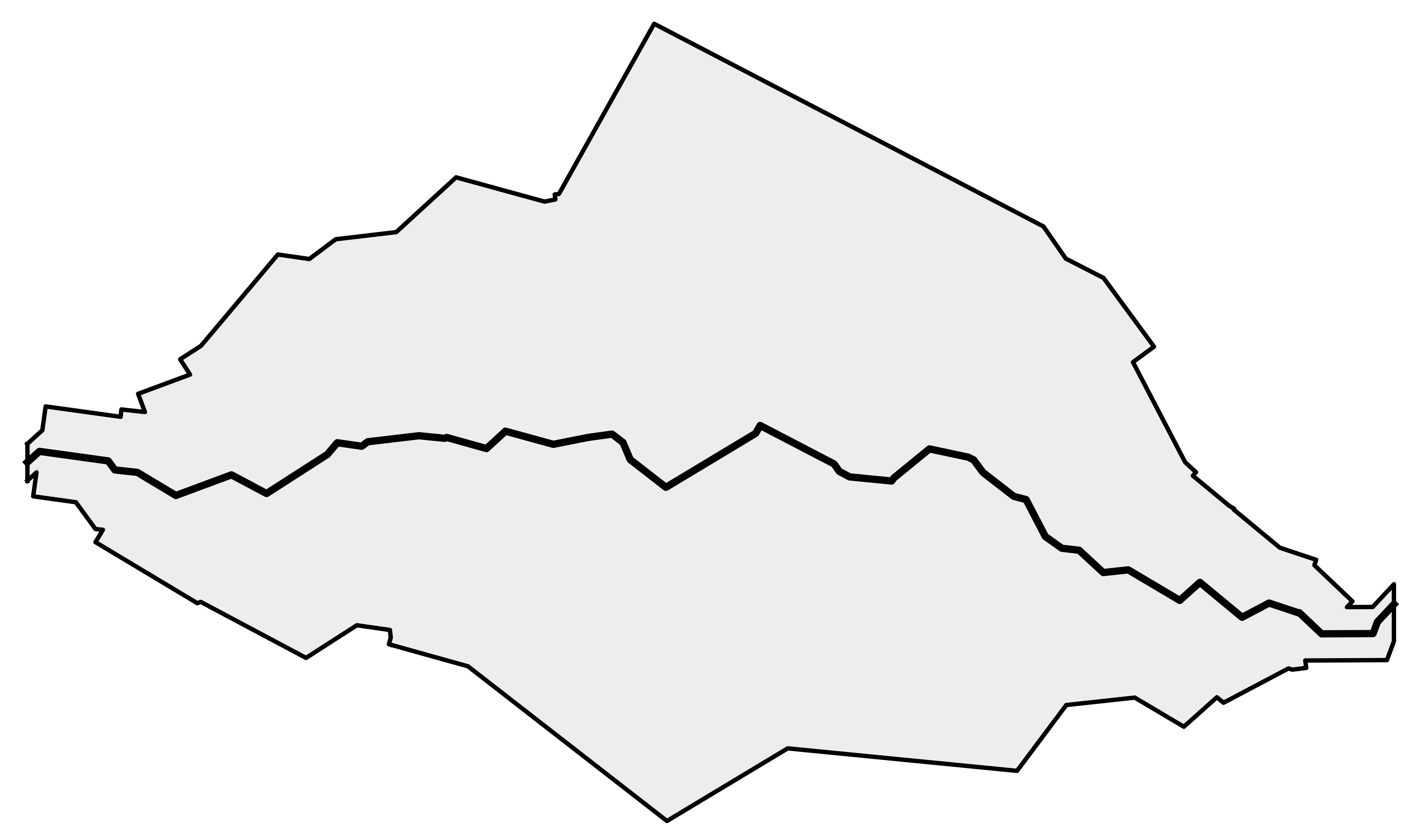}};
	   	\node[inner sep=0pt] (whitehead) at (11,-6) {\includegraphics[width=.27\textwidth,cframe=black 1pt 0pt]{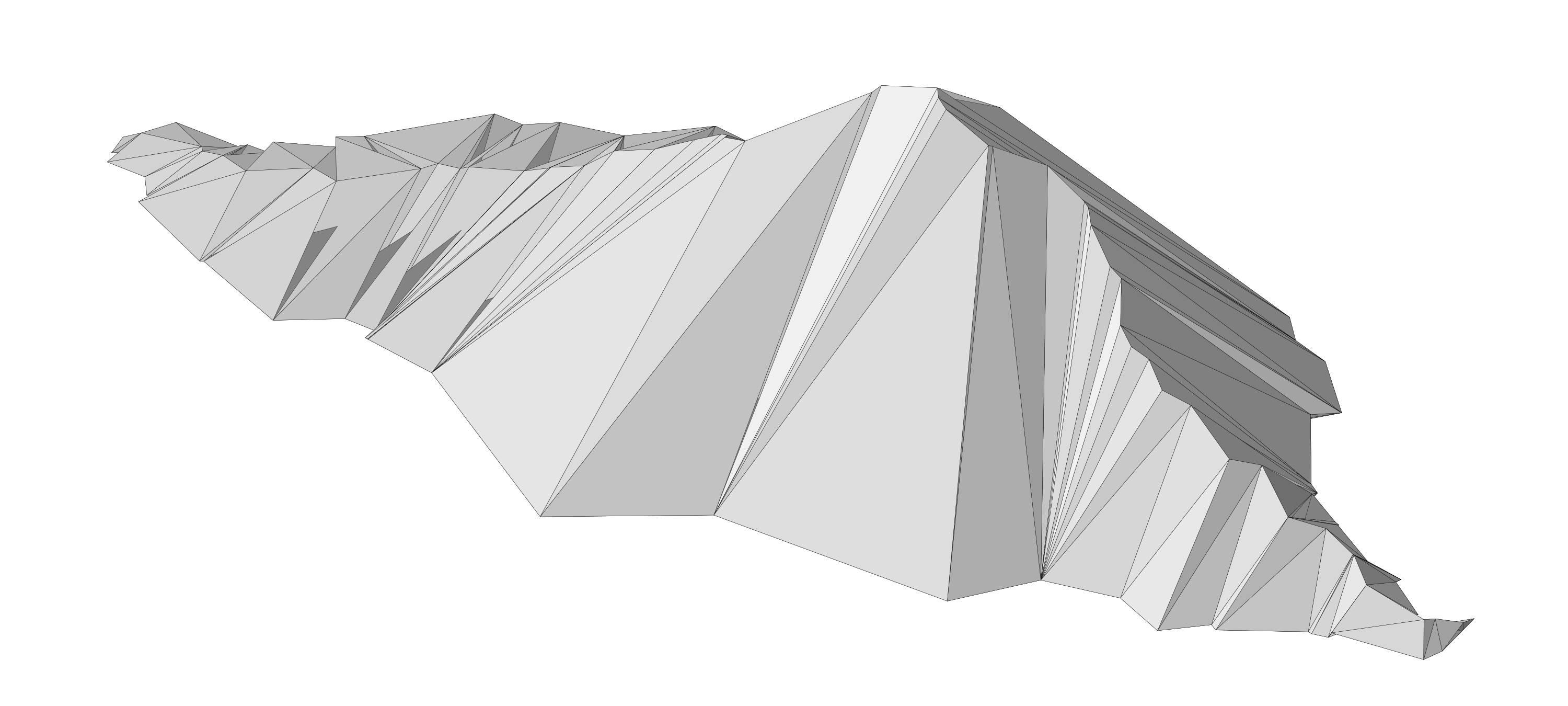}};
			
	   	\node[rotate=90] at (13.7,-9) {\textbf{Coat lift}};
	   	\node[inner sep=0pt] (whitehead) at (5.5,-9) {\includegraphics[width=.27\textwidth,cframe=black 1pt 0pt]{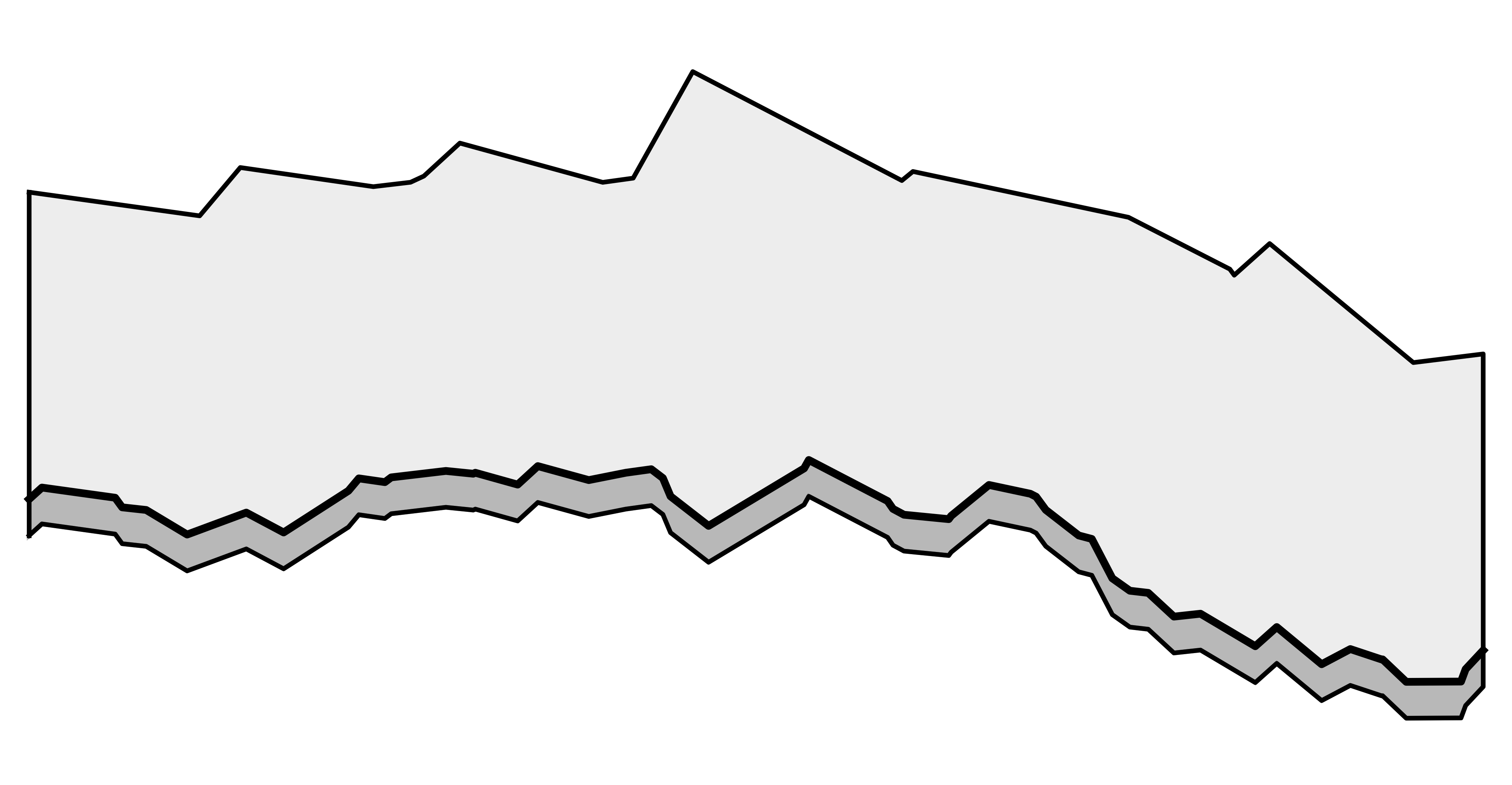}};
	   	\node[inner sep=0pt] (whitehead) at (11,-9) {\includegraphics[width=.27\textwidth,cframe=black 1pt 0pt]{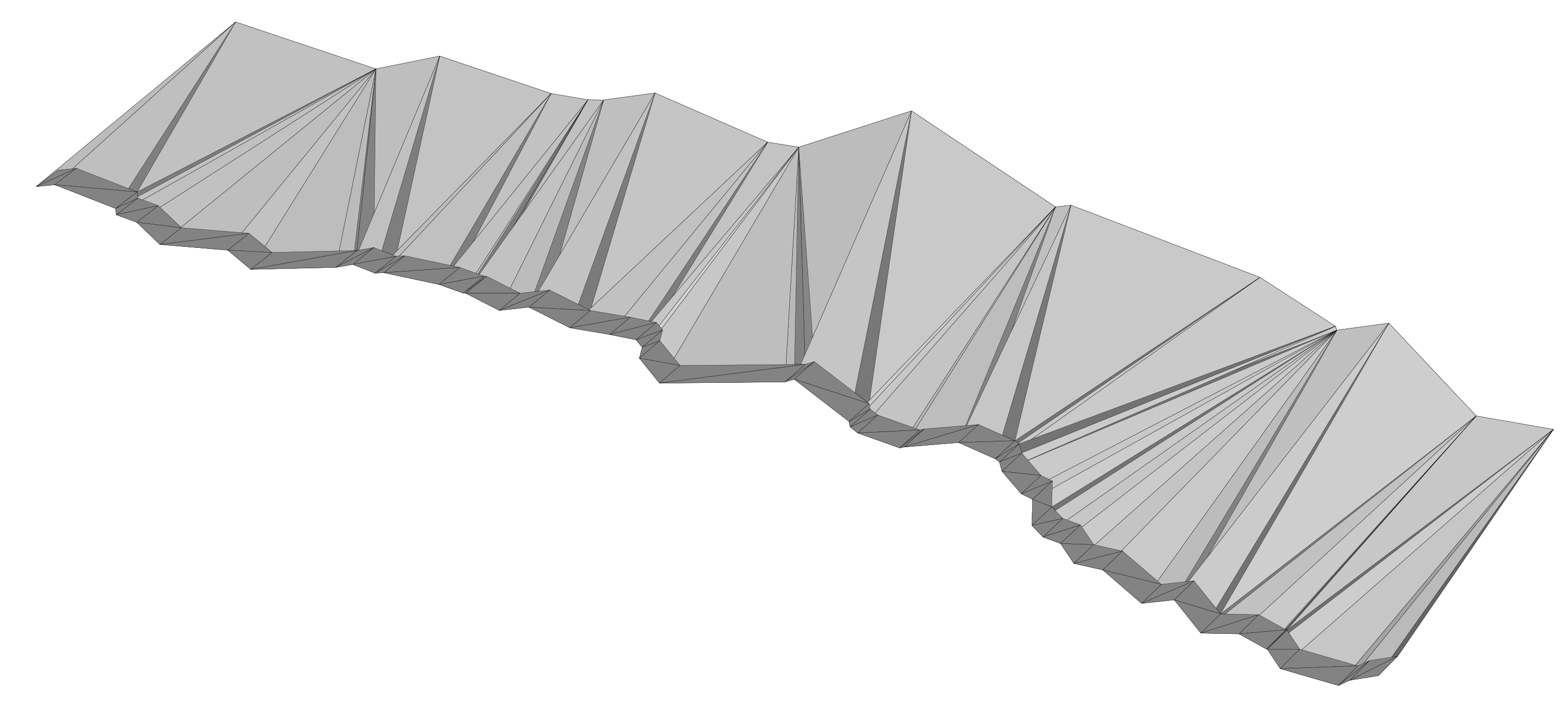}};
			
	   	\node[rotate=90] at (13.7,-12.2) {\textbf{Cold shut}};
	   	\node[rotate=90] at (14.05,-12.2) {\textbf{Rat tail}};
	   	\node[inner sep=0pt] (whitehead) at (0,-12.2) {\includegraphics[width=.27\textwidth,cframe=black 1pt 0pt]{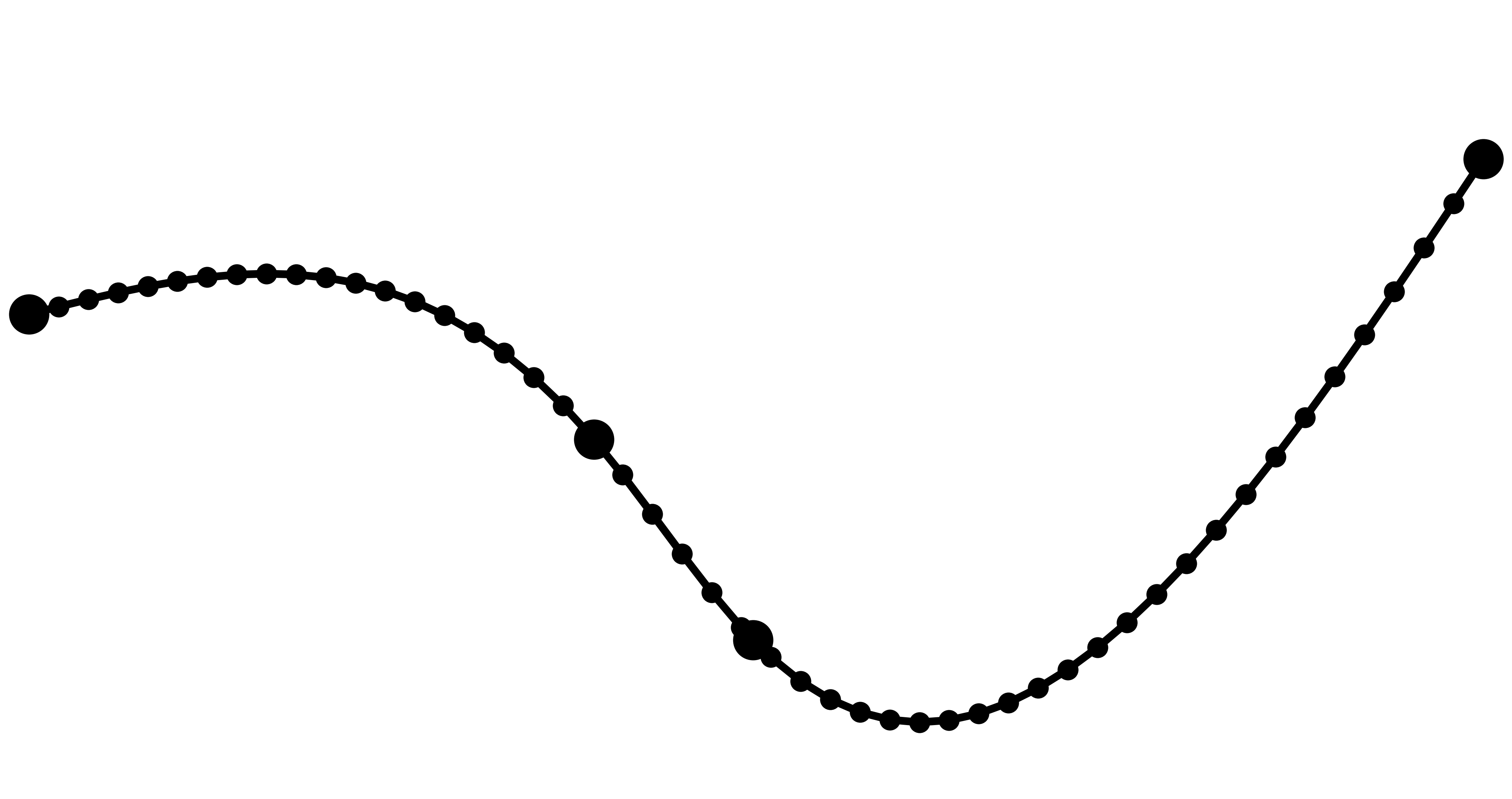}};
	   	\node[inner sep=0pt] (whitehead) at (5.5,-12.2) {\includegraphics[width=.27\textwidth,cframe=black 1pt 0pt]{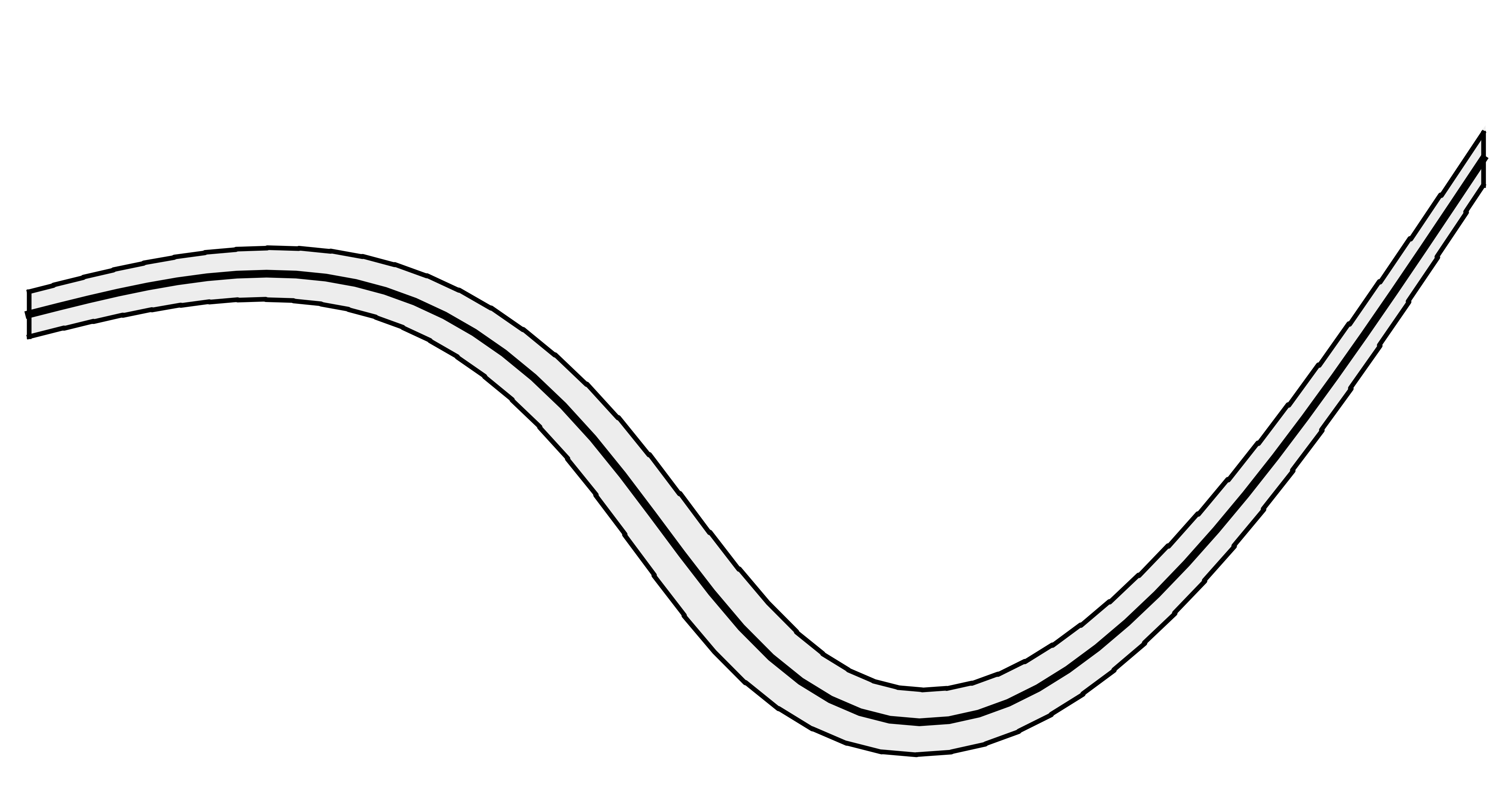}};
	   	\node[inner sep=0pt] (whitehead) at (11,-12.2) {\includegraphics[width=.27\textwidth,cframe=black 1pt 0pt]{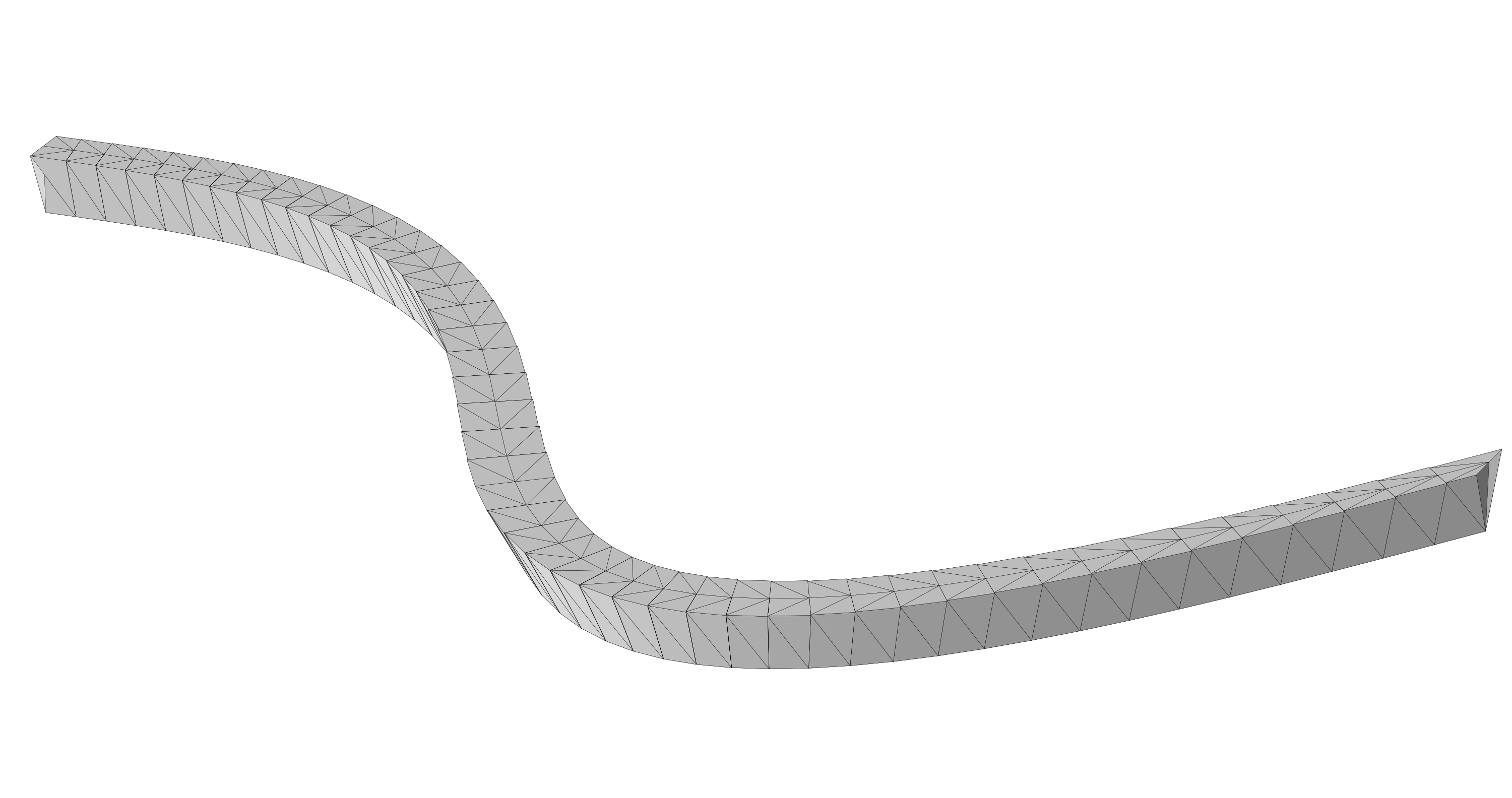}};
			
	   	\draw[line width=1pt] (2.25,-1.85) -- (3.25,0);
	   	\draw[line width=1pt] (2.25,-4.15) -- (3.25,-6);
	   	\draw[line width=1pt] (2.25,-4.15) -- (3.25,-9);
	   	\draw[line width=1pt] (2.25,-12.2)   -- (3.25,-12.2);
			
	   	\draw[line width=1pt] (7.78,0) -- (8.72,0);
	   	\draw[line width=1pt] (7.78,-6) -- (8.72,-6);
	   	\draw[line width=1pt] (7.78,-9) -- (8.72,-9);
	   	\draw[line width=1pt] (7.78,-12.2) -- (8.72,-12.2);
			
	   	\draw[line width=1pt] (11,-1) -- (11,-1.95) node [midway, right] (TextNode) {\Large$\ominus$};
	   	\draw[line width=1pt] (11,-4.05) -- (11,-4.95) node [midway, right] (TextNode) {\Large$\oplus$};
		\end{tikzpicture}}
		\caption{Overview of modeling pipeline including the different steps, adaptation to and relations between different defect types.}
		\label{fig:LinDefectsOverview}
    \end{figure}
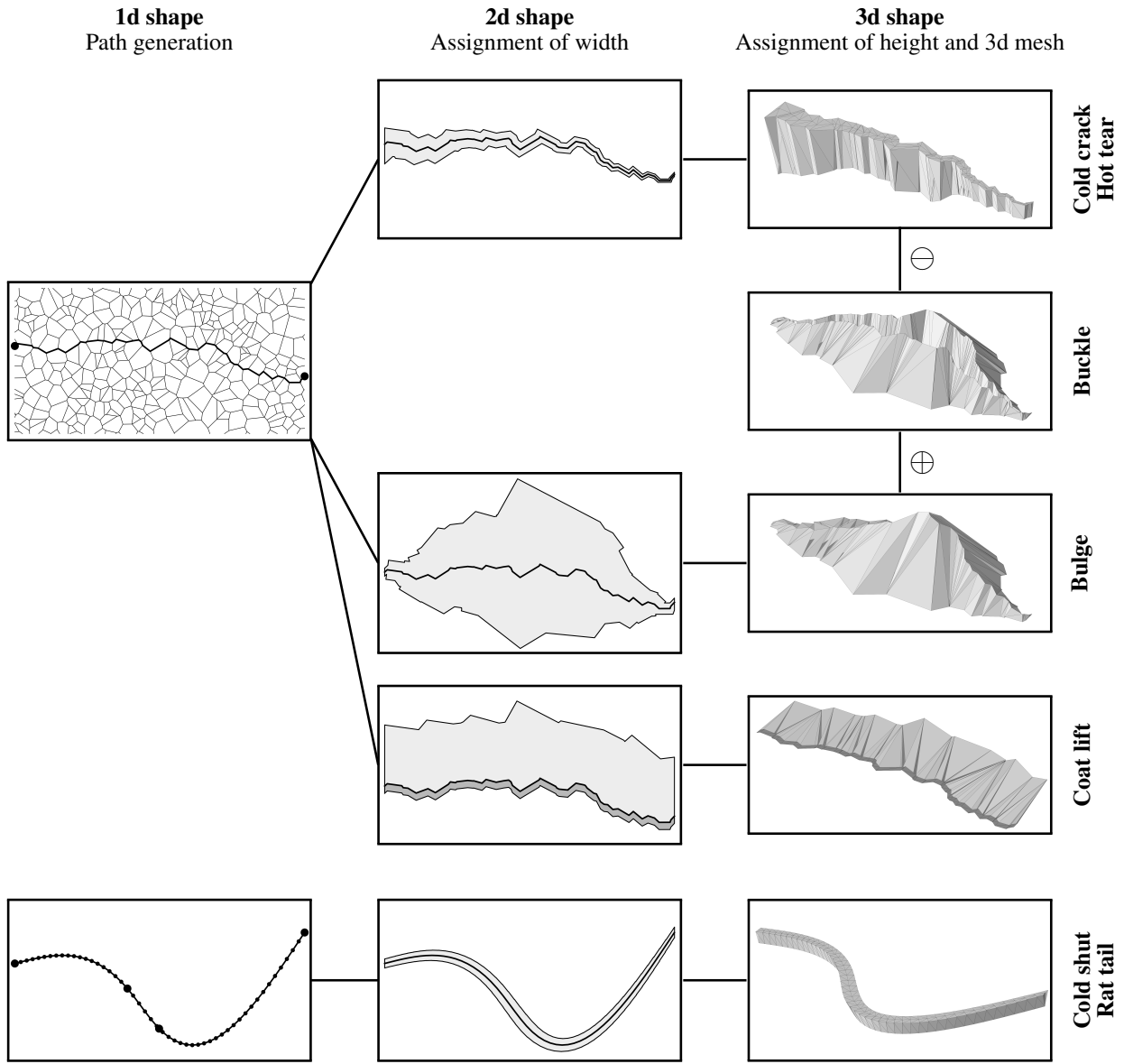

    All defect types, except for the scabs, are characterized by a dominant elongated structure.
    The behavior of that predominant line can be different, e.g. smooth for cold shuts and rat tails and jagged for the others, and the defects are indented into or added to the product geometry. 
    Nevertheless, one basic model can be used and adapted to generate all the individual defect shapes.
    The approach is based on Jung and Redenbach's \cite{Jung2022CrackModel} model to create semi-synthetic cracks in 3d \textmu CT images of concrete.
    In this approach, minimum-weight surfaces computed from a 3d Voronoi tessellation are discretized, dilated and then embedded into real \textmu CT images.
    For applications in surface inspection, the focus is on modeling the planar (2d) shape of a defect. 
    The third dimension is included by locally assigning height values to the positions of the 2d shape.
    Our basic defect model thus subdivides into the following steps, see Figure~\ref{fig:LinDefectsOverview}.
    \begin{enumerate}
		\item Generate a path.
		\item Dilate the path using segment-wise widths.
		\item Assign height values to the vertices and compute a triangulation such that a valid mesh of a 2d surface is created.
		\item Compute the 3d mesh of the defect geometry.
    \end{enumerate}

    The first step consists of generating the defect-specific 1d shape which can either be a smooth or a jagged path, obtained by a piecewise linear approximation of a smooth curve or a path through the edge system of a 2d Voronoi tessellation, respectively.
    Then, that path is dilated to obtain the 2d shape of the defect. 
    We use segment-wise widths to better control the spatial extension of the defects.
    Next, height values are assigned to the vertices.
    Finally, the triangulation of the 3d vertices is computed to obtain a valid 3d mesh of the defect geometry. 
    Figure \ref{fig:LinDefectsOverview} shows an overview of all adaptations and relations between the defect models. 
    A detailed description of the individual steps and their defect-specific adaptations is given below.
    Moreover, Figure \ref{fig:linDefRend} shows synthetic images of the different defect types rendered with parameters of a realistic imaging setup. 

    \begin{figure}
		\centering
        \begin{subfigure}[t]{0.195\textwidth}
			\includegraphics[width=\textwidth]{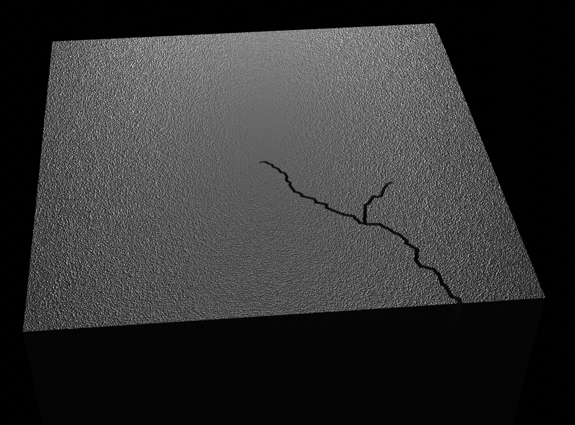}\\[2pt]
			\includegraphics[width=\textwidth]{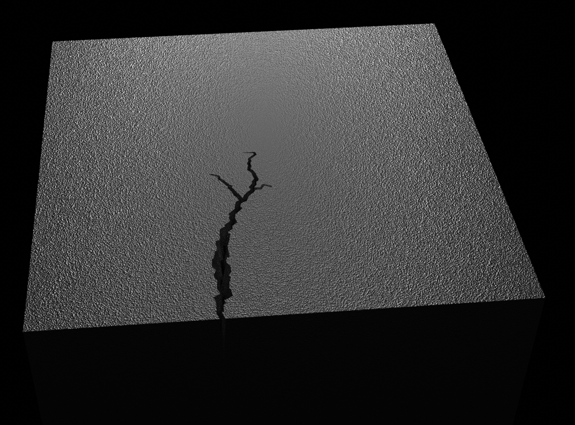}\\[2pt]
			\includegraphics[width=\textwidth]{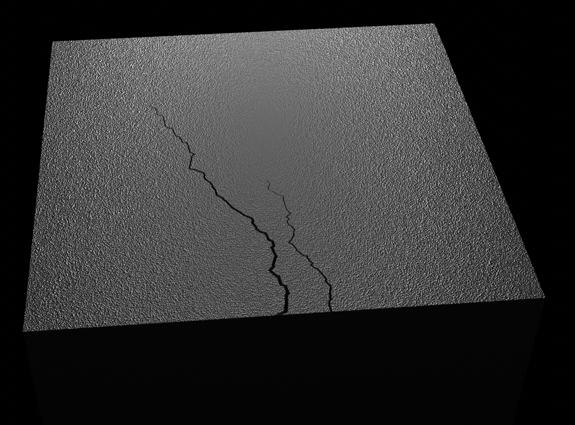}
			\caption*{\begin{tabular}{c}Cold crack\\Hot tear \end{tabular}}
		\end{subfigure}\hfill
		\begin{subfigure}[t]{0.195\textwidth}
			\includegraphics[width=\textwidth]{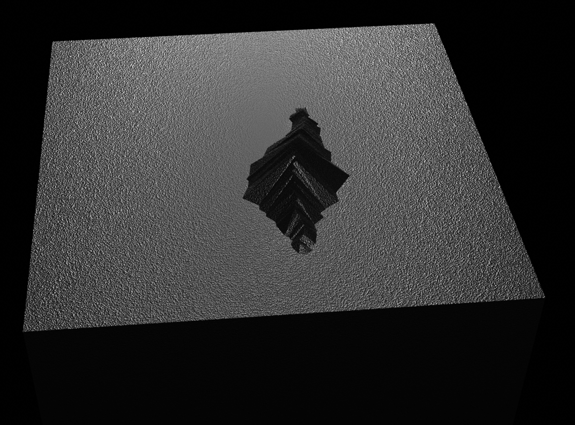}\\[2pt]
			\includegraphics[width=\textwidth]{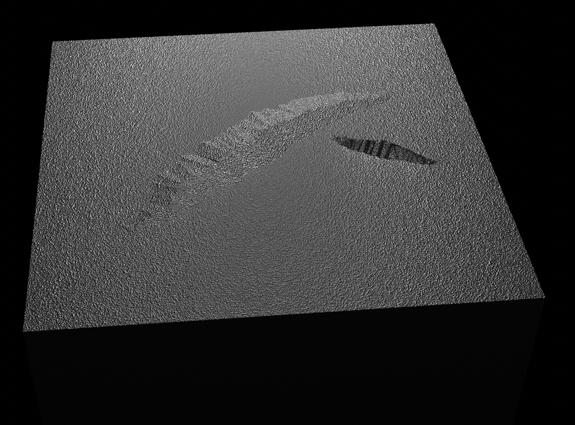}\\[2pt]
			\includegraphics[width=\textwidth]{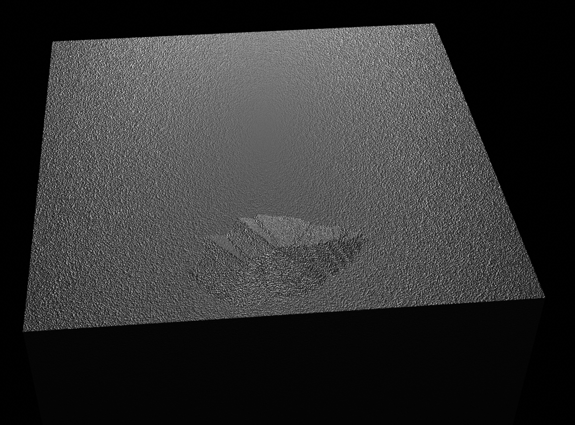}
			\caption*{Bulge}
		\end{subfigure}\hfill
		\begin{subfigure}[t]{0.195\textwidth}
			\includegraphics[width=\textwidth]{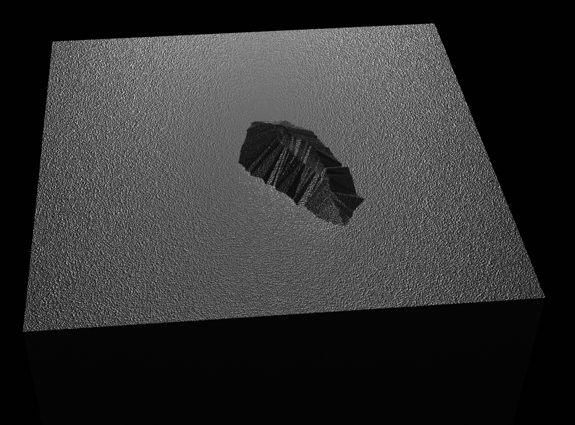}\\[2pt]
			\includegraphics[width=\textwidth]{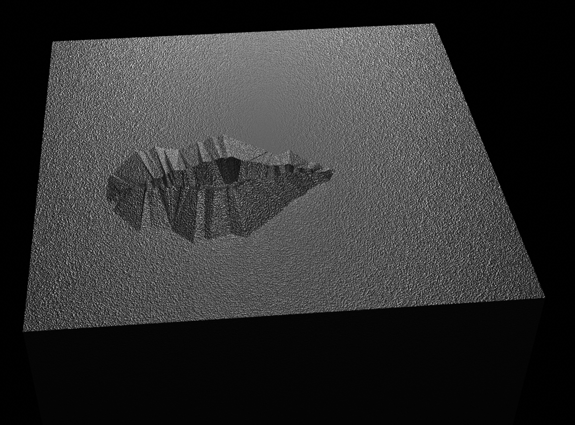}\\[2pt]
			\includegraphics[width=\textwidth]{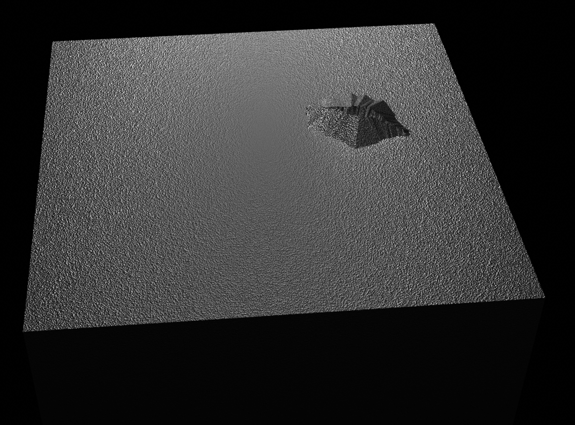}
			\caption*{Buckle}
		\end{subfigure}\hfill
		\begin{subfigure}[t]{0.195\textwidth}
			\includegraphics[width=\textwidth]{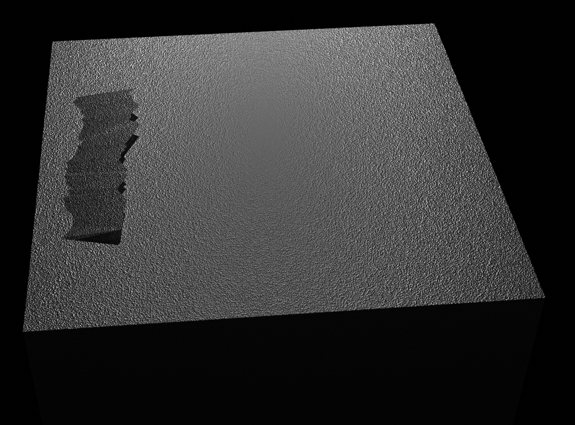}\\[2pt]
			\includegraphics[width=\textwidth]{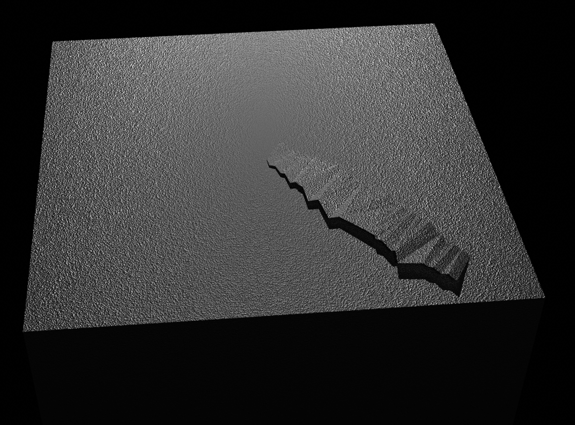}\\[2pt]
			\includegraphics[width=\textwidth]{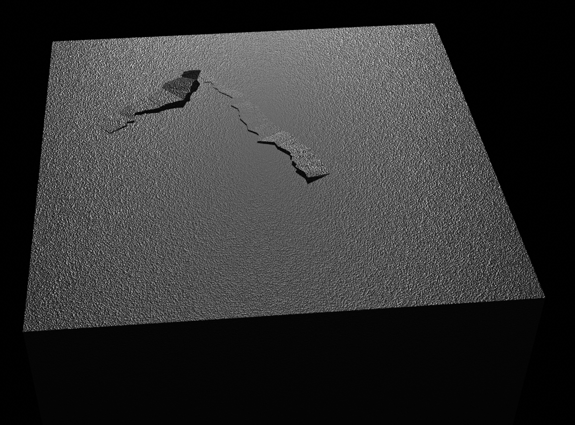}
			\caption*{Coat lift}
		\end{subfigure}\hfill
		\begin{subfigure}[t]{0.195\textwidth}
			\includegraphics[width=\textwidth]{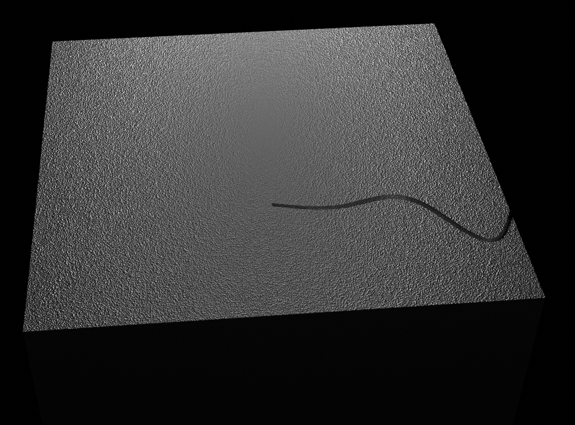}\\[2pt]
			\includegraphics[width=\textwidth]{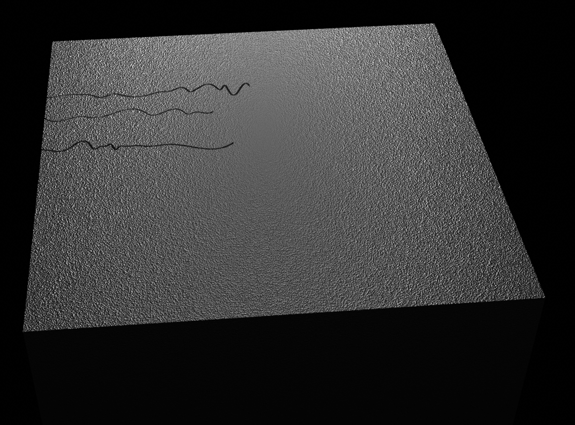}\\[2pt]
			\includegraphics[width=\textwidth]{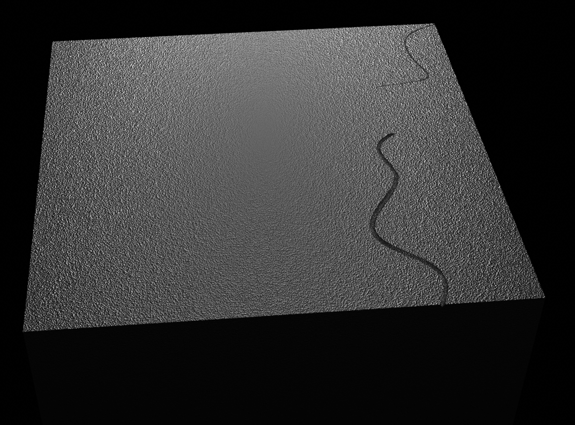}
			\caption*{\begin{tabular}{c}Cold shut\\Rat tail \end{tabular}}
		\end{subfigure}
		\caption{Synthetic images of different defect types in a cast metal cube obtained by rendering from a 
        synthetic industrial image acquisition system created in Blender \cite{blender}.
        A pinhole camera and two plane lights are used for illumination. Rendering is performed using 24 samples per pixel and 16 light bounces. The surface texture is chosen to resemble the appearance of a cast metal object. The texture model was integrated by Lovro Bosnar from Fraunhofer ITWM and is based on the cellular texture introduced by Worley \cite{castingWorley}.}
		\label{fig:linDefRend}
	\end{figure}

    \subsection{Details of individual modeling steps}  
    \subsubsection{Computation of path} \label{sec:path}
    To obtain the jagged path, two steps are needed.
    First, we generate a 2d Voronoi tessellation followed by computing a minimal path through its edge system.
    Since the path computation is performed exclusively in 2d, we restrict the following introduction to $\mathbb{R}^2$.

    \paragraph{Voronoi tessellation} 
    In general, a tessellation subdivides space into disjoint cells.
    Voronoi tessellations are generated by a set $\mathscr{P}=\left\{p_1,\dots,p_n\right\}$, $2\leq n<\infty$ of distinct points $p_i\in\mathbb{R}^2$, the generator points.
    The Voronoi cells are defined by $$C_i=\left\{x\in\mathbb{R}^2:\Vert x-p_i\Vert\leq\Vert x-p_j\Vert \text{ for all }\,j\in\left\{1,\dots,n\right\}\right\}.$$
    Thus, a point is assigned to the cell whose generator is closest with respect to the Euclidean distance $\Vert\cdot\Vert$. 
    The Voronoi tessellation $\text{Vor}(\mathscr{P})$ is the collection of all Voronoi cells of the points of $\mathscr{P}$. The vertices and edges form lower-dimensional sub-structures of the tessellation.
    We consider a bounded region $W\subseteq\mathbb{R}^2$ and use the bounded Voronoi tessellation $\left\{C_1\cap W,\dots,C_n\cap W\right\}$.
    
    \paragraph{Minimal path} 
    Let $G=(V\times A)$ be a directed graph with $V =\{v_i, i=1,\dots,n \}$ the set of vertices and $A\subseteq V\times V$ the set of directed arcs.
    The start vertex of an arc $a$ is denoted by $\alpha$ and its end vertex by $\omega$.
    A path $P$ in $G$ is defined as finite sequence of arcs $P=(a_0,a_1,\dots,a_{k})$, $k\geq 0$ with $\omega_i=\alpha_{i+1}$ for $i=0,\dots,k-1$. 
    Let $c:A\rightarrow\mathbb{R}_{>0}$ be a weight function that assigns each arc a positive value.
    The weight of a path is then defined by $c(P)=\sum_{i=0}^k c(a_i)$.
    Given two vertices $v,v^\prime\in V$, the shortest path with $\alpha_0=v$ and $\omega_k=v^\prime$ is obtained by minimizing the weight of all paths connecting $v$ and $v^\prime$.
    Famous algorithms for solving the shortest-path problem are Dijkstra's \cite{Dijkstra}, Bellman and Ford's \cite{BellmanFord} or the A* search algorithm \cite{Astar}.
    
    \paragraph{Path generation for defect modeling} 
    For our application, we need shortest paths computed in a 2d Voronoi tessellation.
    Therefore, the vertices and edges of the tessellation serve as vertices and arcs to define a graph $G$. Note that for every edge in the Voronoi tessellation we introduce two directed arcs in opposite directions in the planar graph.
    The weights are given by the lengths of the edges.
    The shortest path is computed using Dijkstra’s algorithm \cite{Dijkstra}.
    
    We use a bounded Voronoi tessellation with i.i.d generator points $p_i\sim\mathcal{U}\left(W\right)$ that follow a uniform distribution on a rectangular window $W=\left[w_0^\text{min},w_0^\text{max}\right]\times\left[w_1^\text{min},w_1^\text{max}\right]$.
    As start and end vertices of the shortest path we choose Voronoi vertices located on the left and right side of the window, that is,  $(\alpha_0)_0=w_0^\text{min}$ and $(\omega_k)_0=w_0^\text{max}$.
    See Figure \ref{fig:LinDefectsOverview} for an example.

    \subsubsection{Dilation of a path}
\begin{figure}
        \centering
        \begin{subfigure}{0.48\textwidth}
            \centering
            \includegraphics[width=\textwidth]{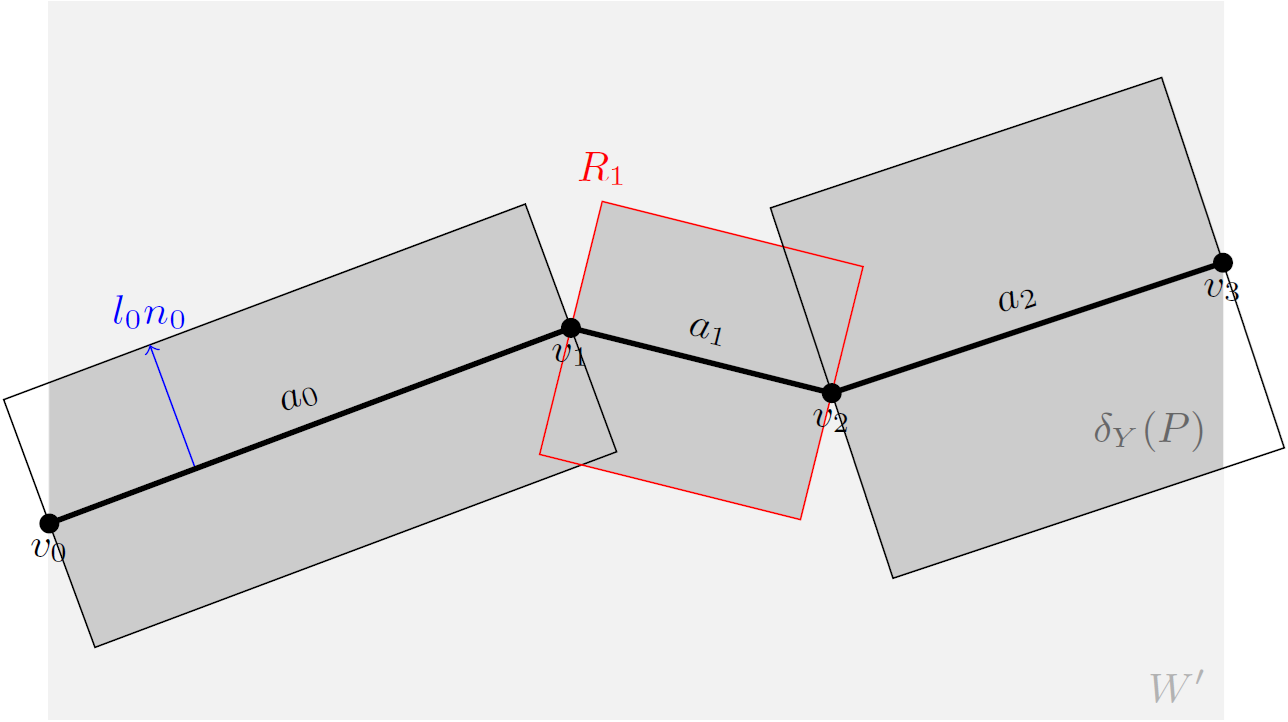}
            \caption{Primitive dilation of a path.\\ \textcolor{white}{we need much more text for space}}
            \label{fig:tikzDil:easy}
        \end{subfigure}
        \hspace{10pt}
        \begin{subfigure}{0.48\textwidth}
            \centering
            \includegraphics[width=\textwidth]{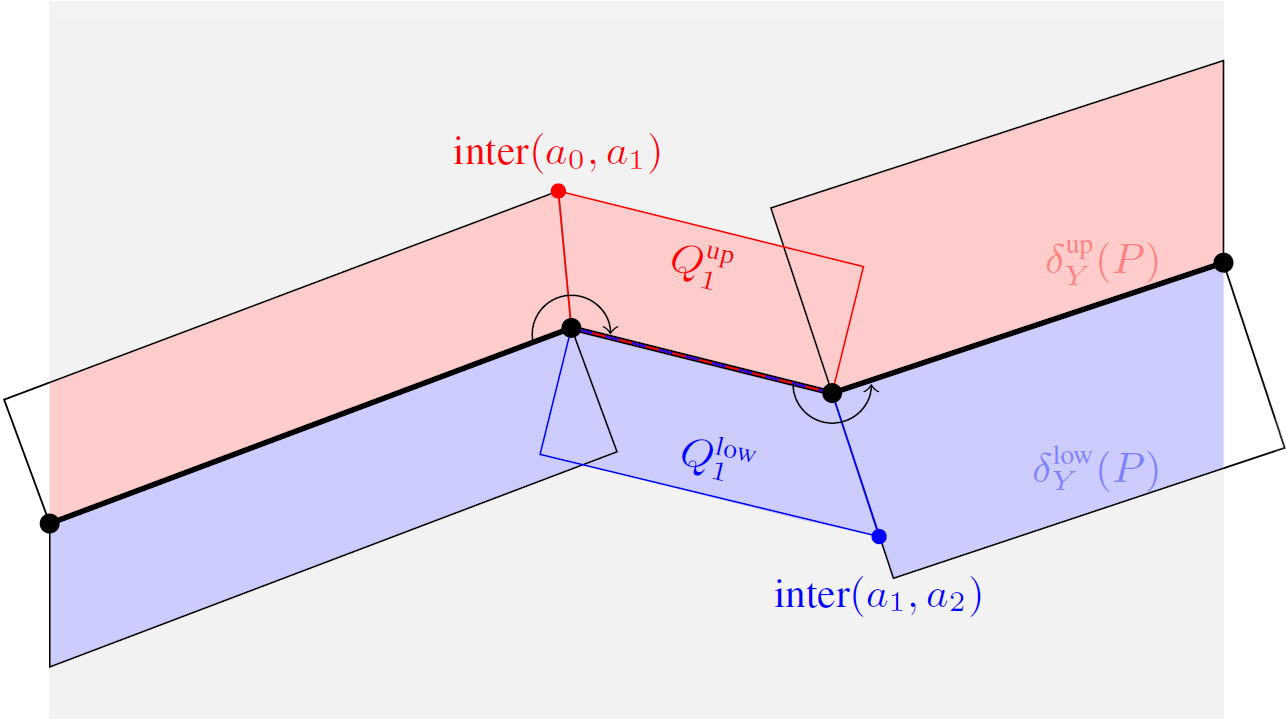}
            \caption{Adjustment such that no cone-shaped gaps appear for reflex angles.}
            \label{fig:tikzDil:advanced}
        \end{subfigure}
        \caption{Illustrations of different definitions for a dilated path.}
        \label{fig:tikzDil}
    \end{figure}

    In general, the dilation of $X\subset\mathbb{R}^2$ by a structuring element $Y\subset\mathbb{R}^2$ is defined as
    \begin{equation}
        \Delta_Y(X)=\bigcup_{y\in Y}(X-y).
        \label{eq:dilation1}
    \end{equation}
    Now, we want to dilate a path $X=P$ as given in Section \ref{sec:path}.
    Therefore, we adapt definition \eqref{eq:dilation1} to an arc-wise dilation with structuring elements $Y=\left\{Y_0,\dots,Y_{k}\right\}$, 
    \begin{equation}
        \delta_Y(P)=\left(\bigcup_{i=0}^{k}\Delta_{Y_i}(a_i)\right)\cap W^\prime.
        \label{eq:dilation2}
    \end{equation} 
    Here, we choose $W^\prime=\left[w_0^\text{min},w_0^\text{max}\right]\times\mathbb{R}$ to restrict the dilation on the length of window W.
    As structuring elements we choose line segments $$Y_i=\left\{  t \, l_i\, n_i:t\in[-1,1]\right\}$$ defined by the upwards-pointing unit normal vectors $n_i$ of the segments and positive half lengths $l_i\in\mathbb{R}_{>0}$.  
    Thus, $\Delta_{Y_i}(a_i)$ is a rectangle $R_i$ with vertices $\alpha_i\pm l_in_i,\omega_i\pm l_in_i$, see Figure \ref{fig:tikzDil:easy} for an illustration. 
    Therefore, equation \eqref{eq:dilation2} can be reformulated as
    \begin{equation*}
        \delta_Y(P)=\left(\bigcup_{i=0}^{k}R_i\right)\cap W^\prime.
    \end{equation*}
    The dilation's contours are then given as an upper path $P^\text{up}$ and a lower path $P^\text{low}$.
    In analogy, we divide the dilation of $P$ into an upper and a lower dilation $\delta^\text{up}_Y(P)$ and $\delta^\text{low}_Y(P)$, respectively. They are defined by a separation of the rectangles into an upper rectangle $R_i^{\text{up}}$ and a lower rectangle $R_i^{\text{low}}$ with the vertices 
    \begin{align*}
        R_i^\text{up} &: \alpha_i,\alpha_i+l_in_i,\omega_i,\omega_i+l_in_i\\
        R_i^\text{low} &: \alpha_i,\alpha_i-l_in_i,\omega_i,\omega_i-l_in_i.
    \end{align*}

    Now consider the angle $\sphericalangle\left(a_i,a_{i+1}\right)$ between consecutive arcs. For the upper dilation, this angle is computed in clockwise direction and for the lower dilation in counterclockwise direction. In case of a reflex angle $\sphericalangle\left(a_i,a_{i+1}\right)\in(\pi,2\pi)$, the union of the corresponding rectangles has a cone-shaped gap.
    We adapt the corresponding rectangles to fill the gaps, see Figure \ref{fig:tikzDil:advanced}.
    
    It is sufficient to explain the procedure for the upper dilation, since the approach can directly be transferred to the lower dilation.
    Compute the intersection 
    \begin{equation*}
        \text{inter}(a_i,a_{i+1})=\left\{\omega_i+l_in_i+\lambda a_i : \lambda>0\right\} \cap \left\{\alpha_{i+1}+l_{i+1}n_{i+1}-\tau  a_{i+1}:\tau>0\right\}.
    \end{equation*}

    If $l_i \neq l_{i+1}$, this intersection can be empty.
    In such a case, elongate the edge having the smaller scaling factor until it intersects the nearest lateral side of the rectangle corresponding to the other edge. For simplicity, the intersection is also denoted by $\text{inter}(a_i,a_{i+1})$.  
    Now, rectangle $R_i^\text{up}$ is replaced by a quadrangle $Q_i^\text{up}$ with vertices $\alpha_i,\omega_i,\tilde{v}_i,\tilde{v}_{i+1}$ with
    \begin{align*}
        \tilde{v}_i&=
        \begin{cases*}
            \alpha_i+l_in_i & if $\sphericalangle\left(a_{i-1},a_i\right)\in(0,\pi]$\\
            \text{inter}(a_{i-1},a_i) & else,
        \end{cases*}\\
        \tilde{v}_{i+1}&=
        \begin{cases*}
            \omega_i+l_in_i & if $\sphericalangle\left(a_i,a_{i+1}\right)\in(0,\pi]$\\
            \text{inter}(a_i,a_{i+1}) & else.
        \end{cases*}
    \end{align*}
          
    Furthermore, note that cone-shaped gaps can occur at the start and end of the dilated path.
    In that case, the upper edge is extended until it hits the boundary of the window, see Figure \ref{fig:tikzDil:advanced}. 
    Finally, the dilation of a path is given by
    \begin{equation*}
        \tilde{\delta}_Y(P) = \bigcup_{i=0}^{k} \left(Q_i^\text{up}\cup Q_i^\text{low}\right)\cap W^\prime.
    \end{equation*}
    Note that the dilation may contain holes, see for example Figure \ref{fig:tikzTri}, which will be filled when computing the triangulation.

    \subsubsection{Transition from 2d to 3d} \label{sec:2dto3d}
    The shape of the resulting defect is determined by the path $P$ and the dilation's upper and lower contours $P^\text{up}$ and $P^\text{low}$.
    To obtain three-dimensional defects, height values are assigned to the vertices that form those paths. 
    Denote the set of path vertices by $V=\left\{v_0,\dots,v_{k+1},v_0^\text{up},\dots,v_{k^\text{up}+1}^\text{up},v_0^\text{low},\dots,v_{k^\text{low}+1}^\text{low}\right\}$ 
    and by $h_i,h_{i^\text{up}}^\text{up},h_{i^\text{low}}^\text{low}$ the assigned height values.
    Assume that defects are imprinted or added to an object with a planar surface of constant height $h>0$.
    To obtain a smooth transition after imprinting, assign that height $h$ to the outermost vertices originating from the contours $P^\text{up}$ and $P^\text{low}$. 
    For the inner vertices belonging to $P$, we follow a defect-specific pipeline detailed in Section \ref{sec:defect_adaptations}.

    \subsubsection{Triangulation of 2d surface in 3d} \label{sec:Triangulation}
    In the next step, we compute a triangulation of the 3d vertices to obtain a mesh of the surface.
    In general, let $T(t_1,t_2,t_3)$ be a triangle with vertices $t_1$, $t_2$ and $t_3$.
    We consider the upper and lower dilation separately and introduce our approach for the upper dilation only. 
    Represent the contour of the upper dilation as a path $P^\text{up}=\left(a_0^\text{up},a_1^\text{up},\dots,a_{k^\text{up}}^\text{up}\right)$ with $\alpha_i^\text{up}=v_i^\text{up}$, $\omega_i^\text{up}=v_{i+1}^\text{up}$ for $i=0,\dots,k^\text{up}$, $(v_0^\text{up})_0=w_0^\text{min}$ and $(v_{k^\text{up}+1}^\text{up})_0=w_0^\text{max}$. 
    For consistency, we use the same notation for the 3d vertices including heights as for the 2d vertices.
    
    We say that $a_j^\text{up}$ originates from $a_i$ if $a_j^\text{up}$ is part of the upper contour of $Q_i^\text{up}$. 
    Note that an arc $a_i$ in $P$ does not necessarily give rise to an arc in $P^\text{up}$. 
    Furthermore, the dilation polygon $Q_i^\text{up}$ of an arc $a_i$ can generate multiple arcs of $P^\text{up}$.
    We differentiate between different cases depending on the edge of $Q_i^\text{up}$ that generates the dilation arc in $P^\text{up}$.
    First assume that the left lateral edge of $Q_i^\text{up}$ contributes at most one arc to $P^\text{up}$ and that, if it exists, this arc is connected to a dilation arc formed by the upper edge of $Q_i^\text{up}$.
    Furthermore, assume that $Q_{l}^\text{up}$ with $l>i$ does not give rise to any arcs in $P^\text{up}$ that lie further to the left than the arcs formed by $Q_i^\text{up}$.
    Then the cases are defined as follows.
    \begin{itemize}[left=1.1cm]
        \item[\textbf{Case 1:}] 
        This case refers to the leftmost (and often the only) dilation arc formed by the top edge of a dilation polygon. 
        This is the most common case.
        \item[\textbf{Case 2:}] 
        This case covers all dilation arcs formed by the top edge of a dilation polygon which are not covered in Case 1.
        Case 2 arcs only appear if a top edge forms multiple arcs.
        \item[\textbf{Case 3:}] 
        This case is defined by the arcs generated by all lateral edges of a dilation polygon.
        \item[\textbf{Case 4:}] 
        This case covers all arcs $a_i$ in $P$ where the top edge of $Q_i^\text{up}$ does not contribute an arc to $P^\text{up}$. 
    \end{itemize}
        
    Now assume that there is at least one arc $a_j^\text{up}$ formed by the left lateral edge of $Q_i^\text{up}$ that is not connected to an arc formed by the upper edge of $Q_i^\text{up}$.
    Moreover, no $Q_l^\text{up}$ with $l>i$ forms an arc that lies further to the left than $a_j^\text{up}$.  
    All such arcs formed by the lateral edges belong to Case 3. 
    All arcs generated by the upper edge of $Q_i^\text{up}$ are defined to be Case 2. 
    We extend the definition of Case 4 by the cases where it is the first time that $Q_i^\text{up}$ forms an arc in $P^\text{up}$, but not from its top edge.

    Finally, assume that at least one arc in $P^\text{up}$ is formed by $Q_i^\text{up}$ but there exists a $Q_l^\text{up}$ with $l>i$ that gives rise to arcs in $P^\text{up}$ further to the left.
    Then, all arcs formed by the lateral and top edge of $Q_i^\text{up}$ are added to Case 3 and Case 2, respectively.
    Moreover, Case 4 is applied before visiting the first arc of $Q_l^\text{up}$.
    
\begin{figure}
        \centering 
    	\includegraphics[width=0.54\textwidth]{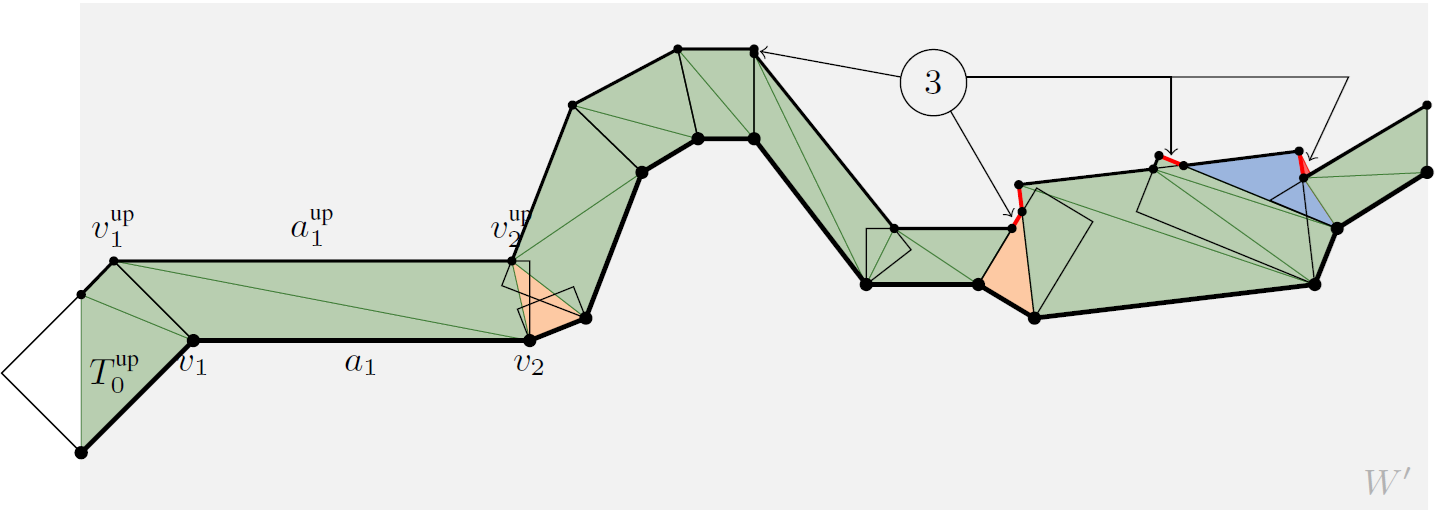}\hfill
        \includegraphics[width=0.45\textwidth]{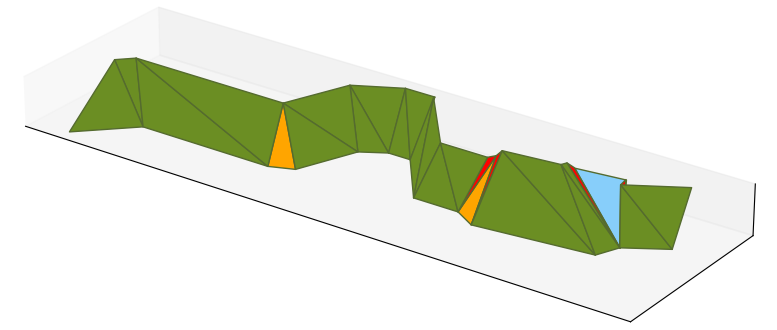}
        \caption{Illustration of triangulation with marking the four cases described in Section \ref{sec:Triangulation}: 1. green, 2. blue, 3. red, and 4. orange. 
        Left: 2d drawing as projection on the x-y-plane. Right: 3d view with the path vertices deeper than the dilated path.
        Note that in 2d the blue triangle is partially and almost all red triangles are totally covered by the subsequent triangles. 
        Therefore, the corresponding arcs of all Case 3 triangles are highlighted in the contour of the upper dilation. }
        \label{fig:tikzTri}
    \end{figure}

    The triangulation is computed by iterating over the arcs $a_j^\text{up}$ while simultaneously keeping track of the arcs $a_i$. 
    To do so, let $i_\text{last}=0$ be the last visited polygon among the $Q_i^\text{up}$ and iterate over the arcs $a_j^\text{up}, j=0,\dots,k^\text{up}$.
    An example of the algorithm including the classification into the four cases is provided in Figure \ref{fig:tikzTri}. 

    First, check from which arc $a_i$ the current arc $a_j^\text{up}$ originates.
    
    If $i>i_\text{last}$, then the polygon $Q_i^\text{up}$ is visited for the first time. In this case, proceed as follows.
    If $j=0$ and $i>0$, then $T\left(\alpha_0,\omega_0,\alpha_0^\text{up}\right)$ must be added first.
    If there are edges in $P$ located between $a_{i_\text{last}}$ and $a_i$ that do not generate edges in $P^\text{up}$ prior to $a_j^\text{up}$, then add all triangles $T\left(\alpha_m,\omega_m,\alpha_j^\text{up}\right)$ for $m=i_\text{last}+1,\dots,i-1$ (\textbf{Case 4}).
    Then, we have to differentiate whether $a_j^\text{up}$ originates from the top, left or right edge of $Q^\text{up}_i$:
    \begin{itemize}[left=1cm]
        \item[\textbf{Top:}]
        Add the triangles $T\left(\alpha_i,\omega_i,\alpha_j^\text{up}\right)$ and $T\left(\omega_i,\alpha_j^\text{up},\omega_j^\text{up}\right)$ (\textbf{Case 1}).
        Update $i_\text{last}=i$.
        
        \item[\textbf{Left:}]
        Add triangle $T\left(\alpha_i,\alpha_j^\text{up},\omega_j^\text{up}\right)$ (\textbf{Case 3}).
        If the subsequent arc $a_{j+1}^\text{up}$ is also a lateral edge of its corresponding polygon (left or right is irrelevant), further add triangle $T\left(\alpha_i,\omega_i,\omega_j^\text{up}\right)$ (\textbf{Case 4}) and update $i_\text{last}=i$. 
        Otherwise, if $a_{j+1}^\text{up}$ originates from $a_i$ as well, meaning from the top edge of $Q^\text{up}_i$, then set $i_\text{last}=i-1$ such that the triangles of Case 1 are added properly.

        \item[\textbf{Right:}] 
        If $a_{j-1}^\text{up}$ is also a lateral edge of its corresponding polygon (left or right is irrelevant), first add triangle $T\left(\alpha_i,\omega_i,\alpha_j^\text{up}\right)$ (\textbf{Case 4}).
        Then, independently of $a_{j-1}^\text{up}$, triangle $T\left(\omega_i,\alpha_j^\text{up},\omega_j^\text{up}\right)$ is added (\textbf{Case 3}).
        Update $i_\text{last}=i$.
    \end{itemize} 
    
    If $i\leq i_\text{last}$, that is, polygon $i$ was already visited, then just triangle $T\left(\omega_{i_\text{last}},\alpha_j^\text{up},\omega_j^\text{up}\right)$ is added, regardless of whether it is part of the top (\textbf{Case 2}), left or right edge (\textbf{Case 3}) of $Q^\text{up}_i$. 
    
    Finally, if $i_\text{last}<k$ after finalizing the iteration over $j$, then not all path elements of $P$ were visited.
    To complete the triangulation, add all triangles $T\left(\alpha_i,\omega_i,\omega_{k^\text{up}}^\text{up}\right)$ for $i=i_\text{last}+1,\dots,k$ (\textbf{Case 4}).

    As the 3d mesh of the defect has to be closed for proper imprinting or adding to the object surface, extra triangles and possibly also extra vertices are needed to obtain a closed surface triangulation of the defect geometry. Details are given in the description for the specific defects below.
    Note that depending on the height values and the length of the path segments, the resulting triangles can get very thin which may cause problems during imprinting the defect geometry into the object geometry.
    This can, for example, be solved by removing too short arcs from the path before applying the above algorithm.

    \subsection{Defect-specific adaptations} \label{sec:defect_adaptations}
    \subsubsection{Cold crack and hot tear} \label{sec:crack}
    Cracks are caused by internal tensile stress during the cooling and solidification of the molten metal.
    Two types of cracks are distinguished depending on the predominant temperature in the casting: hot tears occur before the solidification temperature is reached, and cold cracks afterwards.
    The volume reduction during the cooling process starts at different locations and can thus cause solid material separations.
    Hot tears appear as a result of material shrinkage during solidification of the remaining melt.
    Cold cracks emerge if the stress of the solidified material exceeds its strength.
    Thus, cold cracks are trans-crystalline which means that metal grains can be destroyed.
    In contrast, hot tears are inter-crystalline.
    
    Cracks are linear structures of irregular shape caused by taking the path of least resistance.
    Usually, cracks are deeper than wide with a symmetric depth profile when looking from the  object surface.
    However, the width and depth within a crack can vary and additional branches can occur.
    Our model is parameterized so that all those properties are captured.
    In general, the model is divided into the following steps, which are illustrated in Figure \ref{fig:CrackSteps}:
    \begin{figure}
        \centering
        \begin{subfigure}[t]{0.49\textwidth}
            \centering 
            \includegraphics[width=\textwidth]{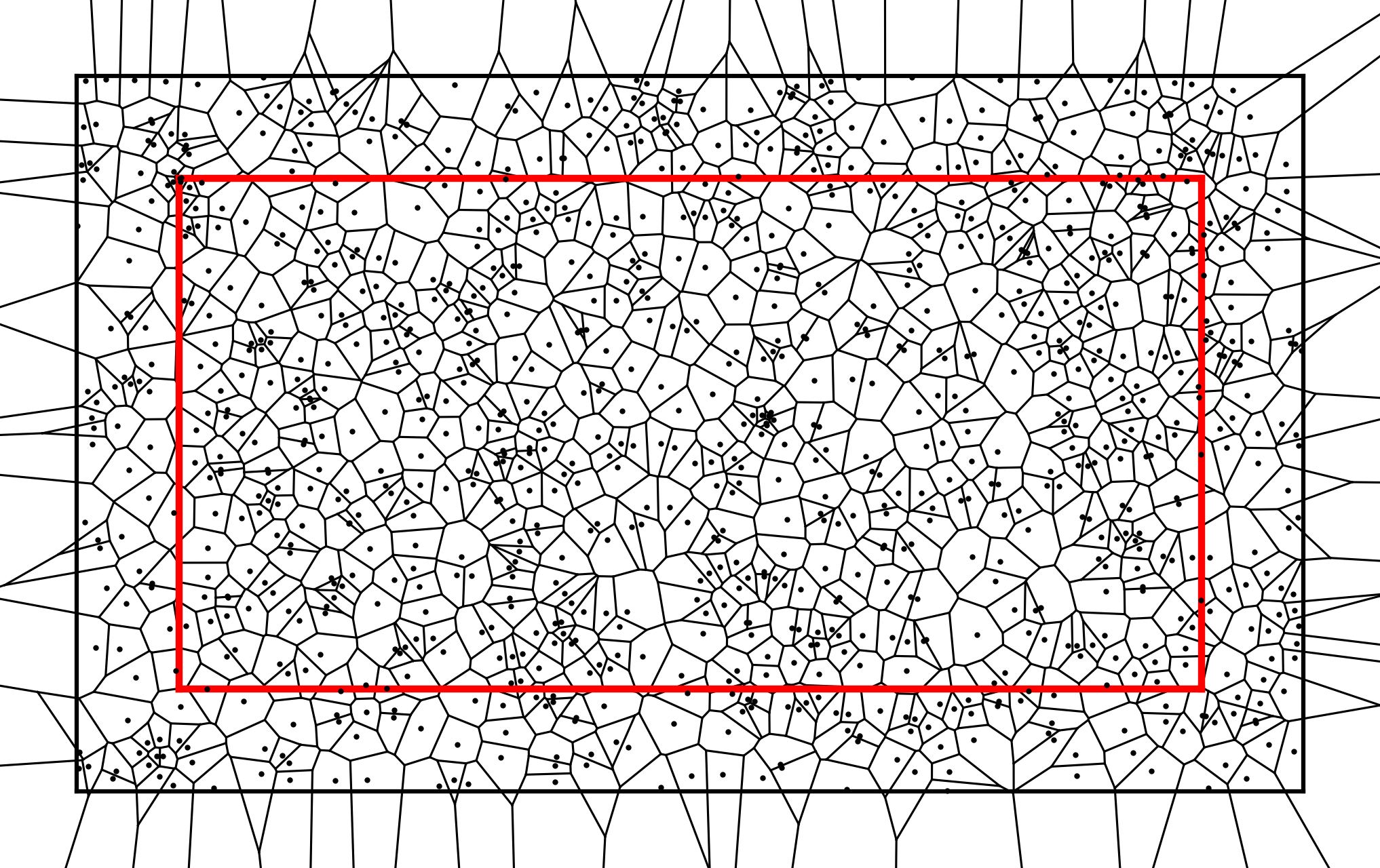}
            \caption{Voronoi tessellation with its generator points in the dilated window (black), the original window is marked in red.}
            \label{fig:CrackStepsVoro}
        \end{subfigure}
        \hfill
        \begin{subfigure}[t]{0.49\textwidth}
            \centering 
            \includegraphics[width=\textwidth]{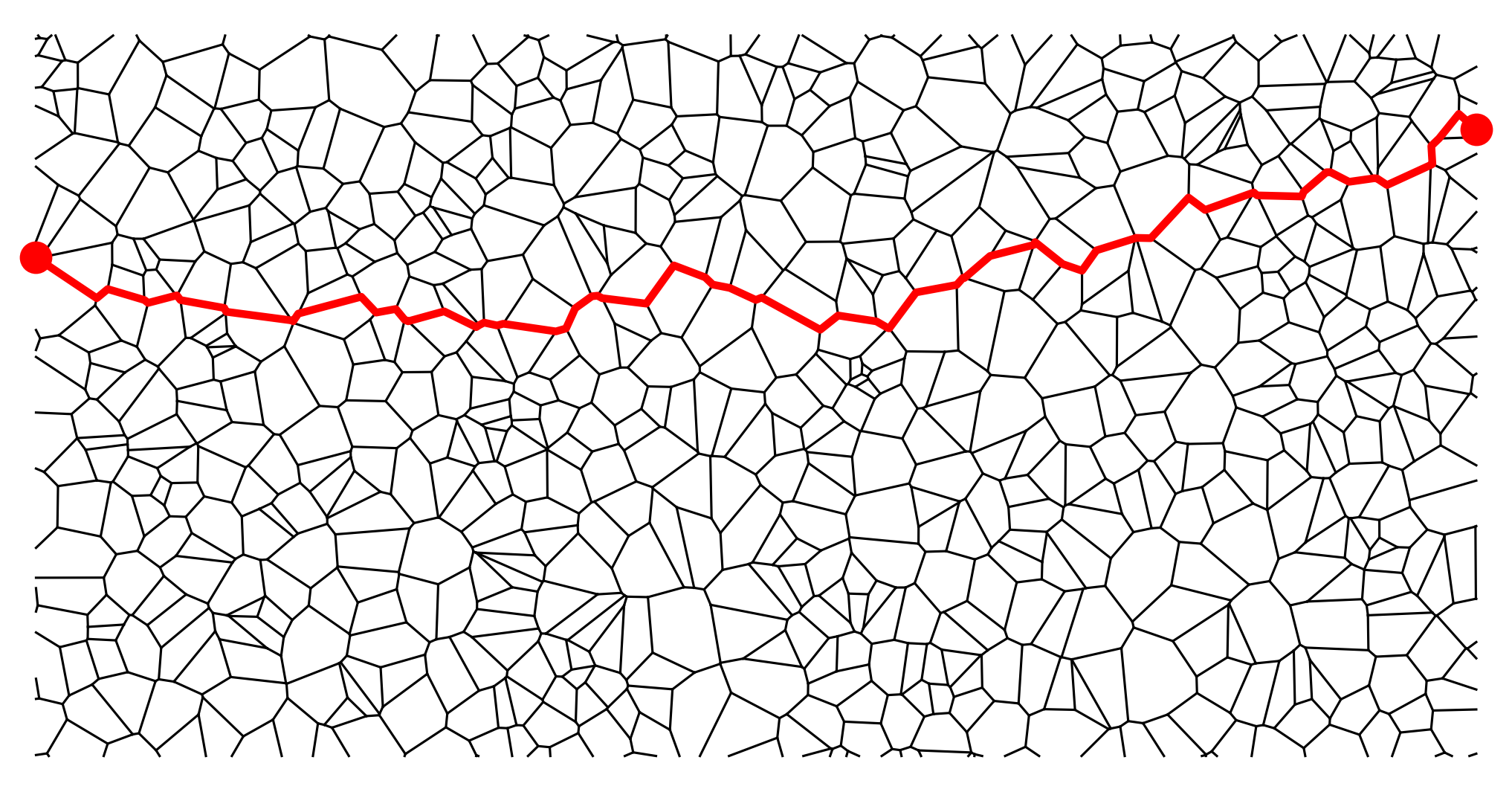}
            \caption{Minimal path (red) between the start and end vertex (circles) in the Voronoi tessellation restricted to the window.}
            \label{fig:CrackStepsPath}
        \end{subfigure}\\
        \begin{subfigure}[t]{0.49\textwidth}
            \centering 
            \includegraphics[width=\textwidth]{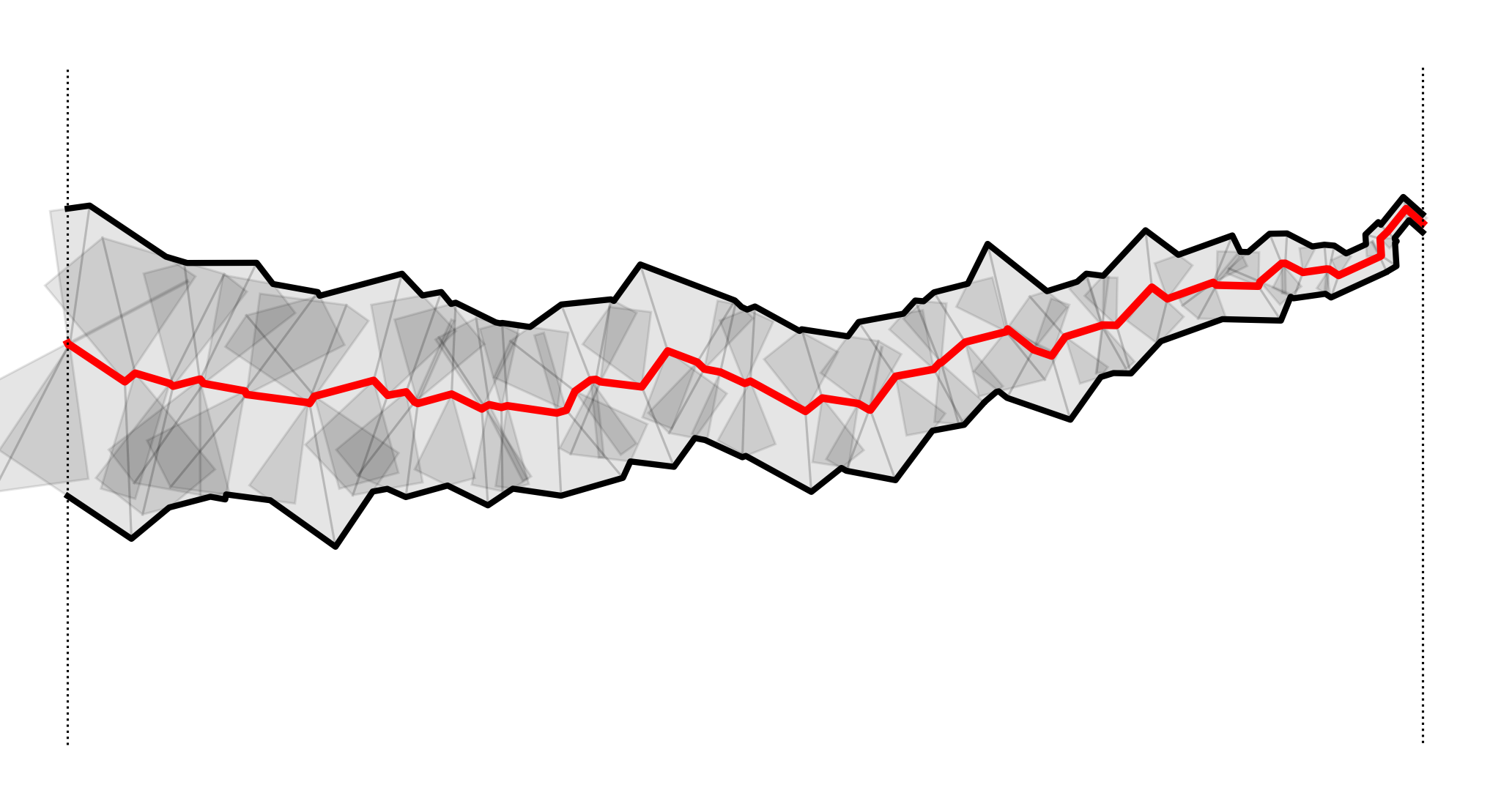}
            \caption{Dilated path (black) bounded to the window edges (dotted). Dilation polygons $Q_i$ in gray.}
            \label{fig:CrackStepsTri}
        \end{subfigure}
        \hfill
        \begin{subfigure}[t]{0.49\textwidth}
            \centering 
            \includegraphics[width=\textwidth]{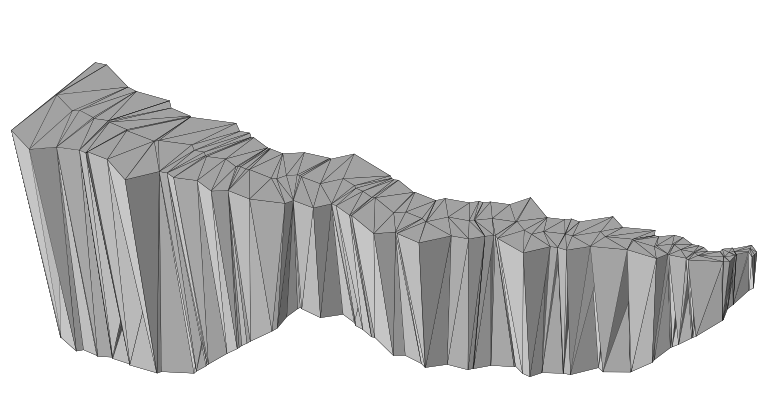}
            \caption{3d mesh of the crack defect with surface triangulation.}
            \label{fig:CrackSteps3d}
        \end{subfigure}
        \caption{Example of the crack defect generation according to the model steps using $n=1000$ generator points and $\gamma=0.1\left(w_0^\text{max}-w_0^\text{min}\right)$.
        }
        \label{fig:CrackSteps}
    \end{figure}
    
    \begin{enumerate}
        \item Choose a window $W$ such that $w_0^\text{max}-w_0^\text{min}$ equals the desired expansion of the defect.
        \item Generate $n$ random points from a uniform distribution in the dilated window $\Delta_\gamma(W)=\left[w_0^\text{min}-\gamma,w_0^\text{max}+\gamma\right]\times\left[w_1^\text{min}-\gamma,w_1^\text{max}+\gamma\right]$ with $\gamma>0$ and compute the corresponding Voronoi tessellation. The dilation is necessary to obtain regular cell shapes even at the window borders.
        \item Choose a start and an end vertex at the left and right borders of the window, respectively, and compute the associated minimal path. 
        \item 
        Dilate the path using segment-wise width values $l_i$.
        
        \item Assign height values $h_i\in\left(0,h\right]$ to the path vertices $v_i$ that define the depth of the crack. 
        \item Compute the 3d surface triangulation. 
        \item Compute the mesh $C=(V,T)$ of the 3d crack, where $V$ is the set of vertices and $T$ the set of triangles. 
        In addition to the triangles provided in Step 6, the crack needs a cover at the top and the sides to be a closed surface mesh.
        Therefore, add a triangulation $T_\text{surface}$ of $\tilde{\delta}_Y(P)$ at the level of the object surface, i.e., at height $h$, with vertices $$\left\{(v_0^\text{up},h),\dots,(v_{k^\text{up}+1}^\text{up},h),(v_0^\text{low},h),\dots,(v_{k^\text{low}+1}^\text{low},h)\right\}.$$
        Note that extra vertices may be added for easier computation.
        To close the mesh, add triangles at the path ends via $T_\text{end}$.
        Thus, the mesh is defined by
        \begin{align*}
            V&=\left\{\left(v_i,h_i\right)\right\}_{i=0,\dots,k+1}\cup\left\{\left(v_i^\text{up},h\right)\right\}_{i=0,\dots,k^\text{up}+1}\cup\left\{\left(v_i^\text{low},h\right)\right\}_{i=0,\dots,k^\text{low}+1},\\
            T&=T_\text{Step 6}\cup T_\text{surface} \cup T_\text{end}. 
        \end{align*}
    \end{enumerate}

\begin{figure}
        \centering
        \begin{subfigure}[t]{0.65\textwidth}
            \includegraphics[width=0.495\textwidth]{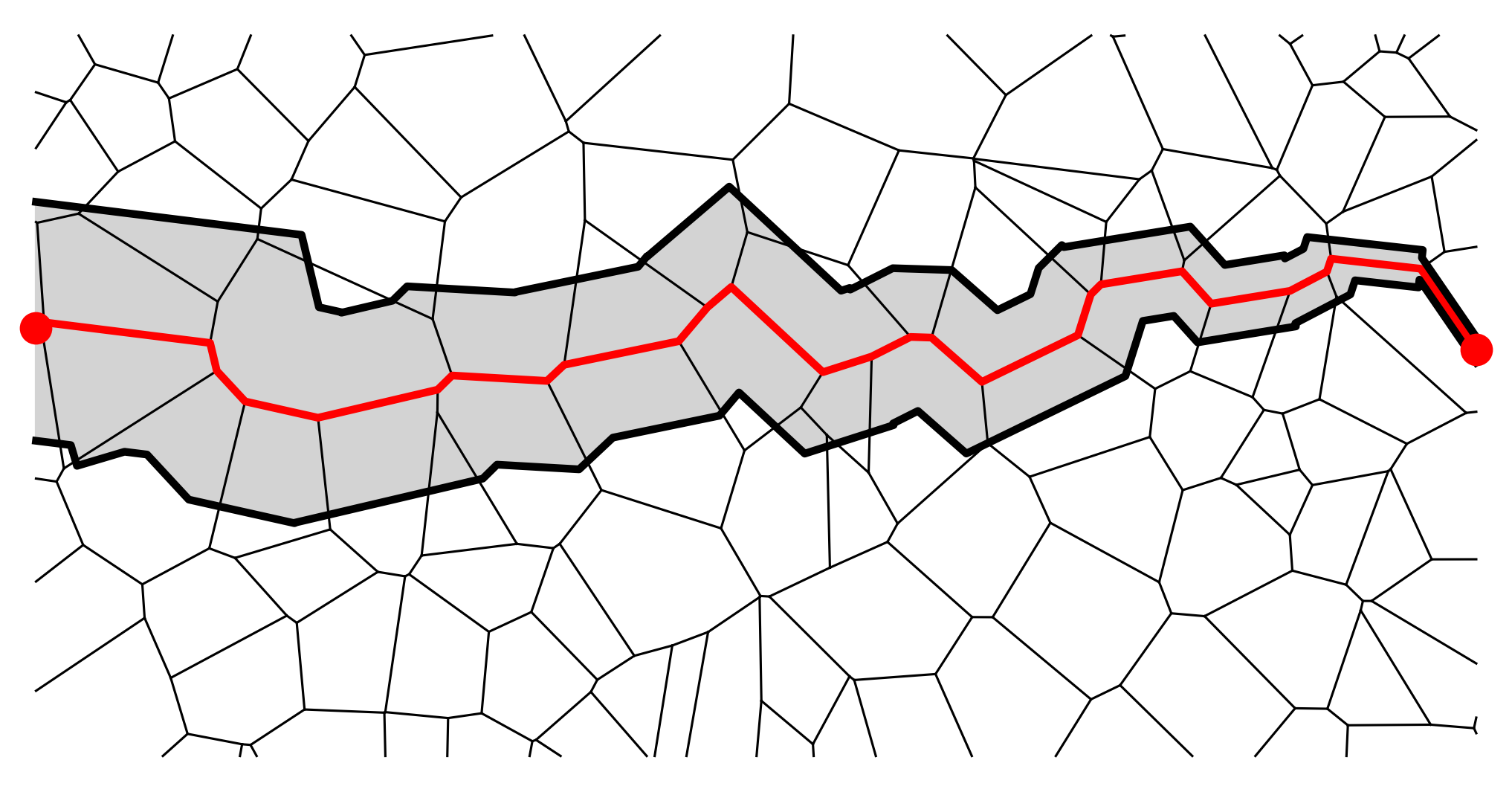}
            \hfill
            \includegraphics[width=0.495\textwidth]{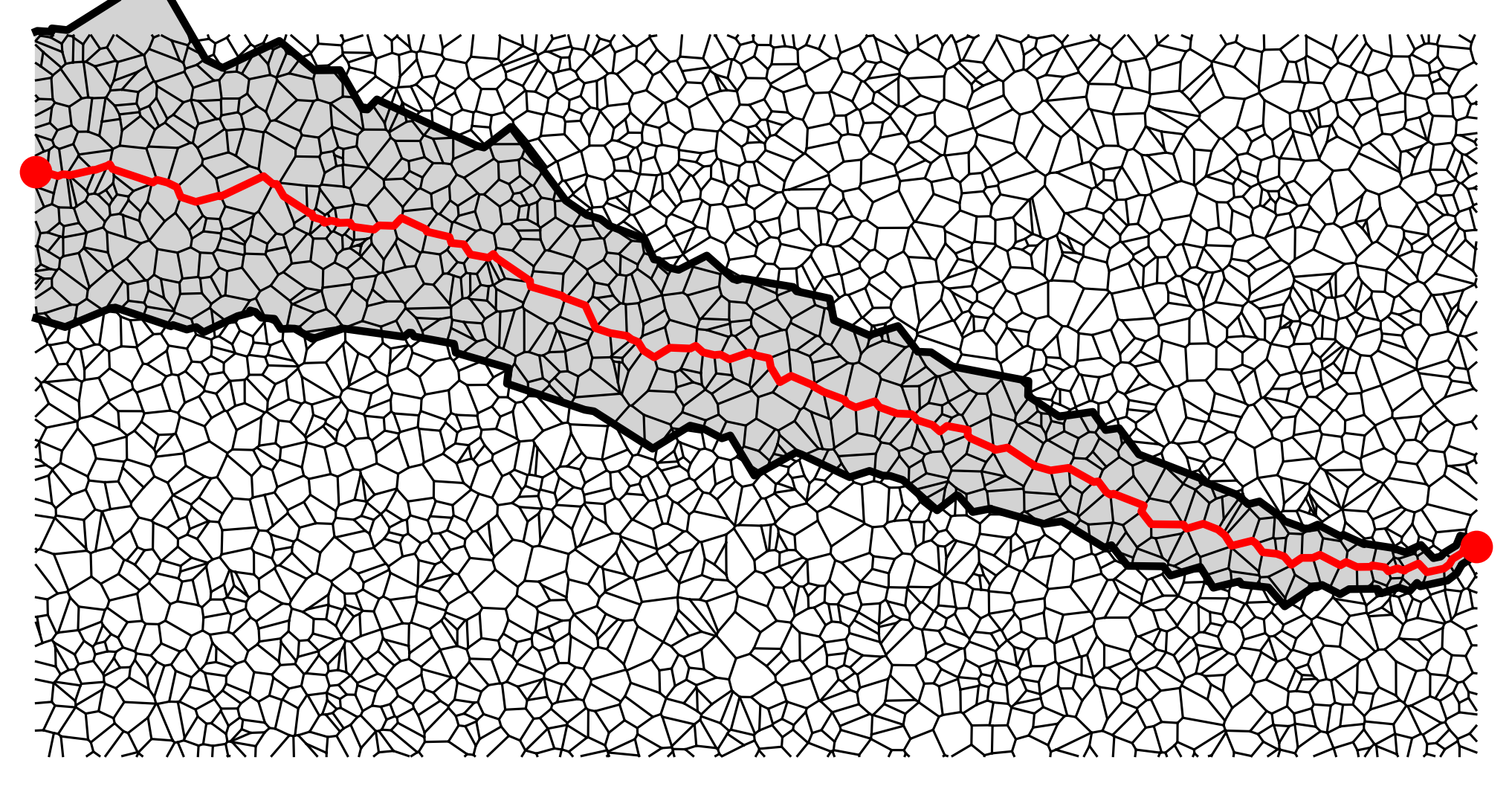}\\
            \includegraphics[width=0.495\textwidth]{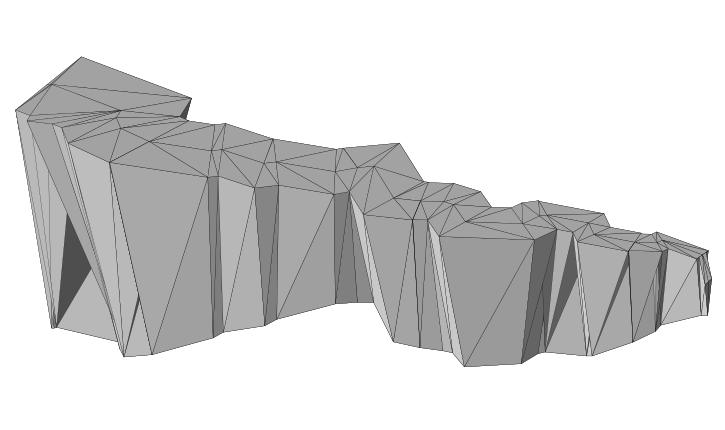}
            \hfill
            \includegraphics[width=0.495\textwidth]{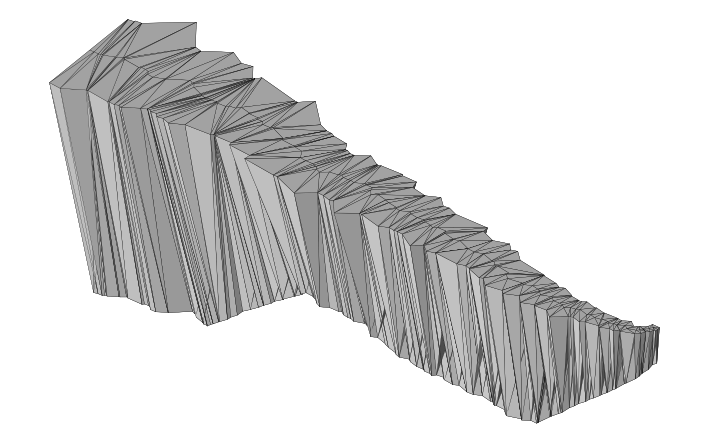}
            \caption{Crack defects using different numbers of generator points. Left: $n=150$, right: $n=2000$.}
            \label{fig:CrackPossInt}
        \end{subfigure}
        \hfill
        \begin{subfigure}[t]{0.325\textwidth}
            \includegraphics[width=0.99\textwidth]{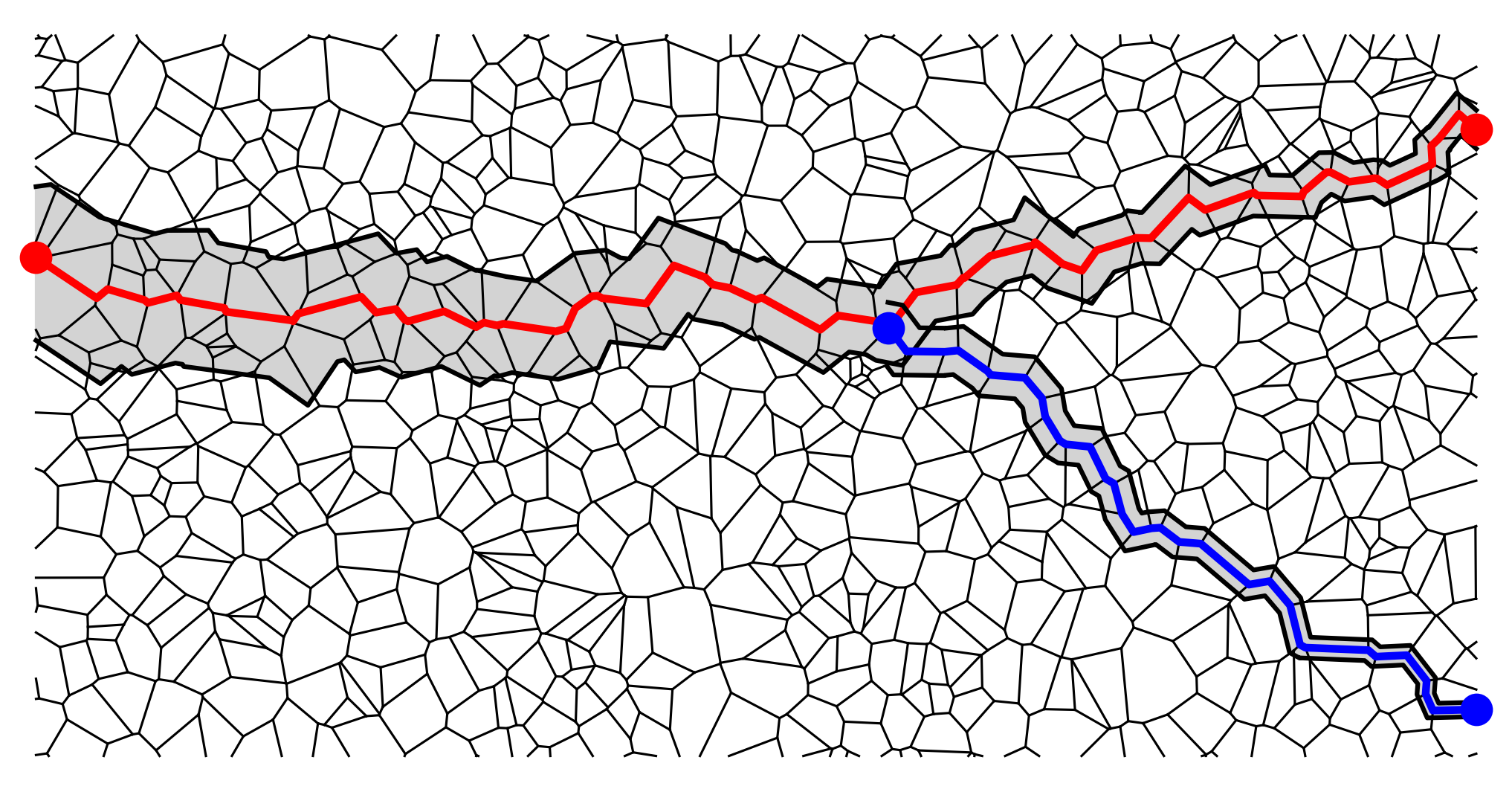}\\
            \includegraphics[width=0.99\textwidth]{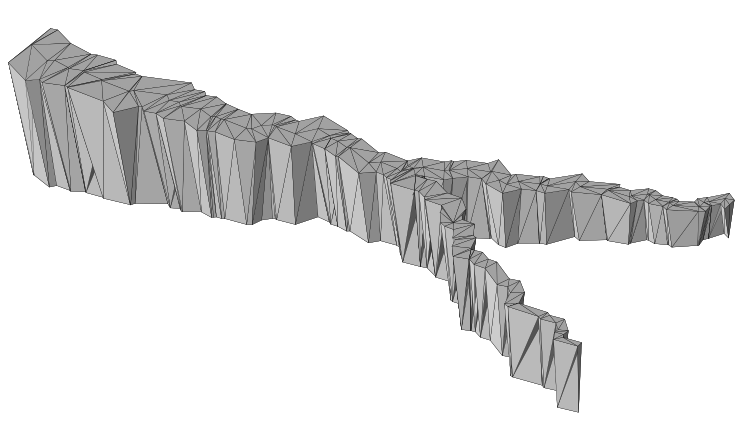}
            \caption{Branching using $n=1000$.}
            \label{fig:CrackBranch}
        \end{subfigure}
        \caption{Parameter variations of the pipeline for crack generation. Top:  Voronoi tessellation with minimal paths of the main crack (red) and possible branch (blue) with their dilations (gray). Associated start and end vertices are marked. Bottom: 3d mesh of the defect.}
        \label{fig:CrackVariation}
    \end{figure}

    In addition to the values for width and depth, the number of generator points for the Voronoi tessellation influences the shape of the crack, see Figure \ref{fig:CrackPossInt}.
    With an increasing number of generator points, the crack structure gets finer.
    However, details may be lost as the width increases.
    
    Branching is a common crack property and should thus be included in the model.
    In the following, we describe the procedure to include one branch, but it can easily be extended to an arbitrary number of branches.
    Let $C=(V_C,T_C)$ be the mesh of a crack defect.
    First, choose a random Voronoi vertex on the crack path and use it as the starting point of the branch. 
    When considering more than one branch, the start vertices may also be chosen from the previously generated branches.
    Next, choose an arbitrary Voronoi vertex outside the path as the end point of the branch. 
    Use the same Voronoi tessellation as for generating the crack and compute the minimal path.
    Then, proceed with Steps 4-7 of the crack simulation procedure to obtain a branch $B=(V_B,T_B)$.
    To connect crack and branch, the triangles of $T_C\cup T_B$ have to be re-meshed to form a valid surface triangulation of $C\cup B$, see Figure \ref{fig:CrackBranch} for an example.
    For this purpose, we use the Boolean union function of the Python package \textsc{trimesh} \cite{trimesh}. 

    \subsubsection{Bulge} \label{sec:bulge}
    Bulges are material accumulations caused by indents in the casting mold and appear as local pyramidal-shaped thickenings of the object surface.
    They resemble a reverse crack, but with larger width and smaller depth.
    Therefore, we reuse the crack model from Section \ref{sec:crack} with an adaptation in Step 6 to generate positive defects, that is $h_i > h$.
    Furthermore, the width values should be chosen so that the resulting dilation is roughly eye-shaped.
    See an example of a synthetic bulge in Figure \ref{fig:FlatDefects:bulge}.

    \subsubsection{Buckle}
    Buckling describes a local lift-off of the top layer of the object surface caused by insufficient bond strength.
    It can take different forms: linear buckling of small width and long length or pyramidal shaped lift-offs.
    Furthermore, a buckle can be closed or open.
    The latter means that the layer is torn off.
    
\begin{figure}
        \centering
        \begin{subfigure}[t]{0.32\textwidth}
            \centering 
            \includegraphics[width=\textwidth]{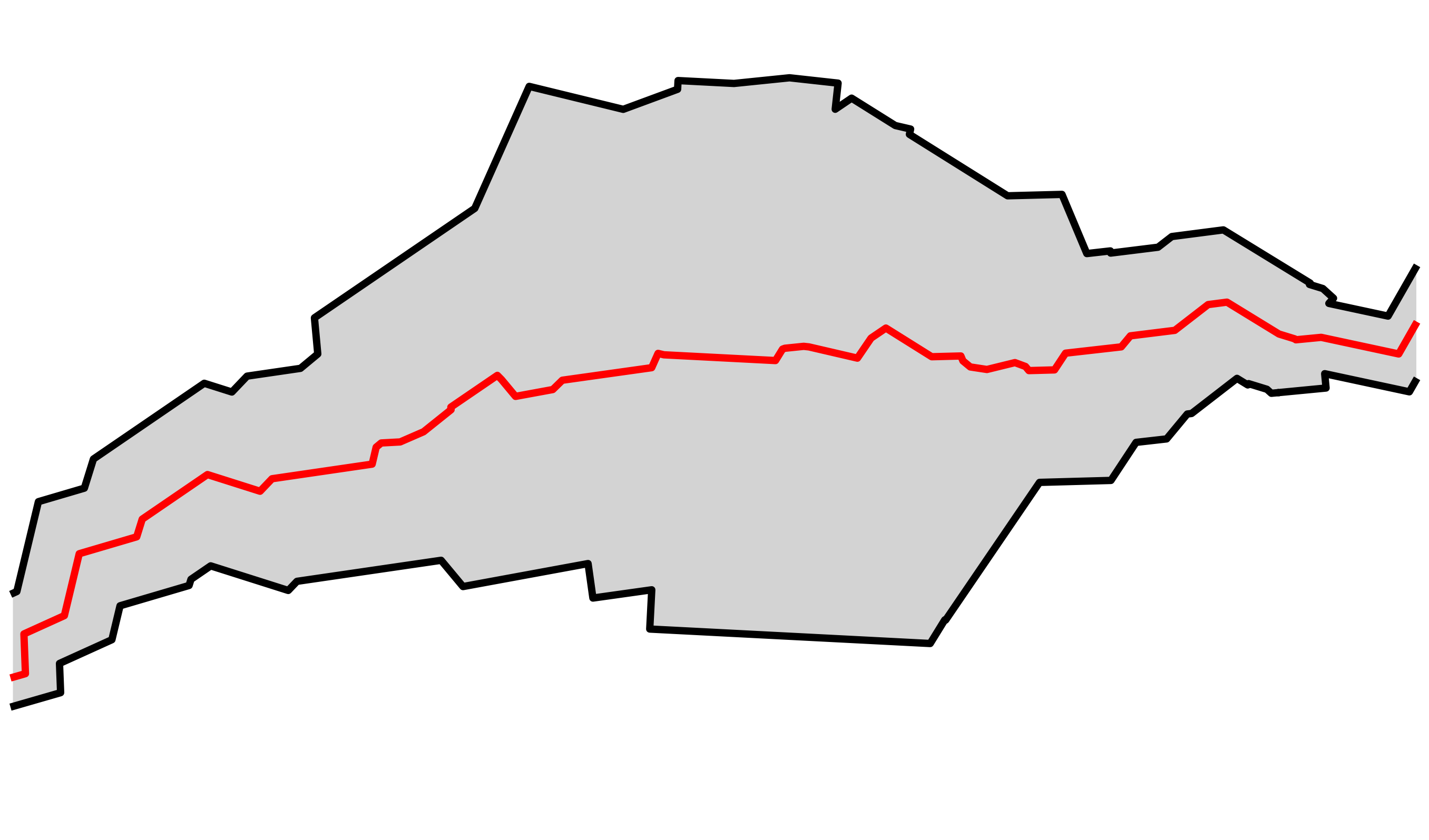}
            \includegraphics[width=\textwidth]{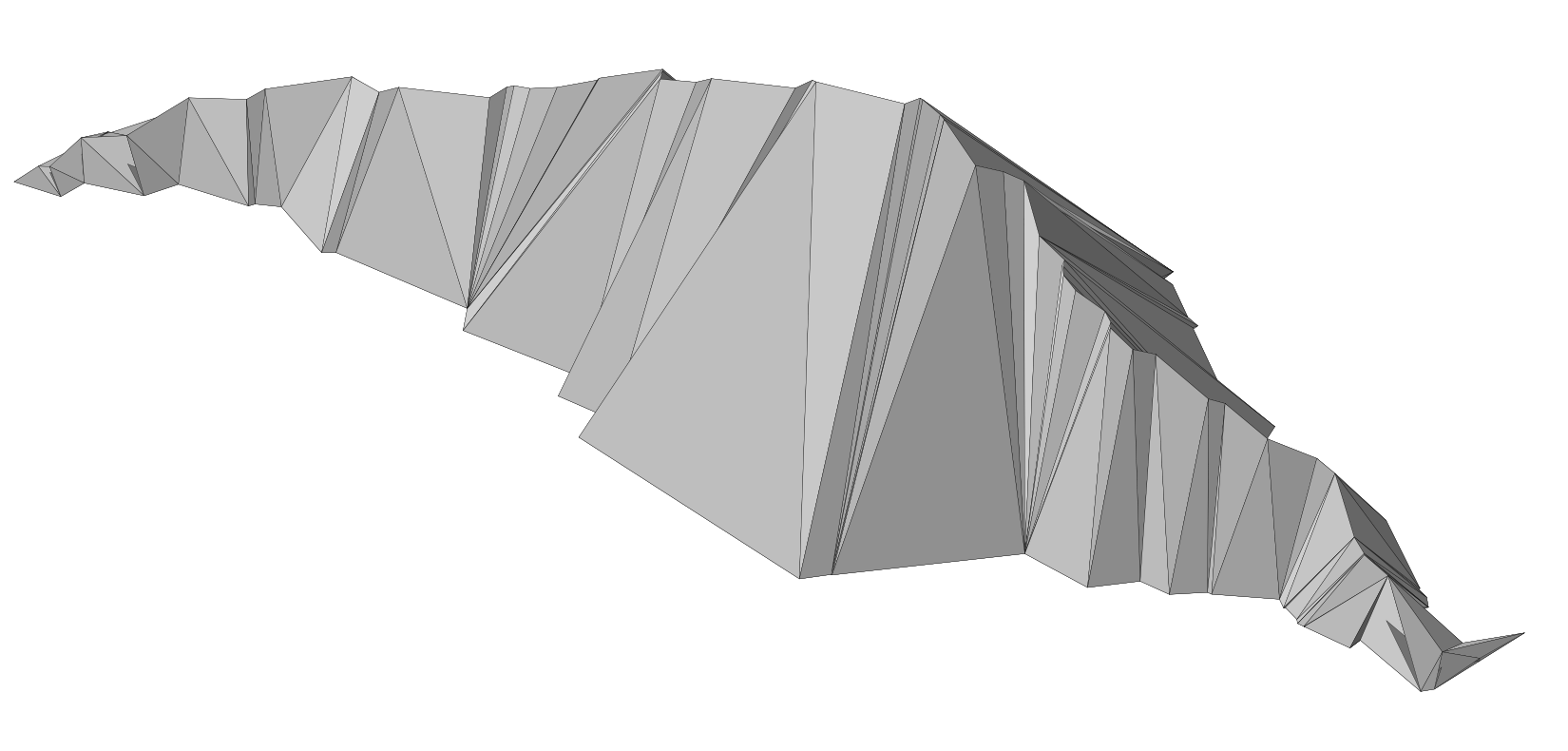}
            \caption{Bulge. Coloring of the top image: Minimal path in red and its dilation in gray.}
            \label{fig:FlatDefects:bulge}
        \end{subfigure}
        \hfill
        \begin{subfigure}[t]{0.32\textwidth}
            \centering 
            \includegraphics[width=\textwidth]{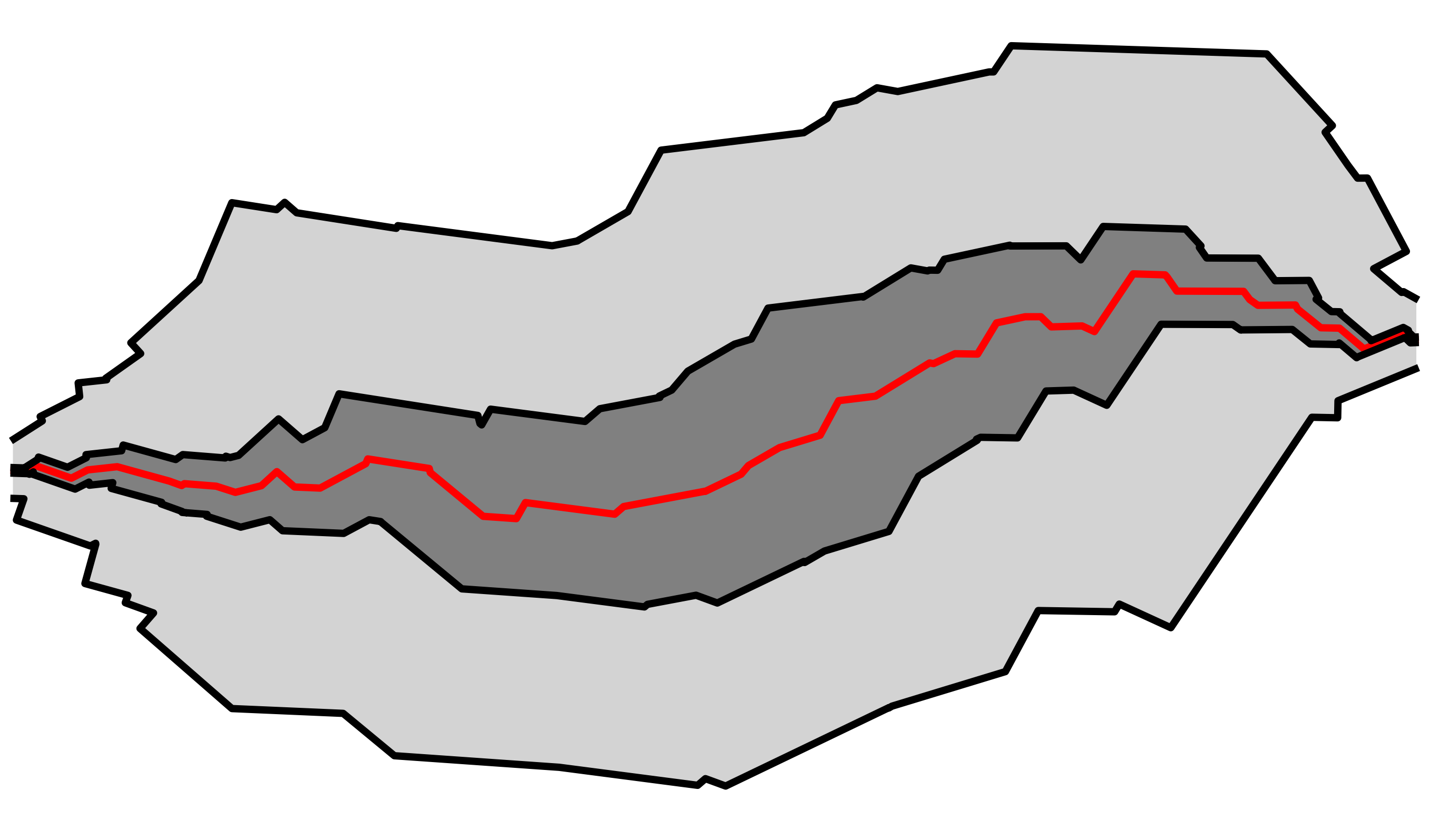}
            \includegraphics[width=\textwidth]{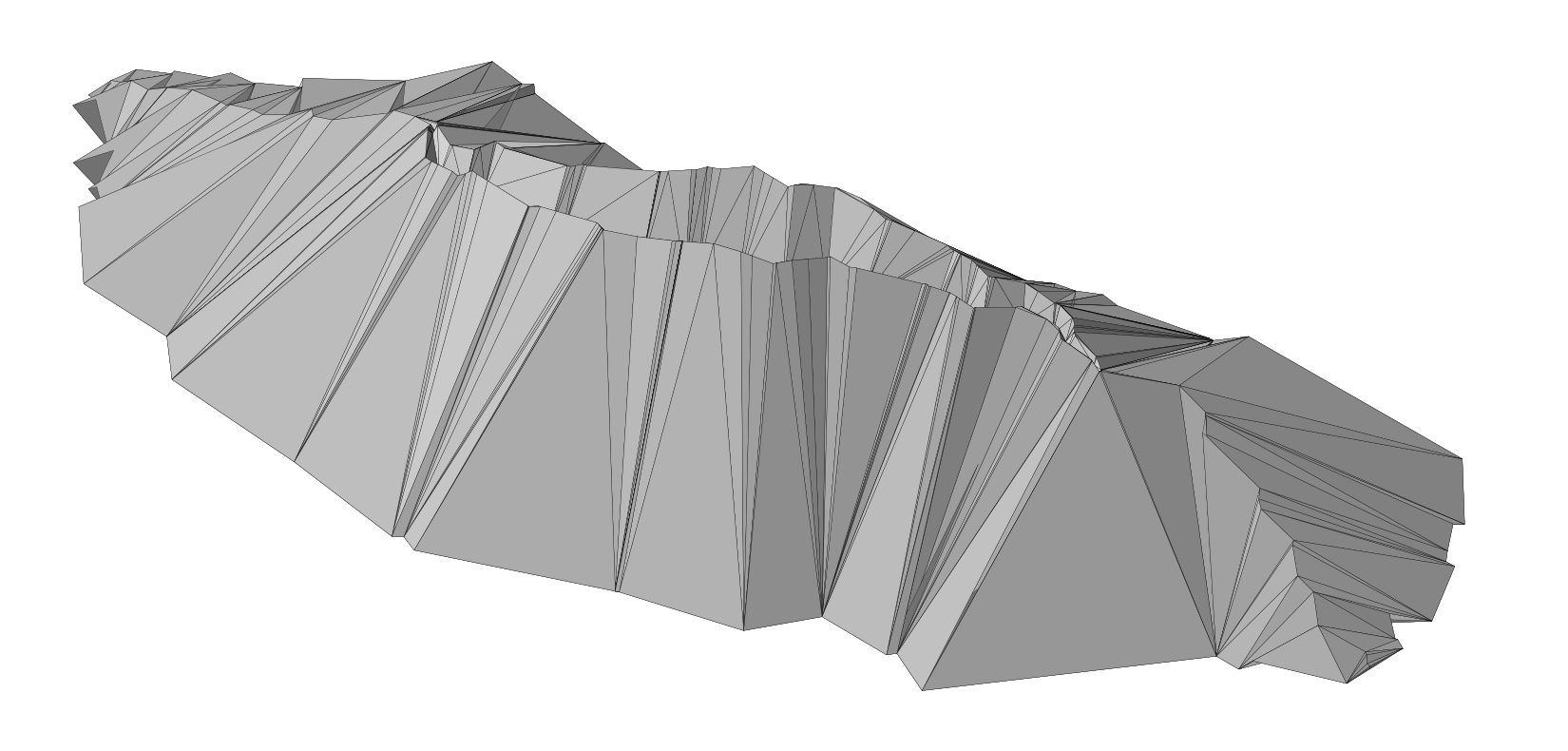}
            \caption{Buckle. Coloring of the top image: Minimal path in red, its small dilation in dark gray and its big dilation in light gray.}
            \label{fig:FlatDefects:buckle}
        \end{subfigure}
        \hfill
        \begin{subfigure}[t]{0.32\textwidth}
        \centering 
        \includegraphics[width=\textwidth]{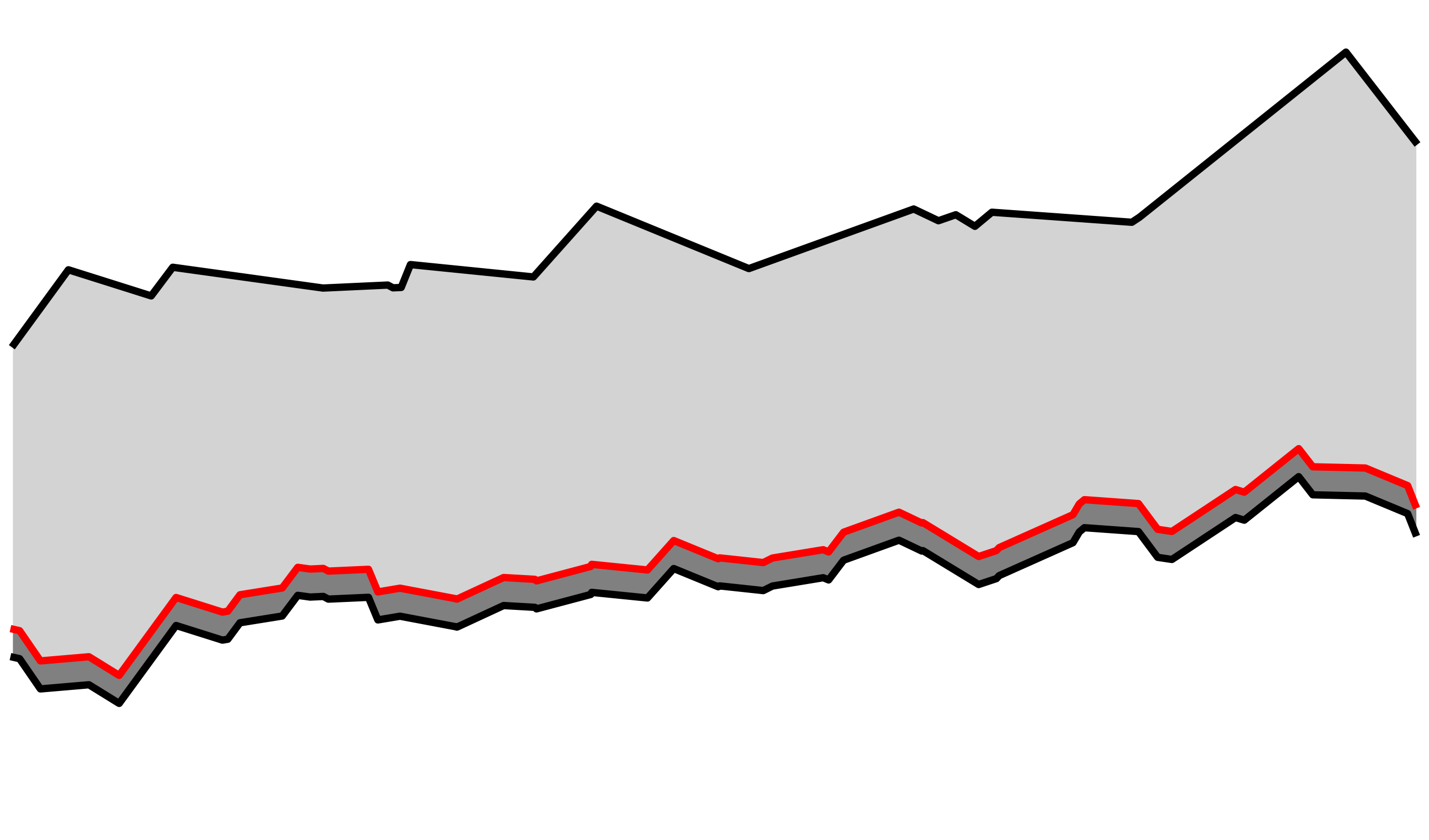}
        \includegraphics[width=\textwidth]{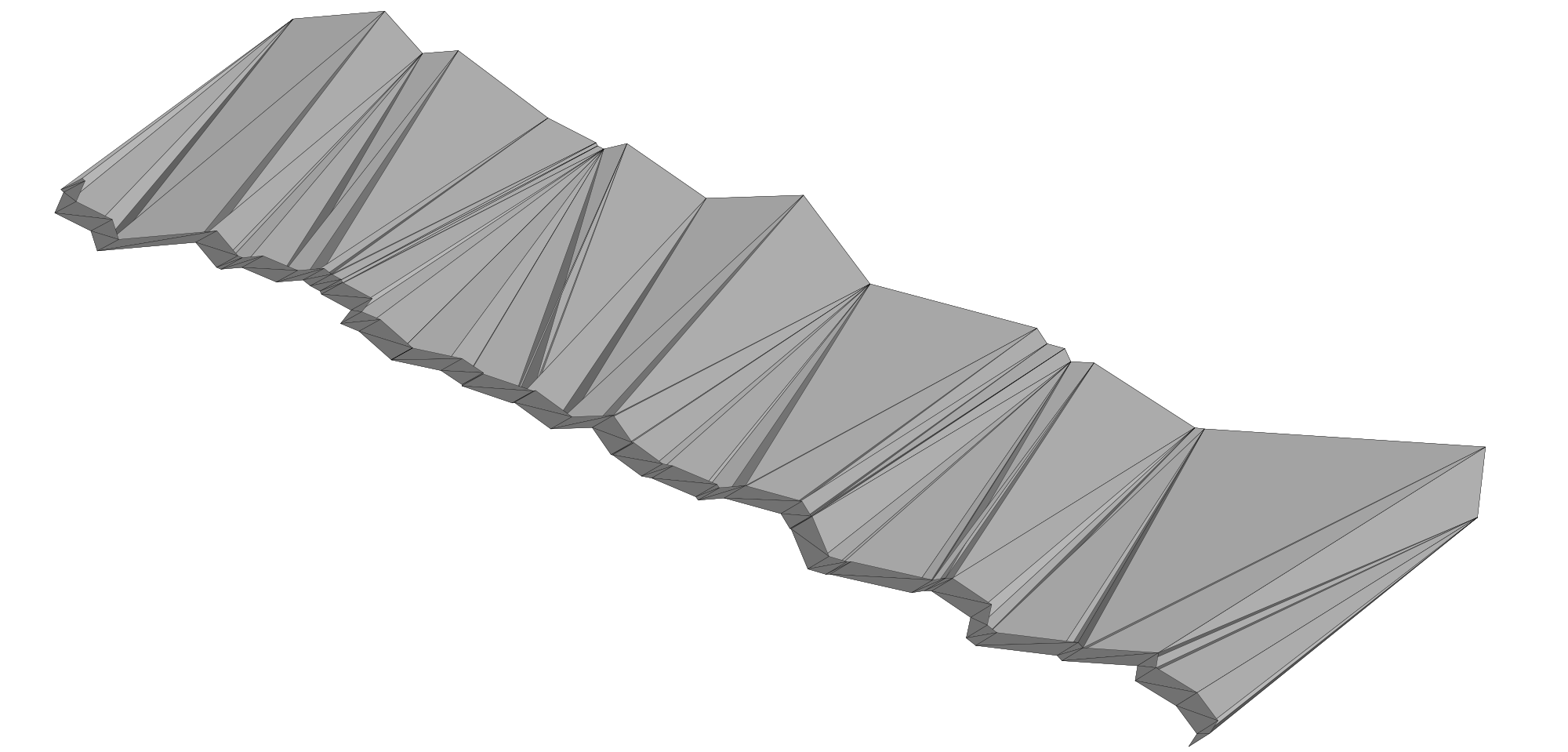}
        \caption{Coat lift. Coloring of the top image: Minimal path in red, its one-sided dilation in light gray and its shift in dark gray.}
        \label{fig:FlatDefects:coatlift}
        \end{subfigure}
        \caption{Examples of bulge, buckle and coat lift defects. Top: Visualization of the 2d model steps, model-relevant coloring defined in the corresponding sub-caption. Bottom: 3d mesh of the objects.}
        \label{fig:FlatDefects}
    \end{figure}

    A closed buckle can be modeled using the same approach as for bulges when choosing appropriate parameter configurations.
    Open buckles are modeled using a Boolean subtraction of a bulge-like and a crack-like defect. 
    Therefore, let $B_P$ be a positive defect shape generated by the bulge model described in Section \ref{sec:bulge} and $B_N$ be a negative defect shape generated by the crack model in Section \ref{sec:crack} generated under the following constraints:
    \begin{itemize}
        \item Both $B_P$ and $B_N$ are modeled using the same underlying minimal path.
        \item Let $l^P_i$ and $l^N_i$ be the widths of the positive and negative defects, respectively.
        We require $l^P_i>l^N_i$ for all $i$, so that the width of $B_P$ is greater than that of $B_N$ for every arc in the path.
        \item 
        The depth of the crack should be smaller than the height of the bulge such that the buckle is never deeper than the object surface.
    \end{itemize}
    Then, an open buckle is computed by $B_P\setminus B_N$.
    To perform that operation, re-meshing of the surface triangulation is necessary.
    Again, the Boolean difference function given in the \textsc{trimesh} package for Python can be used \cite{trimesh}.
    An example is given in Figure \ref{fig:FlatDefects:buckle}.

    \subsubsection{Coat lift}\label{sec:coatlift}
    Coat lifts are caused by buckles in the casting shell.
    Its primary layer lifts off and cracks, allowing melt to fill the space created \cite{AtlasCastingDefects}.
    Sertucha and Lacaze \cite{2022SertuchaDefectExp} describe lustrous carbon inclusions, which are surface defects that look similar, but arise in different ways.
    Coal is a common component of sand  molds. 
    Consequently, carbon films may form on the inner mold surface, which can then be enclosed by the metal melt. This results in protruding metal sheets after solidification.
    
    Coat lifts are therefore one-sided elevations on the surface of the object.
    Both the contours on the object surface and on the elevated side are jagged.
    The model therefore uses the same basis as the previous defect models, a minimal path through a Voronoi tessellation, but the subsequent steps differ, see Figure \ref{fig:FlatDefects:coatlift} for illustration.
       
    \begin{itemize}[left=0.5cm]
        \item[1.-3.] Steps 1-3 are equal to crack modeling in Section \ref{sec:crack}.
        \item[4.] Compute the upper dilation of the path with contour $P^\text{up}$.
        \item[5.] Compute a lower shift $P^\text{low}$ of the path $P$ with vertices $v_i^\text{low}=v_i-\left(0,l^\text{low}\right)$ using a constant width $l^\text{low}>0$. That width models the thickness of the elevated metal layer.
        \item[6.] 
        Assign the object surface height to the vertices of $P$ and $P^\text{low}$, namely $h_i=h_i^\text{low}=h$, and heights $h_i^\text{up}\geq h$ to the vertices of $P^\text{up}$.
        \item[7.] Compute the triangulation $T_\text{up}$ of the vertices belonging to $P$ and $P^\text{up}$, that are $$\left\{(v_0,h),\dots,(v_{k+1},h), (v_0^\text{up},h_0^\text{up}),\dots,(v_{k^\text{up}+1}^\text{up},h_{k^\text{up}}^\text{up})\right\},$$ to connect the elevated contour of the upper dilation and the path as described in Section \ref{sec:Triangulation}. 
        The triangulation $T_\text{low}$ of the area between $P$ and $P^\text{low}$ is given by the triangles $$\left\{T\left(v_i,v_{i+1},v_i^\text{low}\right)\right\}_{i=0,\dots,k}\cup\left\{T\left(v_{i+1}^\text{low},v_i^\text{low},v_{i+1}\right)\right\}_{i=0,\dots,k}.$$
        \item[8.] Compute the mesh of the coat lift $C=(V,T)$ with 
        \begin{align*}
            V&=\left\{\left(v_i,h\right)\right\}_{i=0,\dots,k+1}\cup\left\{\left(v_i^\text{up},h_i^\text{up}\right)\right\}_{i=0,\dots,k^\text{up}+1}\cup\left\{\left(v_i^\text{low},h\right)\right\}_{i=0,\dots,k^\text{low}+1},\\
            T&=T_\text{up}\cup T_\text{low}\cup T_\text{cover}\cup T_\text{end},
        \end{align*}
        where $T_\text{cover}$ connects the vertices of $P^\text{up}$ and $P^\text{low}$ to form the top coverage of the defect.  
        Triangles at the ends of the path are again added via $T_\text{end}$.
    \end{itemize}

    \subsubsection{Cold shut and rat tail}
    Aside from cracks, there exist other types of elongated defects that are curved instead of having a piecewise linear behavior.
    Cold shuts appear if two molten metal streams cannot merge before solidifying.
    This happens if the melt is not fluid enough, which can be caused by low temperature.
    Furthermore, rat tails arise from defected shells.
    For example, a crack in the mold that is too thin to be filled with melt contains air and causes oxidation of the hot metal, which forms an indentation in the object.
    While cold shuts commonly appear individually, rat tails occur accumulated.
    Moreover, rat tails are more curved.
    
\begin{figure}
        \centering
        \begin{subfigure}[t]{0.32\textwidth}
            \centering 
            \includegraphics[width=\textwidth]{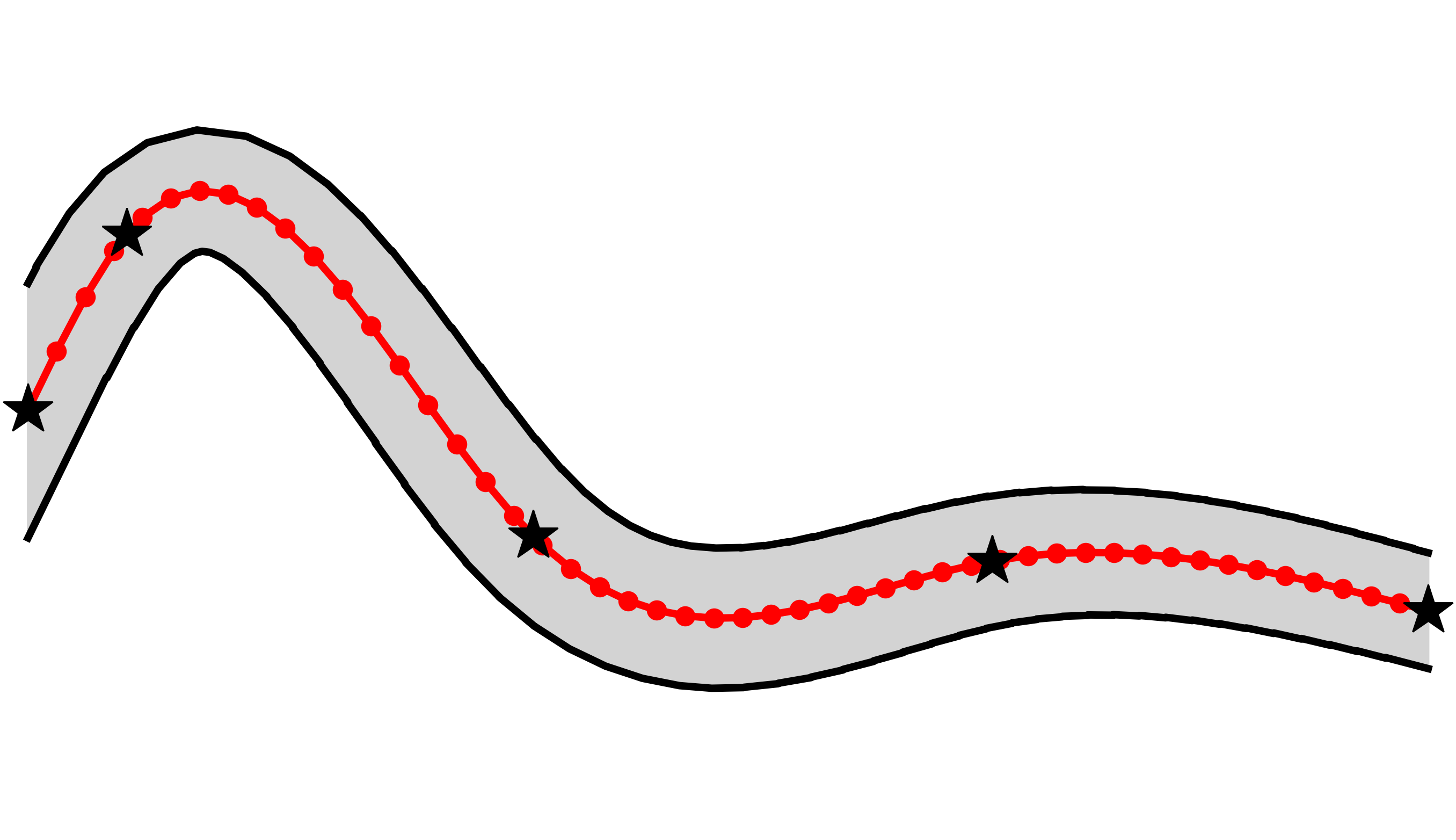}\\
            \includegraphics[width=\textwidth]{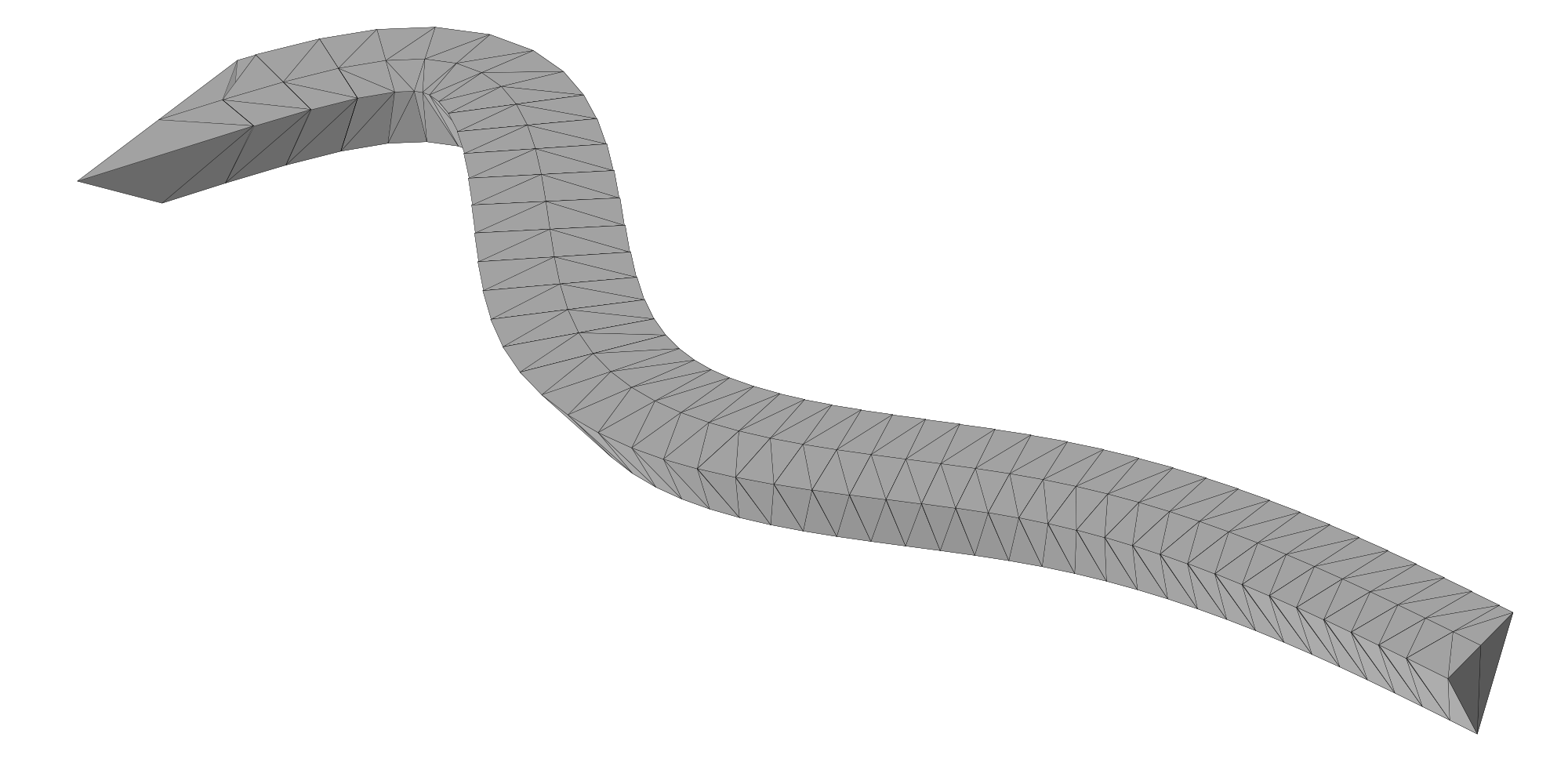}
            \caption*{$n+1=5$, $k+2=50$.}
        \end{subfigure}
        \hspace{4pt}
        \begin{subfigure}[t]{0.32\textwidth}
            \centering 
            \includegraphics[width=\textwidth]{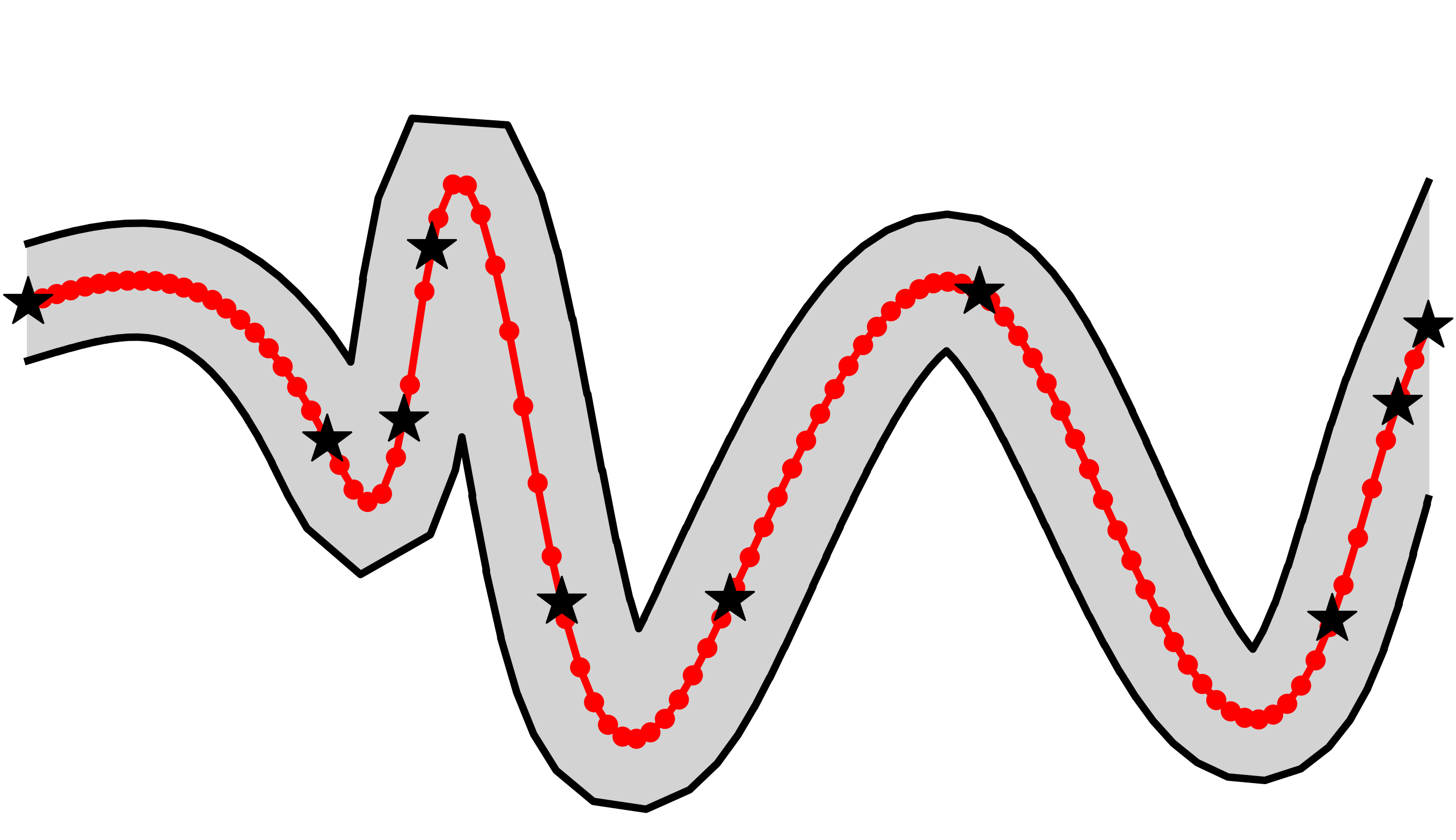}\\
            \includegraphics[width=\textwidth]{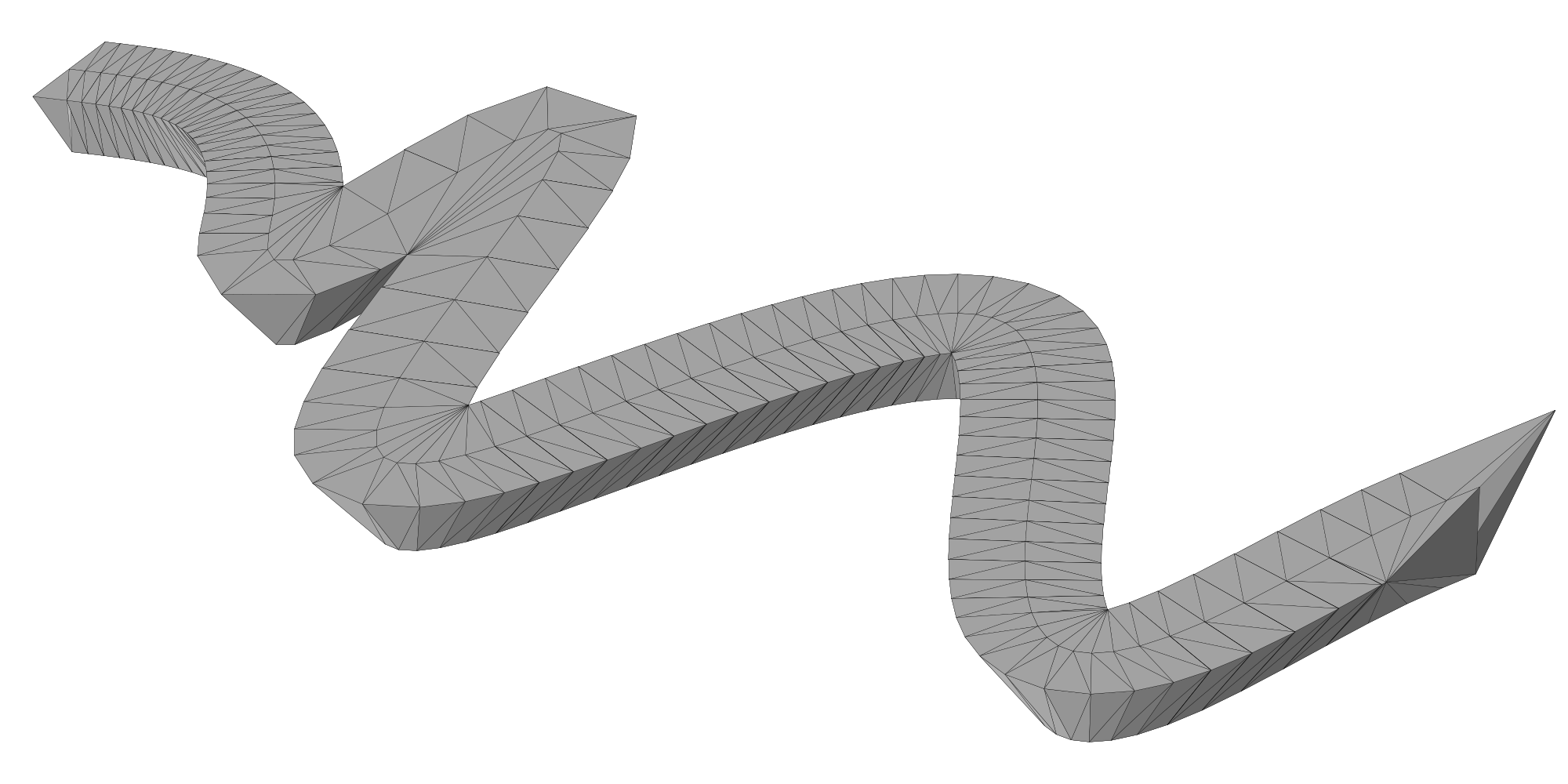}
            \caption*{$n+1=10$, $k+2=100$.}
        \end{subfigure}
        \caption{Realization of the model for cold shuts and rat tails using different parameter configurations. Top: Visualization of the generating procedure of the 2d shape with control points (stars), interpolation with discrete points (red) and the dilated path (black). Bottom: 3d mesh object.}
        \label{fig:coldshut}
    \end{figure}

    Although the causes of both types of defects are different, their appearance is similar.
    Thus, a joint model can be used by choosing suitable parameter configurations.
    The model is related to the crack model, but includes round edges.
    Therefore, instead of the piecewise linear path in the Voronoi tessellation, we use a natural cubic spline interpolation of nodes randomly chosen from a given observation window. 
    The resulting curve is twice continuously differentiable and the second derivative is zero at the curve's end vertices.
    The model is then defined by the following steps, see Figure \ref{fig:coldshut} for an illustration.

    \begin{itemize}[left=0.5cm]
        \item[1.] Choose a window $W$ such that $w_0^\text{max}-w_0^\text{min}$ equals the desired linear expansion of the defect.
        \item[2.] Generate $n+1$ random points $p_0, \ldots, p_n$ from a uniform distribution on $W$ and sort them according to their $x$-values. Generate the points such that the first and the last are located on the left and right border of the window, that is, $(p_0)_0=w_0^\text{min}$ and $(p_n)_0=w_0^\text{max}$.
        \item[3.] Compute a natural cubic spline interpolation using $\left\{p_0,\dots,p_n\right\}$ as control points. Discretize the curve by computing points $v_0,\dots,v_{k+1}$ with $k+1\gg n$ located on the curve with $v_0=p_0$ and $v_{k+1}=p_n$. Define a path by arcs $a_i$ with $\alpha_i=v_i$ and $\omega_i=v_{i+1}$ for $i=0,\dots, k$.
        \item[4.-7.] Steps 4-7 are equal to crack modeling in Section \ref{sec:crack}.
    \end{itemize}

    \section{Model for scab defects} \label{sec:surfDefects}
    Scabs usually occur if the top layer of the mold is defected such that molten metal can fill the space between the layers similarly to coat lifts \cite{2022SertuchaDefectExp}.
    In \cite{AtlasCastingDefects}, this type of defect is called delamination because the top layer of the mold is delaminated.
    In general, delaminations appear when the material breaks into layers.
    For the purpose of this paper, we only consider surface delaminations, that is, the breaking of the top surface layer into individual pieces whose outer regions can be elevated.
    Delaminations can occur individually, in clusters, or as surface-filling defects that cover the entire surface.
    Total surface fracturing rather commonly occurs for coated parts, but is unusual for cast metal products.
    We propose a general approach that is applicable for modeling scabs (similar to individual delaminations), clustered, and surface-filling delaminations.
	
\begin{figure}
    	\centering
    	\begin{subfigure}[t]{4cm}
    		\includegraphics[height=4cm]{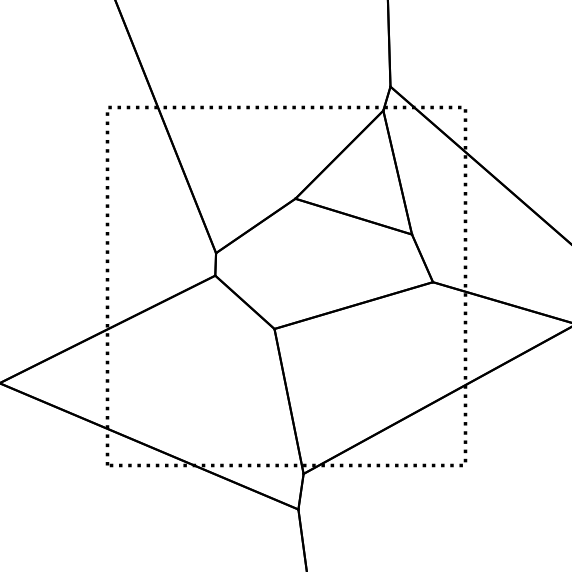}
    		\caption*{\centering Voronoi tessellation in a delaminated window.}
    	\end{subfigure}
    	\hspace{2pt}
    	\begin{subfigure}[t]{4cm}
        	\includegraphics[height=4cm]{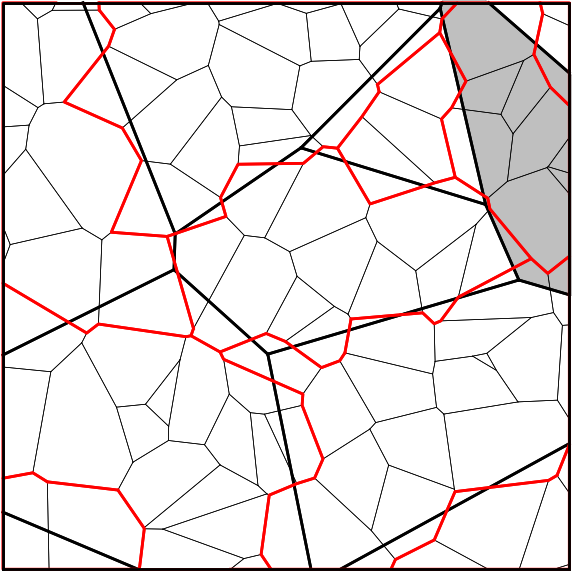}
        	\caption*{\centering Border refinement of individual cells.}
    	\end{subfigure}
    	\hspace{2pt}
    	\begin{subfigure}[t]{2.25cm}
        	\includegraphics[height=4cm]{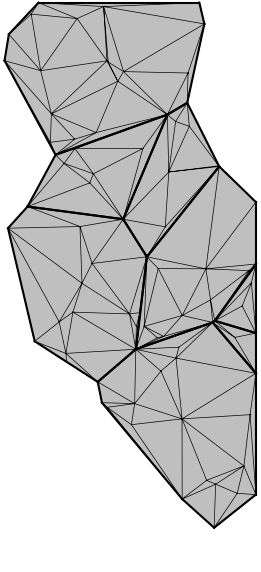}
        	\caption*{\centering Triangulation.}
    	\end{subfigure}
    	\hspace{2pt}
    	\begin{subfigure}[t]{4cm}
        	\includegraphics[height=4cm]{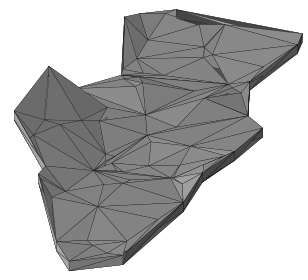}
        	\caption*{\centering Assignment of height values and 3d mesh.}
    	\end{subfigure}
    	\caption{Schematic overview of the model for scab and surface delamination defects.}
    	\label{fig:Delam_short}
    \end{figure}

\begin{figure}
		\centering
		\includegraphics[width=0.195\textwidth]{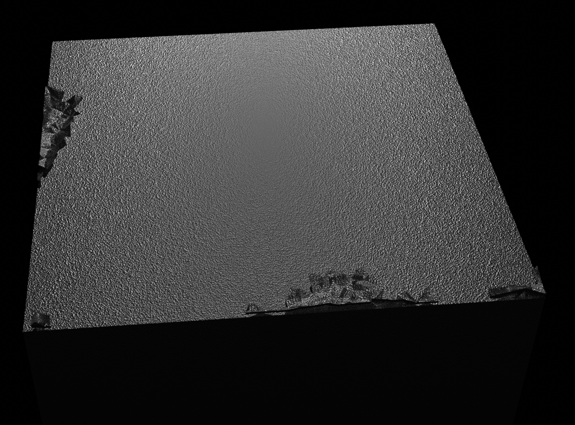}\hfill
		\includegraphics[width=0.195\textwidth]{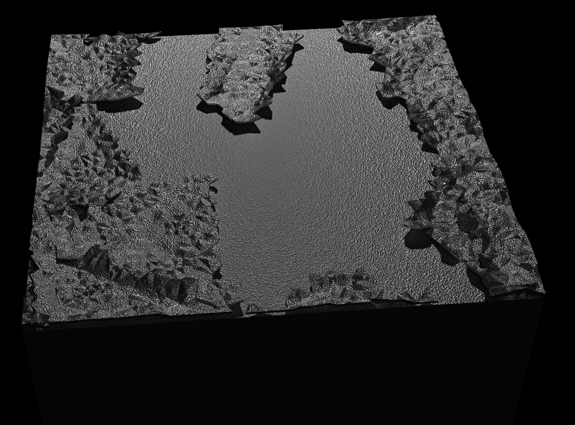}\hfill
		\includegraphics[width=0.195\textwidth]{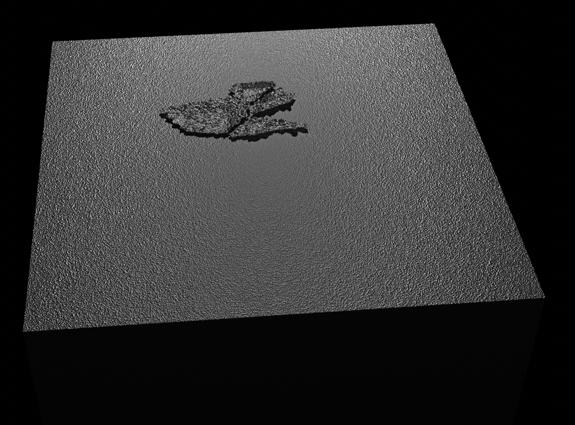}\hfill
		\includegraphics[width=0.195\textwidth]{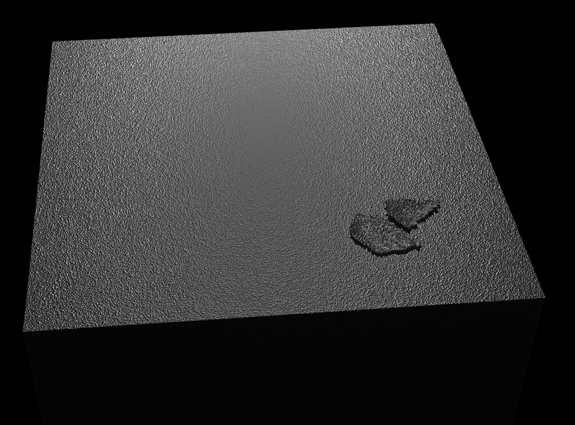}\hfill
		\includegraphics[width=0.195\textwidth]{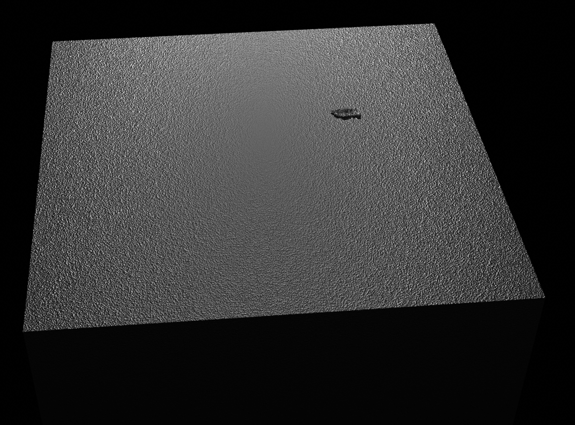}
		\caption{Synthetic images of delaminated surfaces of a cast metal cube rendered as described in Figure \ref{fig:linDefRend}.}
		\label{fig:DelamRend}
    \end{figure}

	Again, we use Voronoi cells for the simulation, which cover the whole surface by construction.
    Our model creates textures of delaminated surfaces by generating individual cells on the object surface whose outer regions may be elevated.
    Since these parts result from breaking of the surface layer, the edges of neighboring cells must fit together.
    Sometimes, just certain parts of the surface are affected by delaminations, for example, in scabs. 
    In this case, the delamination model can be restricted to a subset of the delamination cells.
    We first outline the main steps of the model generation procedure, see also Figure \ref{fig:Delam_short}. Details are explained in the subsequent sections.
    Figure \ref{fig:DelamRend} shows synthetic images of the different delamination types rendered with parameters of a realistic imaging setup.

    \begin{enumerate}
    	\item Select an observation window $W$ that represents the part of the object surface containing the delamination.
    	\item Compute a Voronoi tessellation in the dilated window to obtain the individual delamination cells. 
    	The dilation is necessary to obtain regular cell shapes even at the window borders.
    	\item Refine the cell borders by using a finer Voronoi tessellation and choose how many and which cells should be considered further.
    	\item For all chosen delamination cells:
    	\begin{enumerate}
    		\item[4.1.] Triangulate the refined cell using additional vertices.
    		\item[4.2.] Assign height values to all vertices.
    		\item[4.3.] Compute the 3d mesh.
    	\end{enumerate}
    \end{enumerate}

    \subsection{Voronoi tessellation in the dilated window} \label{sec:restricted_Voro}
    We call the Voronoi tessellation that defines the individual cells the coarse tessellation.    Let $\mathscr{P}^c=\left\{p_1^c,\dots,p_{n^c}^c\right\}$ be a set of generator points with $p_i^c$ uniformly sampled in the dilated window $\Delta_\gamma(W)$. 
    Let $D_i^c=C_i^c\cap W$ be the restriction of cell $C_i^c$ to the window.
    The number of vertices of $D_i^c$ is given by $\# D_i^c$.
    As some generator points can be located outside $W$, each nonempty cell $D_i^c$ is assigned its center point 
    \begin{equation*}
    	c_i^c=\frac{1}{\# D_i^c}\sum_{v \text{ vertex of } D_i^c} v
    \end{equation*}
    as reference points in $W$.
    Let $m^c\leq n^c$ be the number of remaining cells.
    An example is given in Figure \ref{Fig:Del1}.
       
    \subsection{Refinement of cell borders}
    Generate a finer Voronoi tessellation in the same manner as described in Section \ref{sec:restricted_Voro} but with $n^f \gg n^c$ generator points.
    Choose reference points $c^f_j$ as described above and define the adjusted reference points
    \begin{equation*}
        \tilde{c}_j^f=
        \begin{cases*}
            p_j^f &if $p_j^f\in W$\\
            c_j^f &else
        \end{cases*}
    \end{equation*}
    for $D_j^f\neq\emptyset$.
    To refine the borders of the coarse cells, pick all fine cells whose adjusted reference point is located in the coarse cell, that is, for cell $i\in\{1,\dots,m^c\}$ $$\mathscr{D}_i=\left\{D^f_j:\tilde{c}^f_j\in D_i^c, j=1,\dots,m^f\right\}.$$
    An example is given in Figure \ref{Fig:Del2}.  
    
    \subsection{Triangulation of fine cells}
    Let $D$ be a fine cell with vertices $v_1,\dots,v_K$ in clockwise order with $v_{K+1}=v_1$ and center point $c$. 
    To obtain a triangulation of $D$, additional points are added depending on the relative area of the cell.
    The number of newly added points is given by
    \begin{equation*}
        R=\left\lfloor \frac{\text{area}(D)}{\max_{j=1,\dots,m^f} \text{area}(D_j^f)}\cdot R_\text{max}\right\rceil,
    \end{equation*} 
    where $\lfloor\cdot\rceil$ means rounding to the nearest integer and $R_\text{max}\in\mathbb{N}$ is the number of new points for the largest cell in the fine Voronoi tessellation. 
    The computation of the triangulation $T^\text{2d}$ differs depending on $R$, see also Figure \ref{Fig:Del3} for an example.
    \begin{itemize}[left=0.9cm]
        \item[$\mathbf{R=0}$\textbf{:}] Compute a triangulation of the points $V^\text{2d}=\left\{v_1,\dots,v_K,c\right\}$ by $T^\text{2d}=\bigcup_{k=1}^K T(v_k,v_{k+1},c)$.
        \item[$\mathbf{R>0}$\textbf{:}] Generate new points  $u_r, r=1,\dots,R$ from a uniform distribution on $D$. Then, compute the Delaunay triangulation of the points $V^\text{2d}=\left\{v_1,\dots,v_K,c,u_1,\dots,u_R\right\}$, that is, no further points lie in the circumcircle of a triangle.
    \end{itemize}

    \subsection{Assignment of height values}
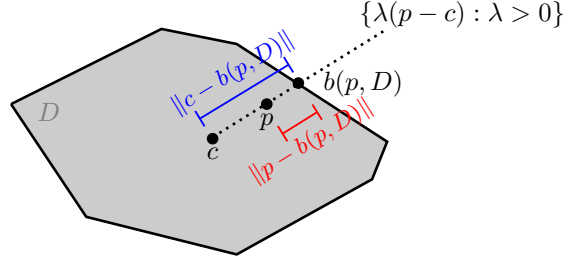
\begin{figure}
    	\centering
    	\begin{tikzpicture}
    		\draw[fill=gray!40,line width=1pt] (0,0) -- (2,1) -- (3,0.8) -- (5,-0.5) -- (4.8,-1) -- (3,-2) -- (1,-1.5) -- (0,0);
    		\draw[fill=black] (2.68,-0.46) circle (0.07) node[below] {$c$};
    		\node at (0.5,-0.1) {\textcolor{gray}{$D$}};
    		\draw[dotted,line width=1pt] (2.68,-0.44) -- (5,1);
    		\node at (6,1.2) {$\{\lambda(p-c):\lambda>0\}$};
    		\draw[fill=black] (3.82,0.28) circle (0.07) node[right] {\hspace{6pt}$b(p,D)$};
    		\draw[fill=black] (3.4,0) circle (0.07) node[below] {$p$};
    		\draw[red,|-|,line width=0.7pt] (3.6,-0.4) -- (4.08,-0.1) node[below,midway,sloped] {\footnotesize $\Vert p-b(p,D)\Vert$};;
    		\draw[blue,|-|,line width=0.7pt] (2.5,-0.2) -- (3.7,0.53) node[above,midway,sloped] {\footnotesize $\Vert c-b(p,D)\Vert$};
    	\end{tikzpicture}
    	\caption{Illustration of intersection $b(p,D)$ and computation of $\text{dist}_\text{rel}\left(p,\partial D\right)$.}
    	\label{fig:intersection}
    \end{figure}

    Height values H(v) are assigned to all vertices $v$ in the triangulation of $\mathscr{D}_i$, that is the union of triangulations of the associated fine cells, by 
    \begin{equation*}
        v\mapsto h+H_\text{elev}(v)+H_\text{tex}(v)
    \end{equation*}
    where $H_\text{elev}(\cdot)$ determines the amount of elevation dependent on the distance to the boundary of $D_i^c$ and $H_\text{tex}(\cdot)$ gives the opportunity to add a surface texture depending on the fine cell that $v$ is located in, see also Figure \ref{Fig:Del5}.
    To define both height functions, let $b(p,D) = \partial D\cap \left\{\lambda(p-c): \lambda >0 \right\}$ be the intersection of the border of cell $D$ and the half-line starting in the center point $c$ that passes through an arbitrary point $p\in\mathbb{R}^2$. 
    See Figure \ref{fig:intersection} for an illustration. 
    The height in $v$ will depend on the distance between $b(v,D)$ and $v$ as described in the following sections.

    \subsubsection{Elevation of outer areas $H_\text{elev}(\cdot)$ }
        This height models the defect-specific shape, that is, the outer part of a cell $\mathscr{D}_i$ is elevated while its center is not, so that the defect remains connected to the object surface. 
        To ensure a natural appearance of the defect, the maximum elevation heights at the boundary of $\mathscr{D}_i$ and the diameter of the non-elevated area in the center of the cell can vary for different directions in $\mathscr{D}_i$.
        
        The size of the non-elevated area in direction of a vertex $v\in\mathscr{D}_i$ is characterized by the distance $\Vert v-b\left(v,D_i^c\right)\Vert$.
        Since that distance varies when traversing over the cell's contour, we use the relative distance
        \begin{equation*}
        	\text{dist}_\text{rel}\left(p,\partial D_i^c\right)=\frac{\Vert p-b\left(p,D_i^c\right)\Vert}{\Vert c^c_i - b\left(p,D_i^c\right)\Vert}
        	\hspace{5pt}
        	\begin{cases*}
        		\leq 1 &if $p\in D_i^c$\\
        		> 1 &else
        	\end{cases*}
        \end{equation*}
        for arbitrary $p\in\mathbb{R}^2$. 
        Note that $\text{dist}_\text{rel}\left(v,\partial D_i^c\right)>1$ may occur because $\mathscr{D}_i\nsubseteq D_i^c$.
        Vertex $v$ will be elevated if $\text{dist}_\text{rel}\left(v,\partial D_i^c\right)$ exceeds a certain threshold $d(v)\in[0,1]$.
        Thus, all vertices $v\notin D_i^c$ are elevated for sure.

        By setting $H_\text{elev}(v)=0$ for all vertices with $\text{dist}_\text{rel}\left(v,\partial D_i^c\right)\leq d(v)$, all such vertices stay at the height of the object surface.
        Height values for all points on the contour of $\mathscr{D}_i$ must be given to define the maximal elevation.
        Then, all vertices located on the contour are assigned these maximal height values.
        The heights of the remaining intermediate vertices between the non-elevated area and the contour are obtained by linear interpolation.

    \subsubsection{Texture height $H_\text{tex}(\cdot)$}
        This height component gives the opportunity to assign a specific texture to the fine cells by individually elevating them using local bulges with maximal height in the center points of the cells.
        Assume that a vertex $v$ is contained in the fine cell $D^f_j$ and choose a maximal height $t_j\in\mathbb{R}_{\geq 0}$ reached at $c^f_j$. 
        The height at $v$ then depends on its relative distance to the center point $c^f_j$ and is  thus defined via. 
        \begin{equation*}
            H_\text{tex}(v)=
            t_j\cdot\frac{1}{2}\left(\cos\left(\pi\frac{\left\Vert c^f_j - v\right\Vert}{\left\Vert c^f_j - b\left(v,D^f_j\right) \right\Vert}\right)+1\right) \in [0, t_j]
        \end{equation*}
        Note that also other shape functions may be used.

    \subsection{Computation of 3d mesh}
\begin{figure}
        \centering
        \begin{subfigure}[t]{0.36\textwidth}
            \centering
            \includegraphics[height=5.2cm]{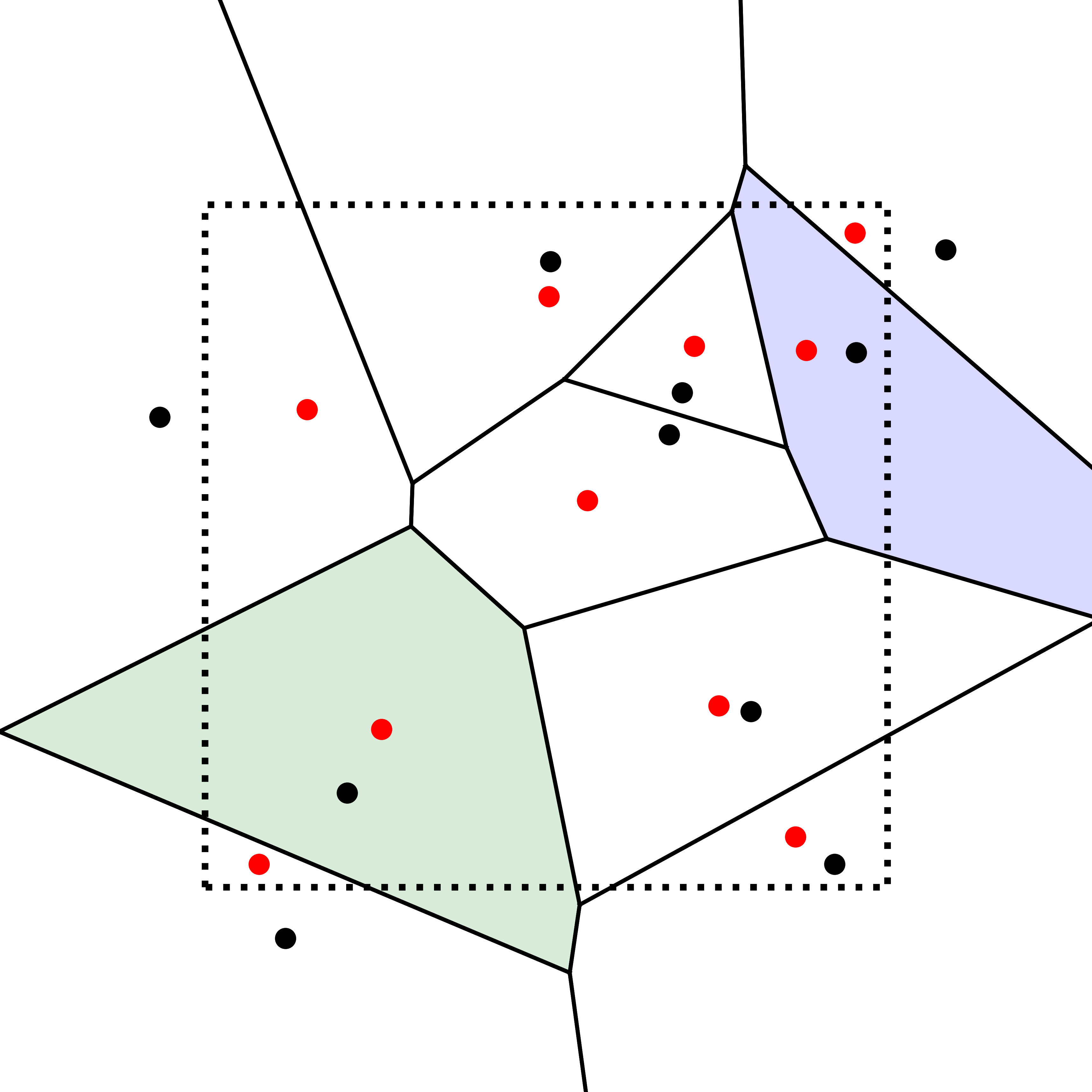}
            \caption{Procedure to restrict Voronoi tessellation to the window (dotted line). Generator (black) and reference points (red) are marked.}
            \label{Fig:Del1}
        \end{subfigure}
        \hfill
        \begin{subfigure}[t]{0.36\textwidth}
            \centering
            \includegraphics[height=5.2cm]{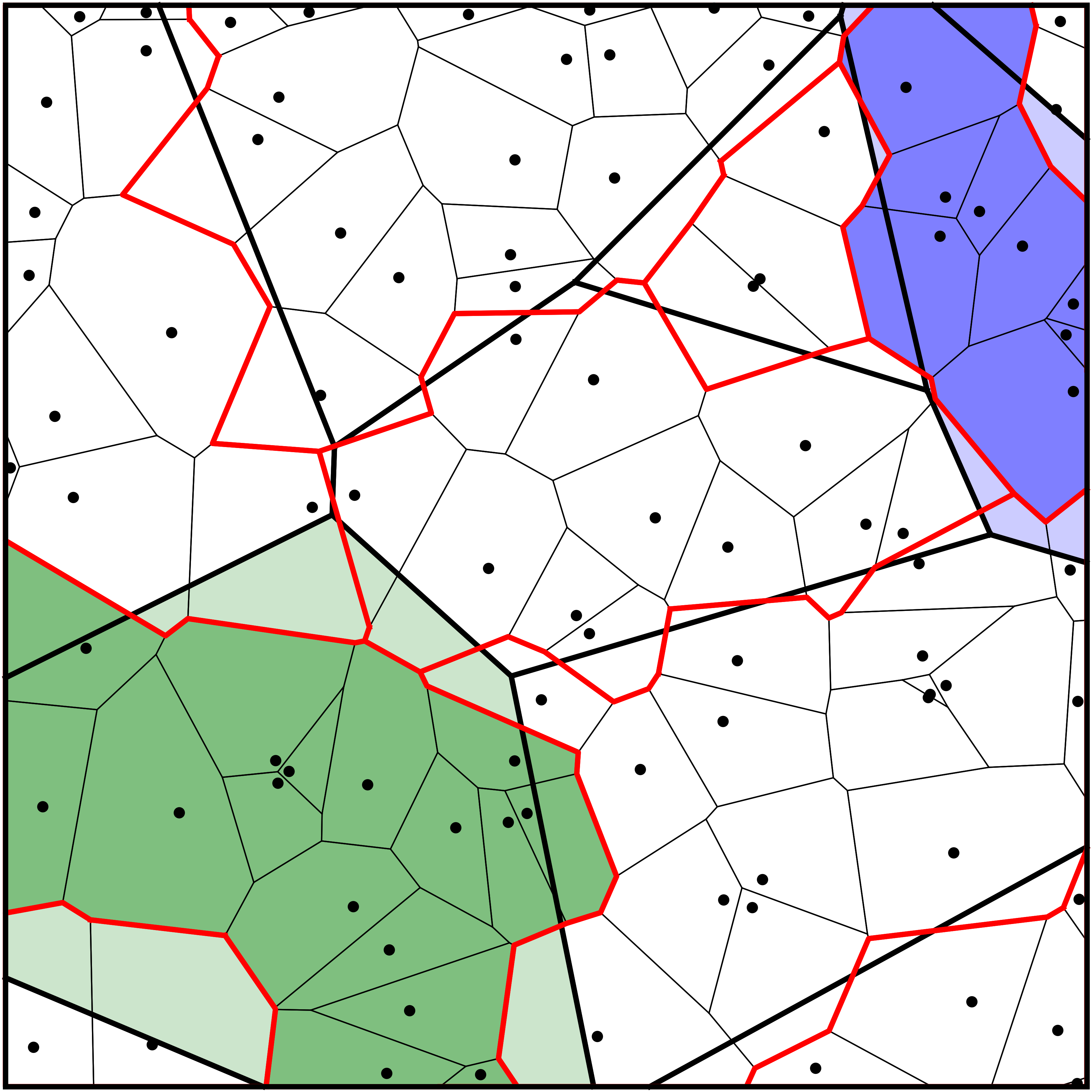}
            \caption{Coarse and fine Voronoi tessellation are represented by bold and thin black lines, respectively. Adjusted reference points of the fine tessellation are marked. Borders of $\mathscr{D}_i$ are illustrated in red. Two cells are highlighted to mark the change from $D_i$ (light) to $\mathscr{D}_i$ (dark).}
            \label{Fig:Del2}
        \end{subfigure}
        \hfill
        \begin{subfigure}[t]{0.25\textwidth}
            \centering
            \includegraphics[height=5.2cm]{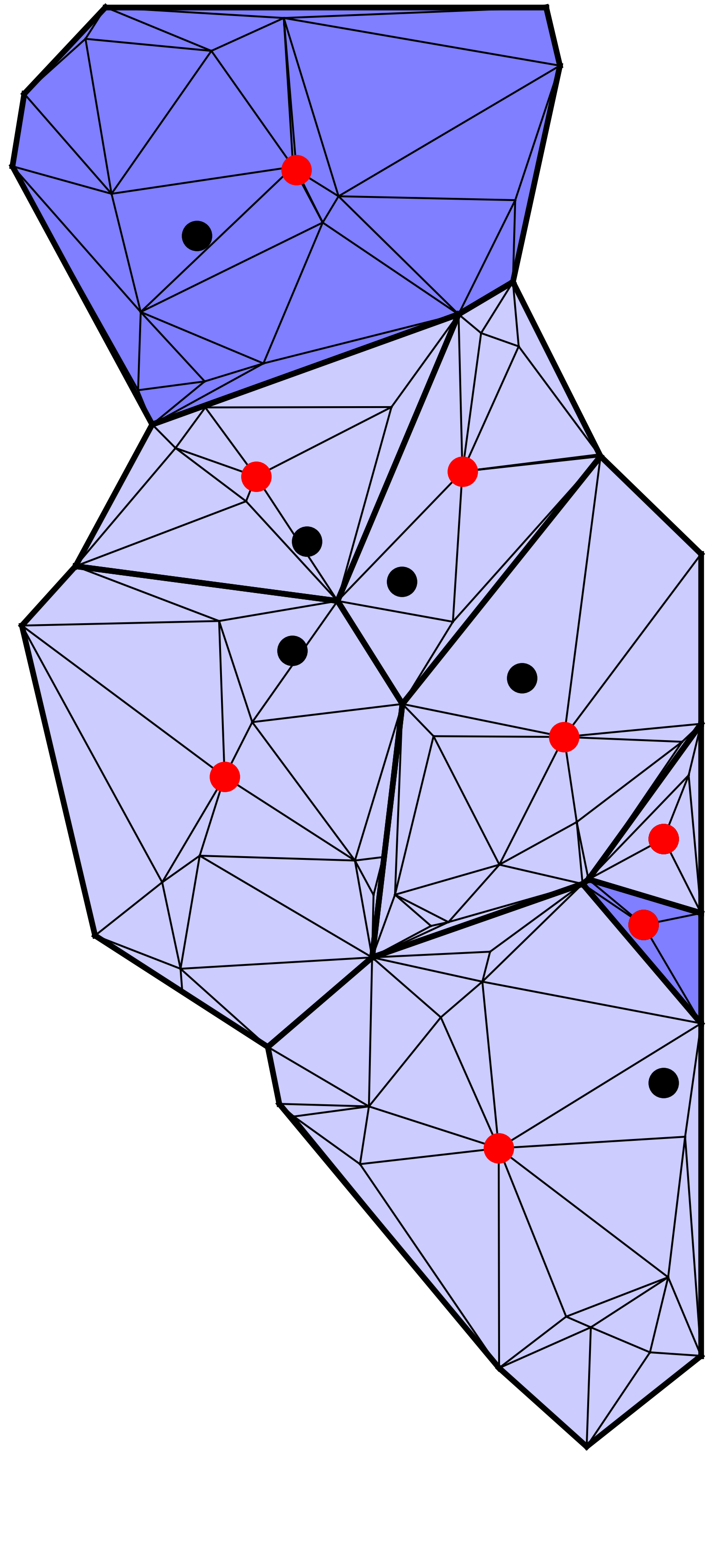}
            \caption{Triangulation of $\mathscr{D}_i$ using $R_\text{max}$=12. 
                The upper highlighted cell gets $R=11$ new points while for the lower highlighted cell we have $R=0$. Reference points are marked in red.}
            \label{Fig:Del3}
        \end{subfigure}\\[4mm]
        \begin{subfigure}[b]{0.535\textwidth}
            \centering
            \subcaptionbox*{$H_\text{elev}$}{\includegraphics[width=0.32\textwidth]{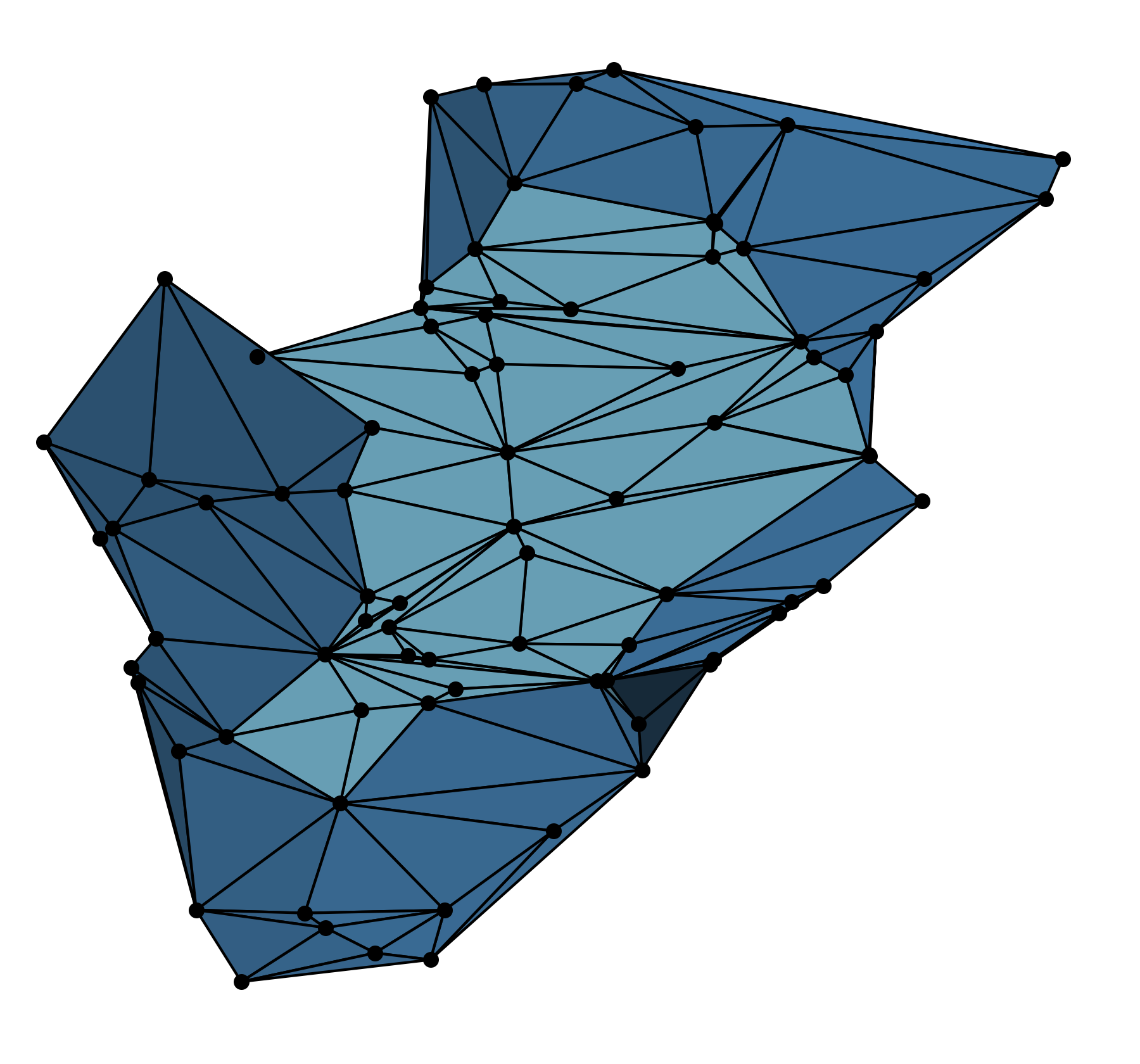}}
            \subcaptionbox*{$H_\text{tex}$}{\includegraphics[width=0.32\textwidth]{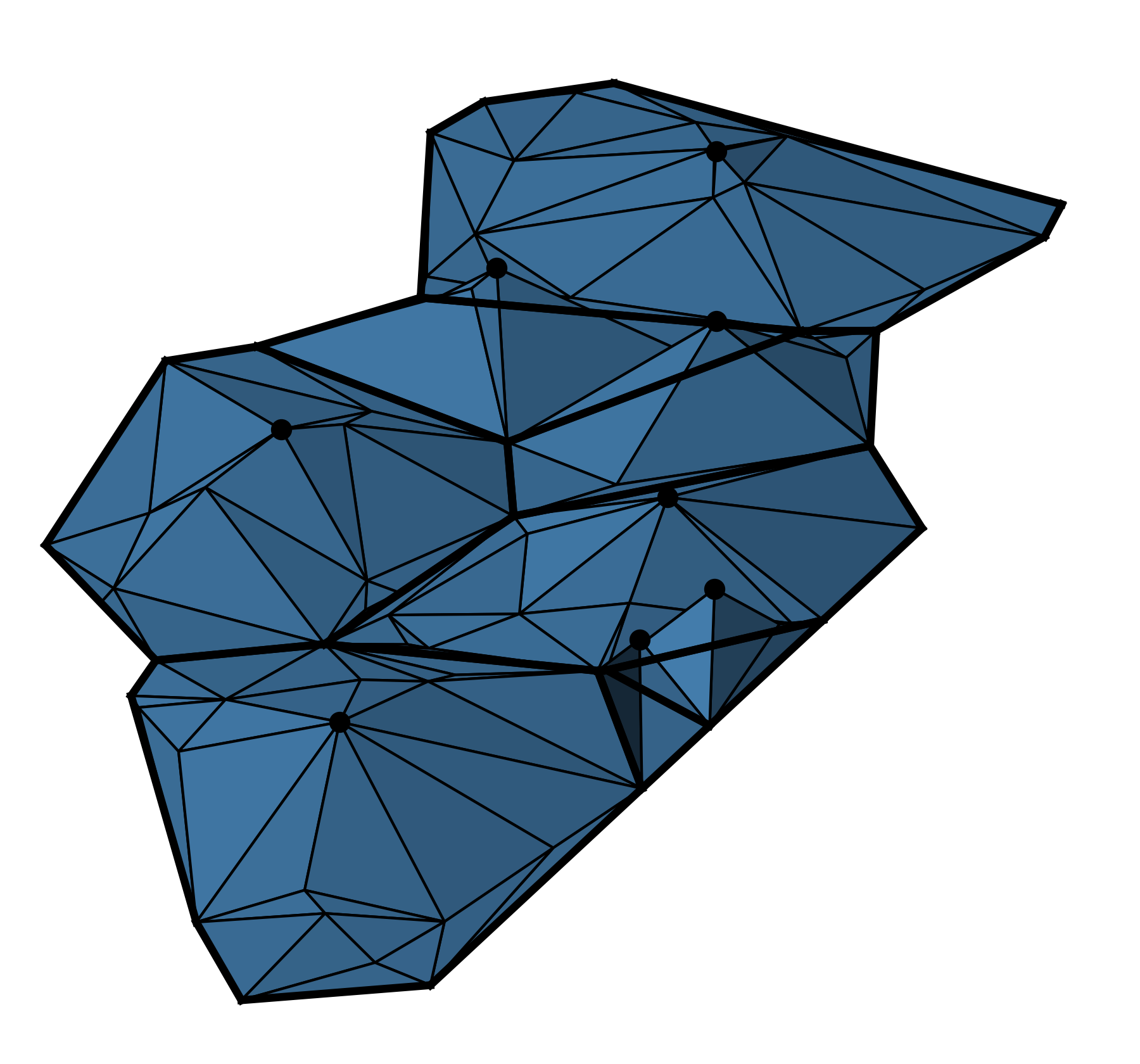}}
            \subcaptionbox*{$H_\text{elev}+H_\text{tex}$}{\includegraphics[width=0.32\textwidth]{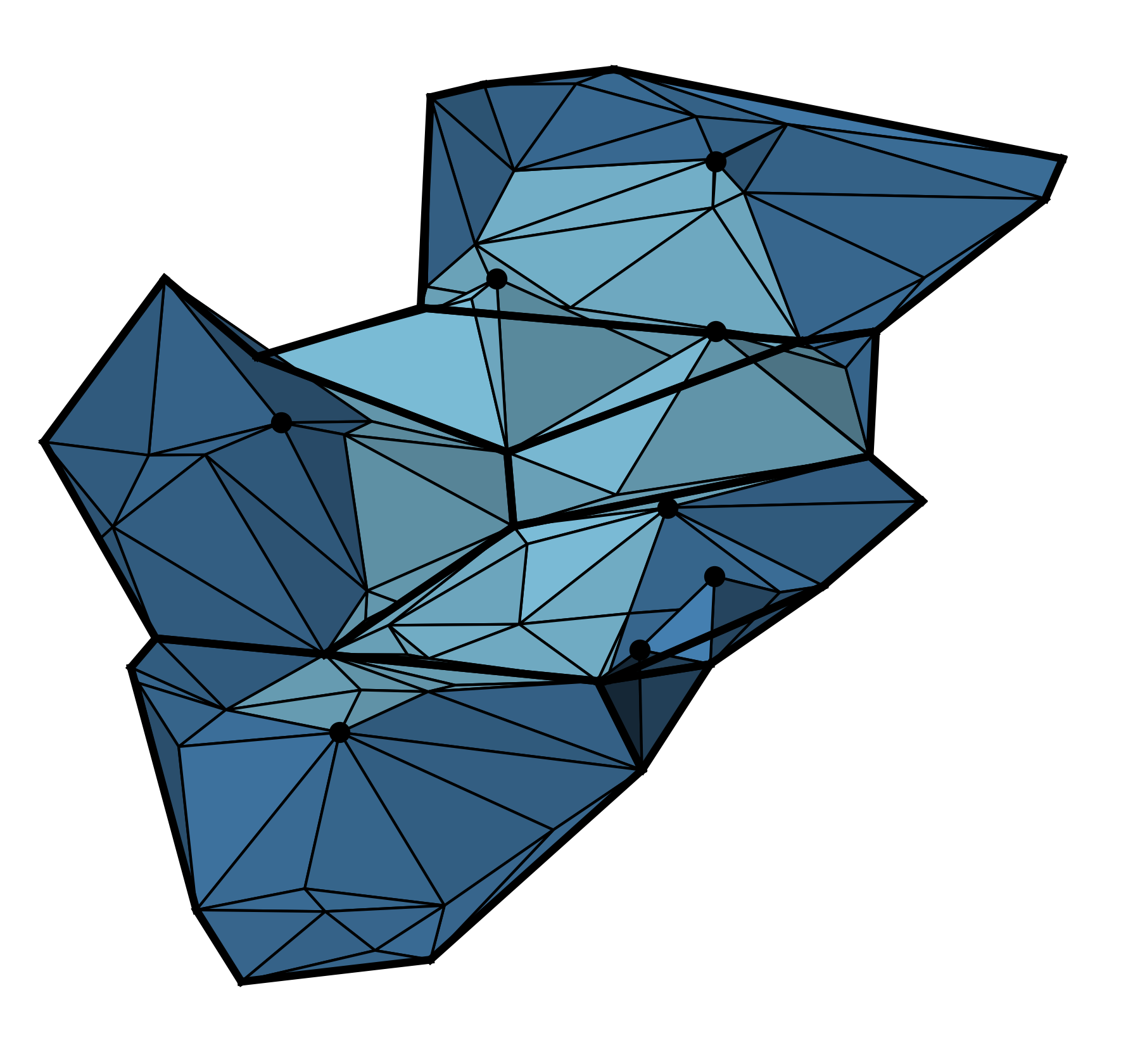}}
            \caption{Assignment of height values by the composition into elevation and texture. Dark blue marks the elevated vertices.}
            \label{Fig:Del5}
        \end{subfigure}\hfill   
        \begin{subfigure}[b]{0.455\textwidth}
            \centering
            \includegraphics[width=\textwidth]{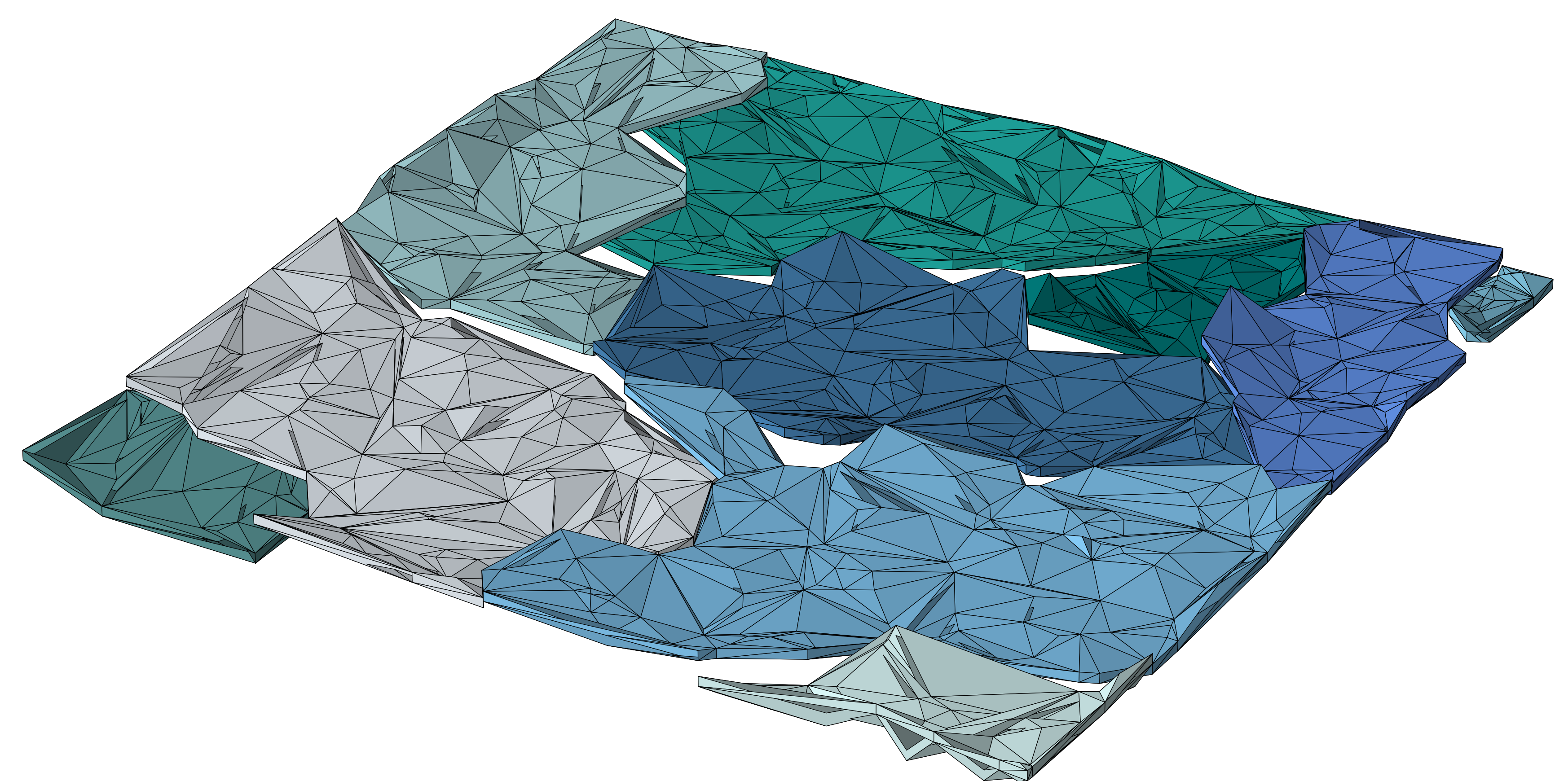}
            \caption{All delamination cells of the Voronoi tessellation as separate 3d volumes.}
            \label{Fig:Del6}
        \end{subfigure}
        \caption{Visualization of the pipeline to generate a delaminated surface or scabs with $n^c=10$, $n^f=110$ and a squared window with dilation width $\gamma=0.1\left(w^\text{max}-w^\text{min}\right)$.}
        \label{Fig:Del}
    \end{figure}

    To generate the 3d mesh of the delamination cell determined by $\mathscr{D}_i$, its vertices and triangles are duplicated first to define the lower and the upper layer. 
    In the first copy, the vertices get height values $h+H_{elev}(v)$.
    The texture height is removed such that the non-elevated cell parts are smoothly connected to the object surface.
    The second copy of the vertices gets the height values $H(v)+l$ where $l\in\mathbb{R}_{>0}$ is the thickness of the delaminated surface layer. 
    Finally, additional triangles are added to connect the upper and lower layer, similar to the coat lifts described in Section \ref{sec:coatlift}.
    Examples of 3d meshes of delaminated cells are given in Figure \ref{Fig:Del6}.
    
    \section{Conclusion}
    We presented parametric models to generate a variety of different surface defects of cast metal objects.
    We use Voronoi tessellations to simulate the fundamental defect behavior.
    Thus, the poly-crystalline material structure is included, but on a coarse scale to reduce the model complexity compared to physically correct defect models.
    This enables fast computation times for both the individual defect geometries and their addition to the product geometry.

    Defect variability is guaranteed by using different parameter configurations.
    Providing real parameter ranges increases the realism of defect shapes.
    Moreover, defect edge case scenarios are covered by using appropriate parameter configurations.
    Thus, a diverse and arbitrarily large synthetic data set can be generated. 
    In addition, particular defect features can be varied individually to fine-tune defect detection algorithms.
    This provides a great advantage over AI-based image generation methods, where systematic data generation is difficult.
    Automatic data annotation is also provided.

    Although our models are motivated by defect geometries for cast metal objects, they can be transferred to other manufacturing processes.
    For example, cracks are frequently occurring defects in various applications and delaminations appear commonly on painted surfaces.
    Synthetic data generation is also not limited to visual surface inspection.
    Placing the digital twins of the inspected object into another virtual inspection environment can provide synthetic data for any NDT method.

\vspace{0.5cm}
\textbf{Funding} This work was supported by the Ministry of Science and Health of the State of Rhineland-Palatinate by the project eQuality.

\bibliographystyle{unsrt}  
\bibliography{lit}

\end{document}